\PassOptionsToPackage{usenames, dvipsnames}{xcolor}
\documentclass[sigconf]{acmart}

\AtBeginDocument{%
  }

\setcopyright{acmlicensed}
\copyrightyear{2026}
\acmYear{2026}
\acmDOI{XXXXXXX.XXXXXXX}
\acmConference[Conference acronym 'XX]{Make sure to enter the correct
  conference title from your rights confirmation email}{June 03--05,
  2018}{Woodstock, NY}
\acmISBN{978-1-4503-XXXX-X/2018/06}

\acmSubmissionID{123-A56-BU3}

\usepackage{hyperref}
\usepackage{url}
\usepackage{booktabs}
\usepackage{makecell}
\usepackage{multirow}
\usepackage{amsmath}

\usepackage{amssymb}
\usepackage[table]{xcolor}
\usepackage{graphicx} 
\usepackage{enumitem}
\usepackage{tabularx}
\usepackage{amsthm}
\usepackage{algpseudocode}
\usepackage{hyperref}
\usepackage{wrapfig}
\usepackage{algorithm}
\usepackage{cleveref}
\usepackage{enumitem}
\usepackage{listings}
\usepackage{subcaption}
\usepackage[most]{tcolorbox}
\usepackage{tikz}

\usepackage[export]{adjustbox}

\usepackage{booktabs}   
\usepackage{longtable}  
\usepackage{xurl}

\definecolor{bestorange}{HTML}{FF8000}
\newcommand{\best}[1]{\textcolor{bestorange}{\textbf{#1}}}

\newcommand{\ico}[1]{\adjustbox{valign=c}{\includegraphics[height=0.95em]{#1}}}

\newcommand{\icoC}{\ico{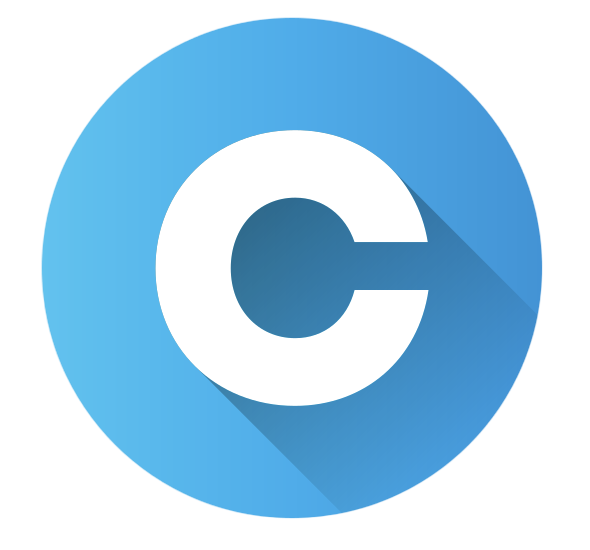}}
\newcommand{\icoCpp}{\ico{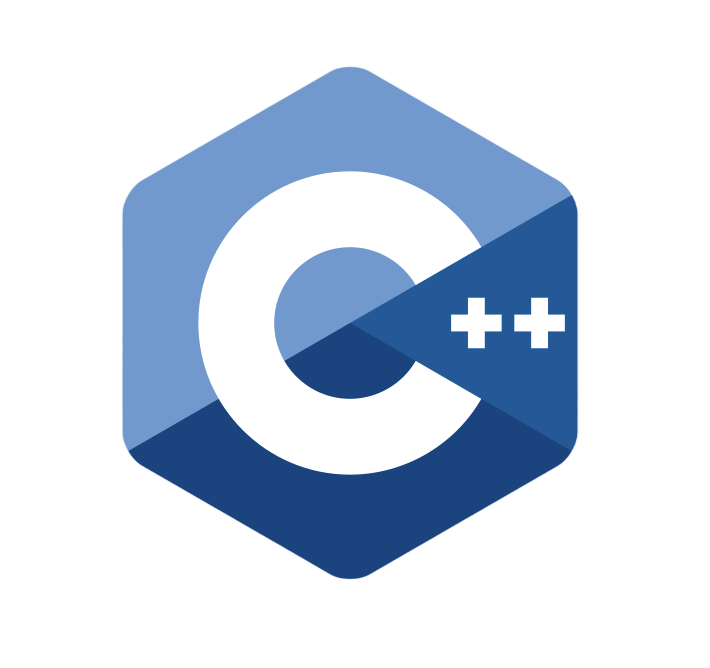}}
\newcommand{\icoGo}{\ico{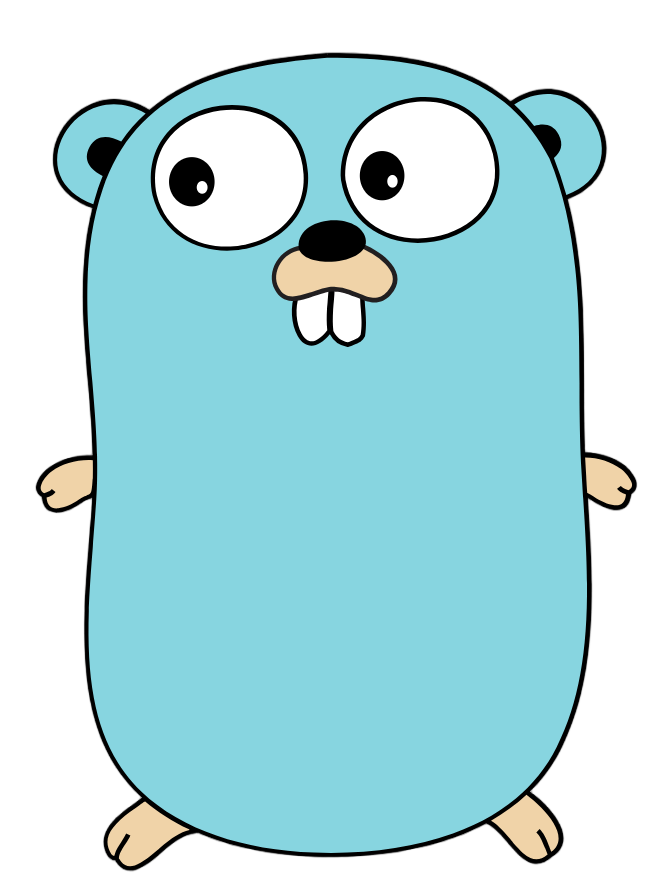}}
\newcommand{\icoGREPO}{\ico{GREPO.png}}
\newcommand{\icoJava}{\ico{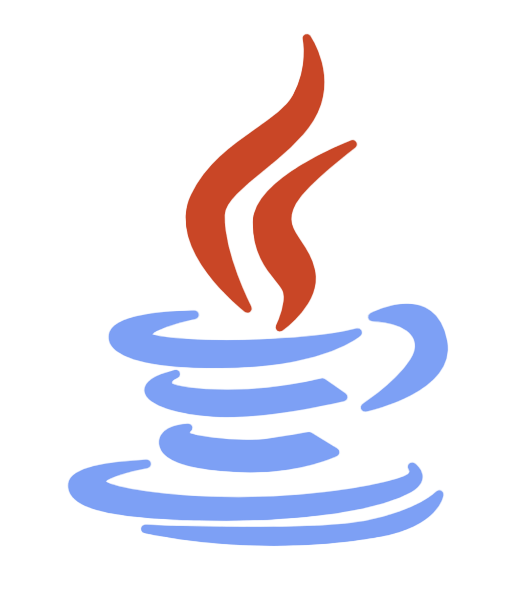}}
\newcommand{\icoJS}{\ico{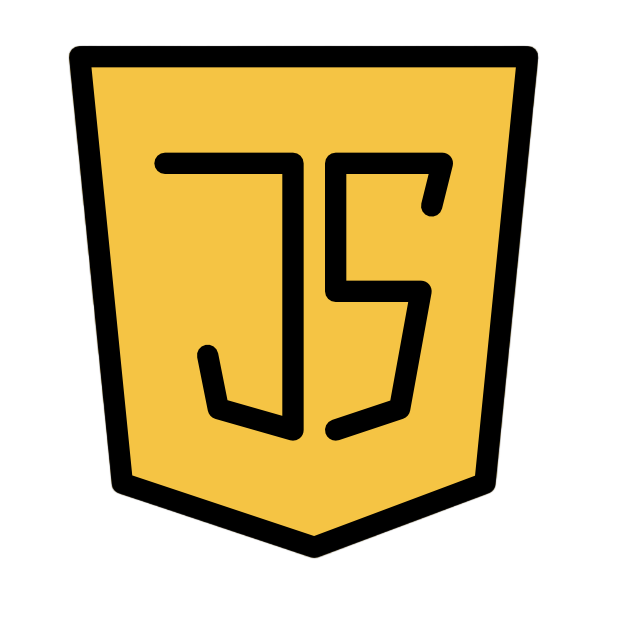}}
\newcommand{\icoPython}{\ico{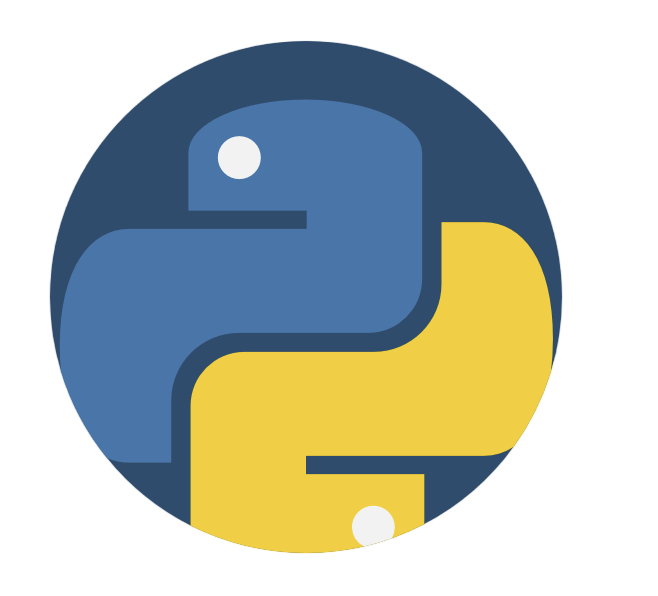}}
\newcommand{\icoRust}{\ico{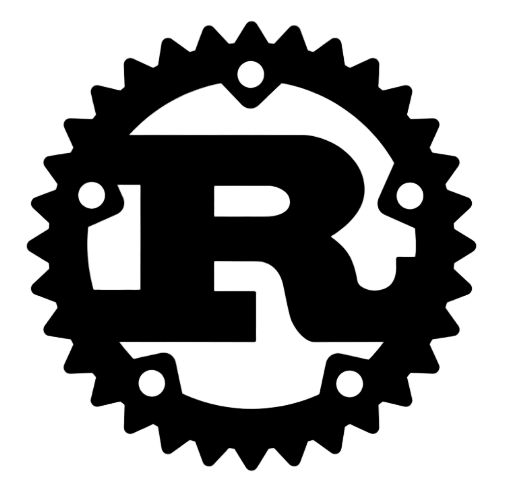}}
\newcommand{\icoTS}{\ico{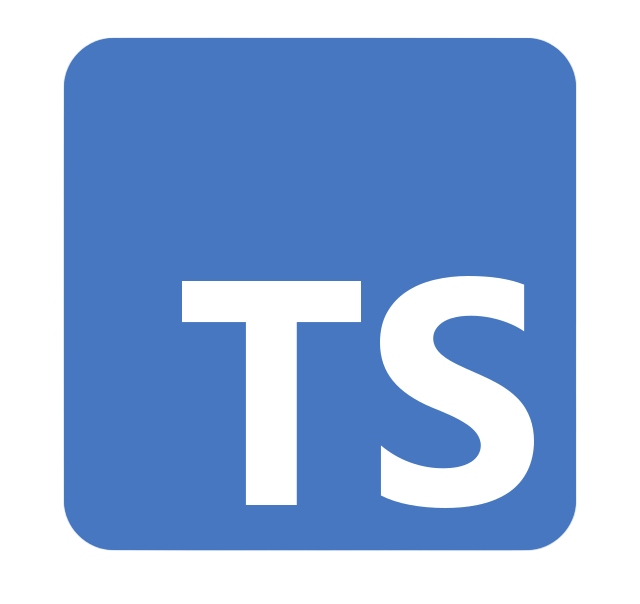}}
\newcommand{\icoCheck}{\ico{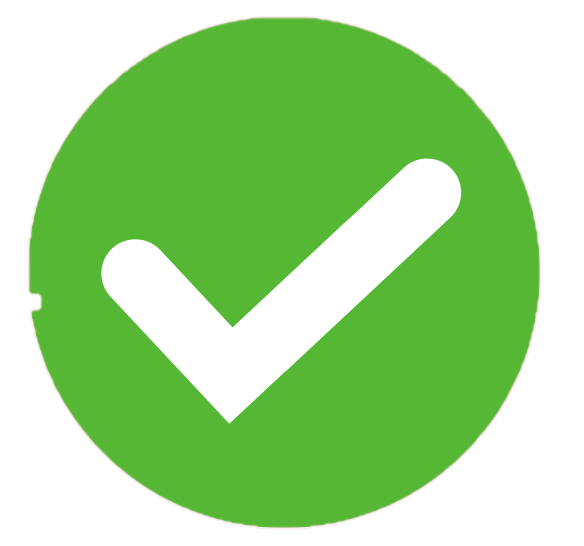}}
\newcommand{\icoCross}{\ico{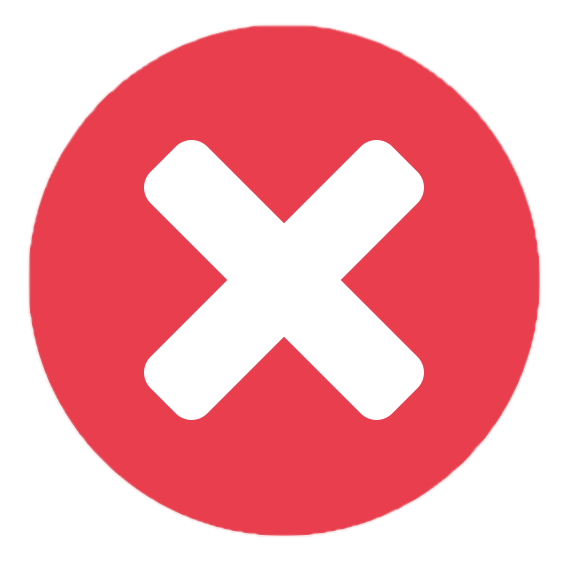}}

\newcommand{\taskBL}{\textsf{BL}}
\newcommand{\taskIR}{\textsf{IR}}
\newcommand{\taskAC}{\textsf{AC}}
\newcommand{\taskGEN}{\textsf{GEN}}

\definecolor{yucky}{HTML}{A52A2A}

\definecolor{codebg}{rgb}{0.97,0.97,0.97}
\definecolor{codeframe}{rgb}{0.80,0.80,0.80}
\definecolor{codetext}{rgb}{0.10,0.10,0.10}

\lstdefinestyle{repo}{
  backgroundcolor=\color{codebg},
  frame=single,
  rulecolor=\color{codeframe},
  frameround=ffff,
  framesep=6pt,
  xleftmargin=0.6em,
  xrightmargin=0.6em,
  basicstyle=\ttfamily\small\color{codetext},
  columns=fullflexible,
  keepspaces=true,
  showstringspaces=false,
  breaklines=true,
  breakatwhitespace=false,
  upquote=true,
  tabsize=2,
  aboveskip=0.6em,
  belowskip=0.6em,
}

\begin{document}


\title{GREPO: A Benchmark for Graph Neural Networks on Repository-Level Bug Localization}

\author{Juntong Wang}
\authornote{Equal Contribution}
\email{jtwang25@stu.pku.edu.cn}
\affiliation{%
  \institution{Institute for Artificial Intelligence, Peking University}
  \country{}
}
\affiliation{%
  \institution{School of Intelligence Science and Technology, Peking University}
  \country{}
}

\author{Libin Chen}
\authornotemark[1]
\email{chenlibin@nudt.edu.cn}
\affiliation{%
  \institution{College of Intelligence Science and Technology, National University of Defense Technology}
  \country{}
}

\author{Xiyuan Wang}
\authornotemark[1]
\email{wangxiyuan@pku.edu.cn}
\affiliation{%
  \institution{Institute for Artificial Intelligence, Peking University}
  \country{}
}

\author{Shijia Kang}
\author{Haotong Yang}
\email{kangshijia@stu.pku.edu.cn}
\email{haotongyang@pku.edu.cn}
\affiliation{%
  \institution{Institute for Artificial Intelligence, Peking University}
  \country{}
}

\author{Da Zheng}
\email{zhengda.zheng@antgroup.com}
\affiliation{%
  \institution{Ant Group}
  \country{}
}

\author{Muhan Zhang}
\authornote{correspondence to Muhan Zhang}
\email{muhan@pku.edu.cn}
\affiliation{%
  \institution{Institute for Artificial Intelligence, Peking University}
  \country{}
}

\renewcommand{\shortauthors}{Wang et al.}

\begin{teaserfigure}
  \centering
  \vskip -0.15in
  \includegraphics[width=0.95\textwidth]{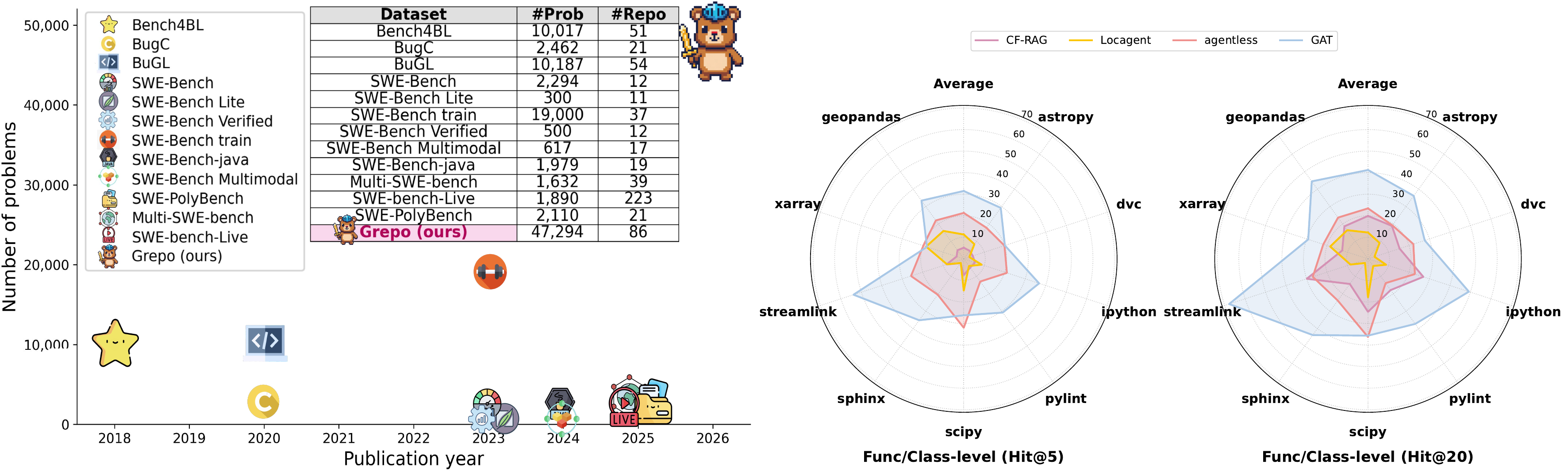}
  \vskip -0.1in
  \caption{ Left: Scale of our dataset compared to prior benchmarks. \textsc{GREPO} covers significantly more issue-linked tasks. Right: Performance on nine representative repositories, demonstrating that the GNN substantially outperforms strong baselines.}
  \label{fig:fist_comb}
  \vskip -0.05in
\end{teaserfigure}

\begin{abstract}
Repository-level bug localization—the task of identifying where code must be modified to fix a bug—is a critical software engineering challenge. Standard Large Language Models (LLMs) are often unsuitable for this task due to context window limitations that prevent them from processing entire code repositories. As a result, various retrieval methods are commonly used, including keyword matching, text similarity, and simple graph-based heuristics such as Breadth-First Search. Graph Neural Networks (GNNs) offer a promising alternative due to their ability to model complex, repository-wide dependencies; however, their application has been hindered by the lack of a dedicated benchmark. To address this gap, we introduce GREPO, the first GNN benchmark for repository-scale bug localization tasks. GREPO comprises 86 Python repositories and 47294 bug-fixing tasks, providing graph-based data structures ready for direct GNN processing. Our evaluation of various GNN architectures shows outstanding performance compared to established information retrieval baselines. This work highlights the potential of GNNs for bug localization and establishes GREPO as a foundational resource for future research. The code is available at \url{https://github.com/qingpingmo/GREPO}.
\end{abstract}



\received{20 February 2007}
\received[revised]{12 March 2009}
\received[accepted]{5 June 2009}

\maketitle
\makeatletter \gdef\@ACM@checkaffil{} \makeatother

\section{Introduction}

Accurate and efficient bug localization within large code repositories is a fundamental challenge in software engineering. Known as repository-level bug localization, this task is a prerequisite for automated program repair, enabling both developers and automated systems to pinpoint the source of a defect and apply an appropriate fix. \citet{bohme2017bug} found that professional developers spend up to 66\% of their debugging time on localization; poor localization often leads to incomplete fixes, the introduction of new bugs, and significantly prolonged development cycles.

Modern software repositories can contain millions of lines of code spread across thousands of files, making it impractical for humans—or even Large Language Models (LLMs)—to inspect the entire codebase directly. Moreover, a bug’s root cause is rarely confined to a single file or function; instead, it often stems from complex, non-local interactions among multiple code entities, necessitating multi-hop reasoning over the repository’s intricate structure.

Prior research has largely framed bug localization as an Information Retrieval (IR) problem, aiming to identify code snippets that align with natural language bug reports~\citep{xia_agentless}. However, effective localization goes beyond simple lexical or semantic matching. It requires multi-hop reasoning and a deep understanding of the issue description and the code’s structural and semantic properties~\citep{chen_locagent}.

Current approaches generally fall into two categories. The first ignores interdependencies within the code repository and evaluates each code entity in isolation, typically by computing the similarity between the bug report and individual code elements. Methods in this category, such as dense vector retrieval, match natural language descriptions to relevant code fragments. While somewhat effective, they are fundamentally limited by their reliance on text similarity and their inability to capture program structure~\citep{lam_dnnloc, liang_flim}.  

The second category attempts to model the repository’s structure but is often constrained by the underlying techniques. LLM-powered agents~\citep{yang_swe-agent_2024, wang_openhands_2024} and graph-based methods~\citep{liu_codexgraph, yu_orcaloca} employ iterative exploration or simple traversal strategies—such as Breadth-First Search (BFS) or Monte Carlo Tree Search (MCTS)—to enable multi-hop reasoning. Although useful, these approaches lack end-to-end learnable mechanisms capable of fully exploiting the rich dependency structure of real-world codebases.

Graph Neural Networks (GNNs) offer a promising alternative. By design, GNNs learn expressive representations of nodes and edges in a graph, making them well-suited to model dependencies among code entities and support the multi-hop reasoning required for repository-level bug localization. Despite this potential, their application to bug localization has been hampered by a critical gap: the \textbf{absence of a dedicated benchmark} that provides ready-to-use graph-structured data for training and evaluating GNNs.

To address this gap and establish a foundation for future GNN research, we introduce GREPO (Graph REPOsitory), the first benchmark specifically designed for GNN-based repository-level bug localization. GREPO comprises \textbf{86 Python repositories} (see the full list in Appendix~\ref{app:reponame}) and includes \textbf{47,294 real-world bug-fixing instances}, offering a diverse and realistic testbed. Crucially, the benchmark provides graph features that can be processed directly by GNNs. As shown in Figure~\ref{fig:fist_comb}, our main contributions are:

\begin{enumerate}[itemsep=2pt,topsep=-2pt,parsep=0pt,leftmargin=10pt]
\item We introduce GREPO, the first repository-level bug localization benchmark tailored for GNNs, providing graph-structured data ready for direct GNN application.
\item Through a scalable collection and preprocessing pipeline, GREPO delivers unprecedented scale: 86 open-source Python repositories and over 47,294 real-world bug-fixing problems. Moreover, we provide a user-friendly toolkit that allows researchers to easily construct GREPO-like datasets from any Python repository.
\item We evaluate a range of representative GNN architectures on GREPO and demonstrate their superior performance against established information retrieval baselines.
\end{enumerate}

\section{Related Work}

\subsection{Bug Localization Benchmarks}

A bug localization benchmark typically consists of a code repository, natural language bug descriptions, and ground-truth labels indicating the location(s) of the fix. Among the most widely used is the dataset introduced by \citet{ye_learning}, which encompasses six open-source Java projects (AspectJ, Birt, Eclipse Platform UI, JDT, SWT, and Tomcat). Other benchmarks also focus primarily on Java~\citep{zhu_cooba,qi_dreamloc,zou_bleser,lee2018bench4bl}. Datasets for other programming languages have been proposed in~\citep{sangle_drast,xia_bugc,muvva2020bugl}.

Beyond dedicated bug localization datasets, recent benchmarks for LLMs and AI agents treat bug localization as an intermediate step in end-to-end issue resolution. RepoBench~\citep{liu2023repobenchbenchmarkinGREPOsitorylevelcode} and SWE-bench~\citep{jimenez2024swebench} are two large-scale benchmarks designed to evaluate LLMs on repository-level code understanding and automated issue fixing, respectively. Subsequent works have extended these efforts to additional languages: \citet{zan2024swebenchjavagithubissueresolving}, \citet{rashid2025swepolybenchmultilanguagebenchmarkrepository}, and \citet{zan2025multiswebenchmultilingualbenchmarkissue} introduce multilingual variants, while \citet{xu2025webbenchllmcodebenchmark} focuses on web development tasks. \citet{yang2025swebenchmultimodel} further incorporates visual elements of software development, such as syntax highlighting and framework-specific UIs. Despite their breadth, all these benchmarks are significantly smaller in scale than GREPO. For a detailed comparison with prior datasets, see Table~\ref{tab:GREPO_vs_prior}.

While numerous bug localization benchmarks exist, none are specifically designed for evaluating Graph Neural Networks (GNNs). Most lack explicit graph-based representations suitable for direct GNN input. GREPO addresses this gap by providing a plug-and-play benchmark with pre-processed, ready-to-use graph structures, enabling seamless training and evaluation of GNN models. Moreover, our design supports straightforward mapping of GNN predictions back to textual code entities (e.g., files or functions), facilitating direct comparison with LLM- and IR-based approaches.

\subsection{Bug Localization Methods}

Early bug localization methods framed the problem as an information retrieval (IR) task, relying on text similarity to match bug reports with relevant code. These approaches commonly employed techniques such as TF-IDF~\citep{lam_dnnloc} or learned text embeddings~\citep{lam_combining, lam_dnnloc} to compute relevance scores. More recently, LLM~\citep{codebert, liang_flim, gunther2023jinaembeddingsnovelset, zhang2024codesage, suresh2024cornstack, li2023towards, SFR-embedding-2} have been used to derive dense vector representations of both bug reports and code for similarity-based ranking.

Recognizing the limitations of purely textual matching, recent work has explored incorporating program structure through graphs such as Control Flow Graphs (CFGs) and Abstract Syntax Trees (ASTs). Models like GraphCodeBERT~\citep{guo_graphcodebert} and UniXCoder~\citep{guo_unixcoder} combine graph structures with code text to learn richer representations. Others, such as CFlow~\citep{zhang_cflow}, leverage code knowledge graphs. While some studies have applied GNNs~\citep{huo_cggnn, ma_icdm}, they encode code pieces only and thus cannot be applied to our repository-level graph.

Recently, the emergence of AI code agents~\citep{yang_swe-agent_2024, wang_openhands_2024} highlights bug localization as a critical intermediate step in autonomous debugging. These agents typically rely on simple heuristics—such as keyword matching and text similarity—combined with file system navigation commands to explore repositories. Although some advanced methods employ hierarchical search strategies~\citep{xia_agentless,swerank} and integrate code graphs with basic traversal mechanisms, including Breadth-First Search (BFS)~\citep{chen_locagent, liu_codexgraph, ouyang_repograph} or Monte Carlo Tree Search (MCTS)~\citep{yu_orcaloca}, they remain largely non-learnable and lack the expressivity to model complex, long-range dependencies across large codebases. Our work addresses this limitation by proposing GNNs as a more powerful, end-to-end learnable alternative to handcrafted traversal strategies. GREPO provides the necessary infrastructure to develop, train, and rigorously evaluate such GNN-based approaches.

\section{Preliminaries}\label{sec:prelim}
\paragraph{Repository-Level Bug Localization}
A bug localization task consists of a code repository, a textual bug description, and ground truth label identifying the localization of the fix at various granularities. In this work, we consider class\&function-level localization.

\paragraph{Message Passing Neural Network (MPNN)~\citep{MPNN}}

MPNN is a popular GNN framework. It consists of multiple message-passing layers, where the $ k $-th layer is:
\begin{equation}
h_v^{(k)} = U^{(k)}(h_v^{(k-1)}, \text{AGG}(\{M^{(k)}(h_u^{(k-1)}) \mid u \in V, (u, v) \in E\})),
\end{equation}
where $ h_v^{(k)} $ is the representation of node $ v $ at the $ k $-th layer, $ U^{(k)} $ and $ M^{(k)} $ are functions such as Multi-Layer Perceptrons (MLPs), and $ \text{AGG} $ is an aggregation function like sum or max. The initial node representation $ h_v^{(0)} $ is the node feature $ X_v $. Each layer aggregates information from neighbors to update the center node's representation.

\section{GREPO Dataset}
\textsc{GREPO} is a large-scale, graph-centric benchmark for \emph{repository-level bug localization}, constructed from real-world GitHub issue reports and their corresponding bug-fixing pull requests (PRs). Designed to support structural reasoning in realistic, multi-file repository settings, GREPO formulates bug localization as a prediction task grounded in concrete \textbf{historical} code states.

Inspired by recent repository-scale benchmarks for code LLMs~\citep{liu2023repobench, jimenez2023swe, yang2024swe, rashid2025swe, zan2025multi}, each instance in GREPO is anchored to a specific repository snapshot: the base commit of a bug-fixing PR serves as the buggy state. The model receives only the issue title and initial description as input—excluding leakage-prone elements such as PR descriptions or developer comments—to isolate the \emph{pre-repair localization} problem. The goal is to predict the set of functions and classes that were actually modified in the fixing PR, which serve as ground-truth labels.

To enable structural and relational reasoning, GREPO represents each repository as a \textbf{heterogeneous} graph. Nodes correspond to code entities, including directories, files, classes, and functions, and edges encode explicit relationships such as containment, function calls, inheritance, and temporal version links across commits. This graph-based representation is central to GREPO's design, making it particularly suitable for Graph Neural Networks and other structure-aware models.

In terms of scale, GREPO comprises \textbf{86 Python repositories} and \textbf{47,294 bug-fixing instances}, significantly surpassing prior bug localization benchmarks in both repository diversity and the number of real-world fixing examples (see Table~\ref{tab:GREPO_vs_prior}). This enables more robust evaluation of generalization.

\begin{figure}[t]
    \centering
    \includegraphics[width=1\linewidth]{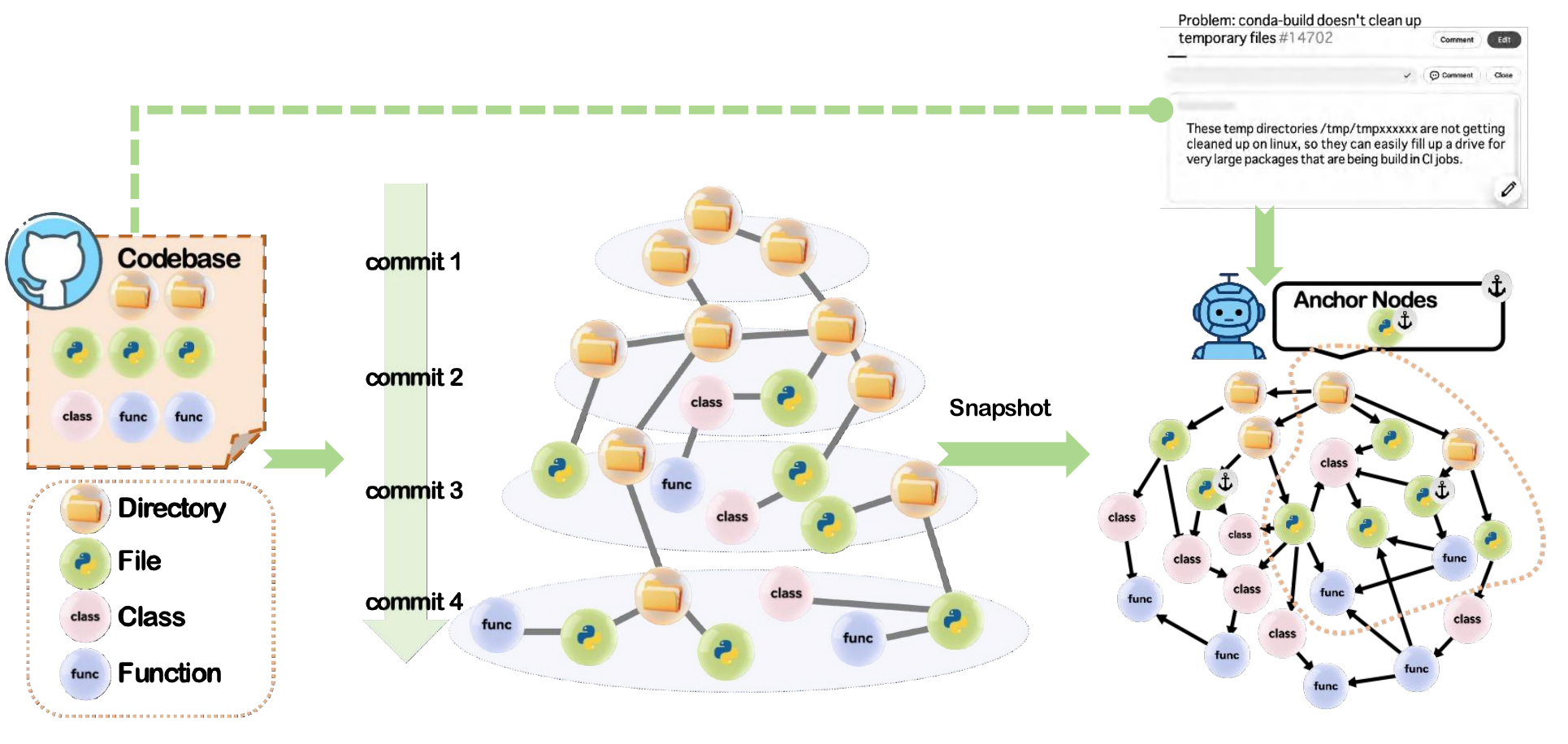}
    \caption{An overview of the dataset construction pipeline consisting of three core stages: (1) converting code repositories into a temporal graph structure with incremental updates, (2) collecting and filtering pull requests and issues to derive bug localization labels, and (3) generating graph features and anchor nodes. For each anchor node, we extract a K-hop subgraph centered at it and run GNN on the subgraph.}
    \label{fig:summary}
    \Description{Our dataset construction pipeline.}
\end{figure}

Figure~\ref{fig:summary} summarizes our dataset construction pipeline, which consists of three main stages:  
\begin{itemize}[itemsep=2pt,topsep=-2pt,parsep=0pt,leftmargin=10pt]
\item Converting repositories into a temporal graph structure where each commit constitutes a graph snapshot (Section~\ref{sec:tempgraph}).
\item Collecting and filtering GitHub issues and PRs to derive high-quality ground-truth labels for the localization task (Section~\ref{sec:label}).  
\item Generating semantic features from source code to enrich graph nodes with textual and syntactic information (Section~\ref{sec:feature}).
\end{itemize}

\begin{table}[t]
\centering
\caption{Comparison of GREPO and existing benchmarks.}
\label{tab:GREPO_vs_prior}
\footnotesize
\setlength{\tabcolsep}{3.5pt}

\rowcolors{2}{blue!6}{white}

\begin{tabularx}{\columnwidth}{@{}p{2.05cm}X p{1.00cm} p{0.72cm} p{0.50cm}@{}}
\toprule
\rowcolor{blue!12}
\textbf{Benchmark} & \textbf{Scale} & \textbf{Lang.} & \textbf{Task}& \textbf{Graph} \\
\midrule
\textbf{Bench4BL}~\citep{lee2018bench4bl} & 51 repos / 10{,}017 tasks & \icoJava & \taskBL  & \icoCross \\
\textbf{BugC}~\citep{niu_survey} & 21 repos / 2{,}462 tasks & \icoC & \taskBL & \icoCross \\
\textbf{BuGL}~\citep{muvva2020bugl} & 54 repos / 10{,}187 tasks & \icoC\,\icoCpp\,\icoJava\,\icoPython & \taskBL  & \icoCross \\
\textbf{RepoBench}~\citep{liu2023repobench} & Multiple repos & \icoJava\,\icoPython & \taskAC  & \icoCross \\
\textbf{SWE-Bench}~\citep{jimenez2023swe} & 12 repos / 2{,}294 tasks & \icoPython & \taskIR  & \icoCross \\
\textbf{SWE-Bench MM}~\citep{yang2024swe} & 17 repos / 617 tasks & \icoJS & \taskIR  & \icoCross \\
\textbf{SWE-PolyBench}~\citep{rashid2025swe} & 21 repos / 2{,}110 tasks & Multi & \taskIR  & \icoCross \\
\textbf{Multi-SWE}~\citep{zan2025multi} &  1{,}632 tasks  & \icoC\,\icoCpp\,\icoJava\,\icoJS\,\icoTS\,\icoRust\,\icoGo & \taskIR  & \icoCross \\
\textbf{WebBench} & -- / 1{,}000 tasks & Web & \taskGEN & \icoCross \\
\textbf{GREPO (Ours)}\,\icoGREPO & \textbf{86 repos / 47{,}294 tasks} & \icoPython & \textbf{\taskBL}  & \icoCheck \\
\bottomrule
\end{tabularx}
\vspace{0.25em}
{\scriptsize
\textbf{Legend:}
\taskBL~bug localization; \taskIR~issue resolving; \taskAC~repo auto-completion; \taskGEN~repo-level generation.
\textbf{Graph:} \icoCheck~provided; \icoCross~not provided.}
\end{table}

\begin{table}[t]
\centering
\scriptsize
\setlength{\tabcolsep}{3pt}
\caption{GREPO task statistics. We report 10\%, 25\%, 50\%, 75\%, 90\%, 99\% quantile. ``Lines changed'':  the number of added plus removed lines;  ``Functions/Classes changed'': the number of changed functions and classes.}
\label{tab:GREPO_stats_overall}

\begin{tabularx}{\linewidth}{l *{6}{>{\raggedleft\arraybackslash}X}}
\toprule
\rowcolor{pink!20}  
Metric & 10\% & 25\% & 50\% & 75\% & 90\% & 99\% \\
\midrule
Issue text length (chars) & 65 & 142 & 376 & 1{,}027 & 2{,}179 & 9{,}092 \\
\#Files changed per fix & 1 & 1 & 2 & 4 & 9 & 40 \\
Lines added & 1 & 5 & 23 & 81 & 231 & 1{,}335 \\
Lines removed & 0 & 1 & 5 & 22 & 80 & 661 \\
Lines changed (add+remove) & 3 & 9 & 33 & 113 & 320 & 1{,}880 \\
\#Functions Changed & 0 & 0 & 2 & 4 & 9 & 47 \\
\#Classes Changed & 0 & 0 & 0 & 0 & 1 & 7 \\

\bottomrule
\end{tabularx}
\end{table}

\begin{figure}[t]
    \centering
    \includegraphics[width=1\linewidth]{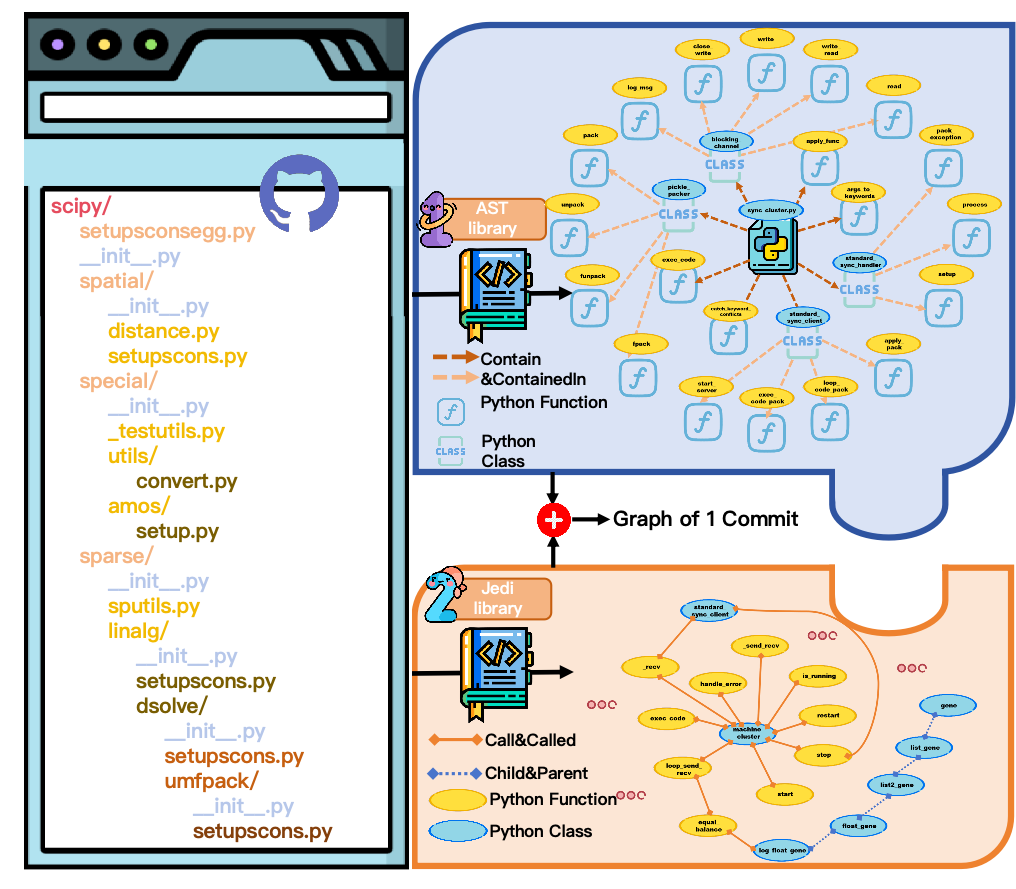}
    \caption{Repository-to-Graph Construction Pipeline for a Single Commit. The left panel shows the repository’s file hierarchy. The upper-right panel illustrates AST extraction using Tree-sitter, which produces containment edges. The lower-right panel depicts inter-procedural relationships, specifically function calls and inheritance (Child/Parent), resolved via static analysis with Jedi.}
    \label{fig:summary2}
    \Description{Graph Construction Pipeline.}
\end{figure}

\begin{figure}[t]
  \centering
  \includegraphics[width=\linewidth]{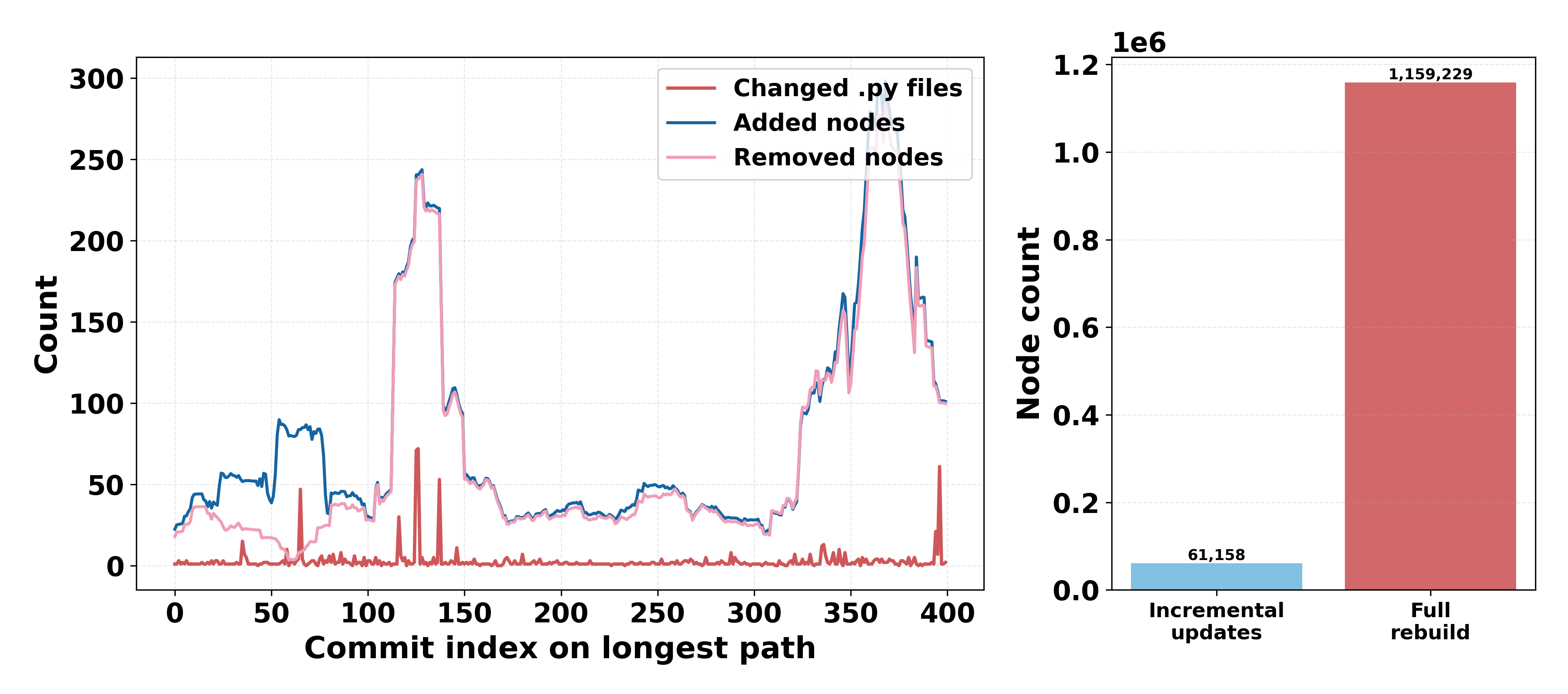}
  \caption{Incremental construction efficiency in SciPy repository. 
  \textbf{Left:} per-commit changed \texttt{.py} files together with incremental node updates (added/removed).
  \textbf{Right:} cumulative incremental update volume (added+removed) versus the sum of per-commit alive nodes. The large gap directly quantifies the gain of incremental construction.}
  \label{fig:incremental-quadrants}  
\Description{Incremental Construction Leads to Better Efficiency than Per-commit Construction.}
\end{figure}

\subsection{Building the Code Repository Graph}\label{sec:tempgraph}

A code repository evolves continuously over time. Since each bug localization task corresponds to a specific historical state---i.e., a particular commit---we model the repository as a sequence of graph snapshots, each representing the code structure at that commit. Constructing a full graph independently for every commit is computationally prohibitive due to the large number of commits and code entities involved. To address this, we build a single temporal graph in which each node is annotated with a \texttt{start\_commit} and \texttt{end\_commit} indicating its lifespan across the commit history.

We initialize the graph using the first commit in the sequence. For subsequent commits, we update the graph incrementally: only code entities modified in the new commit are re-parsed and integrated; unchanged entities are reused from previous snapshots. This design enables efficient, scalable construction by sharing nodes across commits whenever possible.

\subsubsection{Graph Structure for a Single Commit}

For any given commit, we construct a heterogeneous graph that captures both the structural hierarchy and semantic dependencies of the codebase (see Figure~\ref{fig:summary}). Implementation details are in  Appendix~\ref{app:graph-one-commit}.

\paragraph{Nodes.}  
We parse each Python file using the Tree-sitter library~\citep{treesitterlib} to extract its Abstract Syntax Tree (AST). From the AST, we instantiate nodes representing four types of code entities: \texttt{Directory}, \texttt{File}, \texttt{Class}, and \texttt{Function}.

\paragraph{Edges.}  
We define three primary relation types, each accompanied by a reverse edge to facilitate GNNs' message passing:
\begin{enumerate}
    \item \textsc{Contain} / \textsc{ContainedIn}: Encodes hierarchical nesting (e.g., a \texttt{Directory} contains a \texttt{File}; a \texttt{Class} contains a \texttt{Function}). These edges are derived directly from the filesystem layout and AST structure.
    
    \item \textsc{Call} / \textsc{Called}: Represents function or method invocations. We use the Jedi static analysis library~\citep{jedi} to resolve cross-file call relationships.
    
    \item \textsc{Child} / \textsc{Parent}: Models class inheritance (i.e., subclass– superclass relationships), also inferred using Jedi.
\end{enumerate}

Together, these edges form a rich, multi-relational graph that reflects both syntactic containment and semantic dependencies. 

\subsubsection{Incremental Graph Construction}

As illustrated in Figure~\ref{fig:incremental-quadrants}, most commits touch only a small portion of the codebase, and the cumulative incremental update volume (added+removed nodes) is far smaller than the cost of rebuilding full snapshots at each commit. Leveraging this observation, we adopt an incremental construction strategy along a linear commit path.

Each node is created with a \texttt{start\_commit} (and corresponding timestamp); when the entity is deleted or significantly altered in a later commit, its \texttt{end\_commit} is set accordingly. For a new commit, we:
\begin{itemize}[itemsep=2pt,topsep=-2pt,parsep=0pt,leftmargin=10pt]
    \item Extract the list of changed files from the Git patch (\texttt{git diff}).
    \item Re-parse only those files to update their AST-derived nodes and local edges (\textsc{Contain}, \textsc{Call}, \textsc{Child}).
    \item Reuse all unaffected nodes and edges without modification.
\end{itemize}

This approach avoids redundant parsing and dramatically reduces computational overhead, making temporal graph construction feasible at repository scale. Further implementation details are provided in Appendices~\ref{app:incremental-building} and~\ref{app:commit_temp_rel}.

\begin{figure*}[t]
  \centering
  \includegraphics[width=\textwidth]{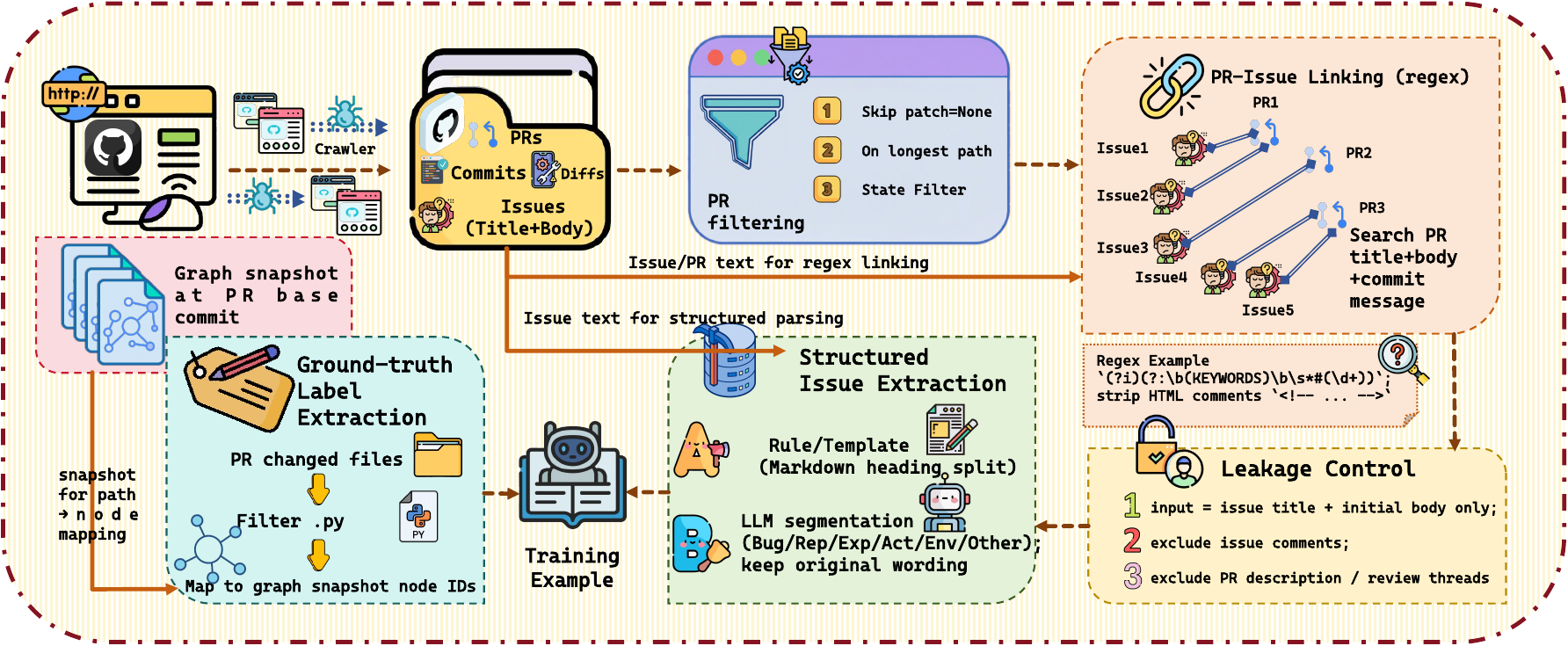}
  \caption{\textbf{Task and label collection schematic.} We link merged PRs to issues via closing-keyword patterns, use only the issue's title/body as model input (excluding comments and PR text), and label each example with the set of modified nodes present in the graph snapshot at the PR base commit.}
  \label{fig:task_label_collection_schematic}
  \Description{Label Collection Pipeline.}
\end{figure*}

\subsection{Label Collection}\label{sec:label}

Our task is to predict the set of code graph nodes that must be modified to fix a bug, given only the natural-language bug report--defined as the GitHub issue title and its initial description (i.e., the body posted at creation time).  

The ground-truth labels consist of Python function and class nodes—specifically, the set of \texttt{Function} and \texttt{Class} nodes corresponding to code entities modified in the bug-fixing pull request (PR). These are mapped to their respective node IDs in the repository's graph snapshot at the PR's base commit (i.e., the buggy state). Figure~\ref{fig:task_label_collection_schematic} illustrates our end-to-end labeling pipeline, which consists of the following five steps:

\begin{enumerate}[itemsep=2pt,topsep=-2pt,parsep=0pt,leftmargin=10pt]
    \item \textbf{Candidate PR collection and filtering.}  
    For each repository, we retrieve PR metadata and associated commits via the GitHub API~\citep{ghapi}, retaining only merged PRs to ensure that the reported issues are meaningful and the fixes have been accepted by maintainers.

    \item \textbf{Issue–PR linking via closing keywords.}  
    We associate PRs with issues using GitHub’s standard closing-keyword convention\footnote{\url{https://docs.github.com/en/issues/tracking-your-work-with-issues/linking-a-pull-request-to-an-issue}}. Specifically, we scan the concatenated text of the PR title, PR description, and all commit messages for patterns such as \texttt{fixes \#123}. Before matching, we strip HTML comments, deduplicate extracted issue IDs, and discard self-references (where the extracted issue number matches the PR number itself).

    \item \textbf{Leakage-safe bug report text.}  
    For each linked issue, we use only the title and the initial body (i.e., the original post) as the model input. We explicitly exclude all subsequent discussion comments, PR descriptions, and review threads---any of which might contain hints or even the solution---to prevent data leakage during training or evaluation.

    \item \textbf{Structured extraction from issue bodies.}  
    To provide clean and comparable inputs across repositories, we convert raw issue bodies into structured text using two complementary strategies:  
    \begin{itemize}
    \item \emph{Rule-based template parsing}: Many repositories use Markdown-based issue templates with headings like \texttt{\#\#\# Steps to reproduce}. We split the body by such headings and map common section titles to canonical slots (e.g., \emph{Bug Description}, \emph{Reproduction Steps}, \emph{Expected Behavior}, \emph{Actual Behavior}, \emph{Environment}). We also extract code blocks, stack traces, and file mentions.  
    \item \emph{LLM-based segmentation}: For repositories without consistent templates, we use a large language model to populate the same canonical slots while preserving the original wording (no paraphrasing). Full details including regular expressions, slot-mapping rules, and exact LLM prompts are provided in Appendix~\ref{app:label_details}.
    \end{itemize}
    \item \textbf{Ground-truth label construction.}  
    Finally, we identify all Python functions and classes modified in the fixing PR (via Git diff analysis) and map them to their corresponding \texttt{Function} and \texttt{Class} node IDs in the graph snapshot at the PR's base commit. These node IDs constitute the ground-truth labels for the localization task.
\end{enumerate}

\begin{figure}[t]
  \centering
  \includegraphics[width=\linewidth]{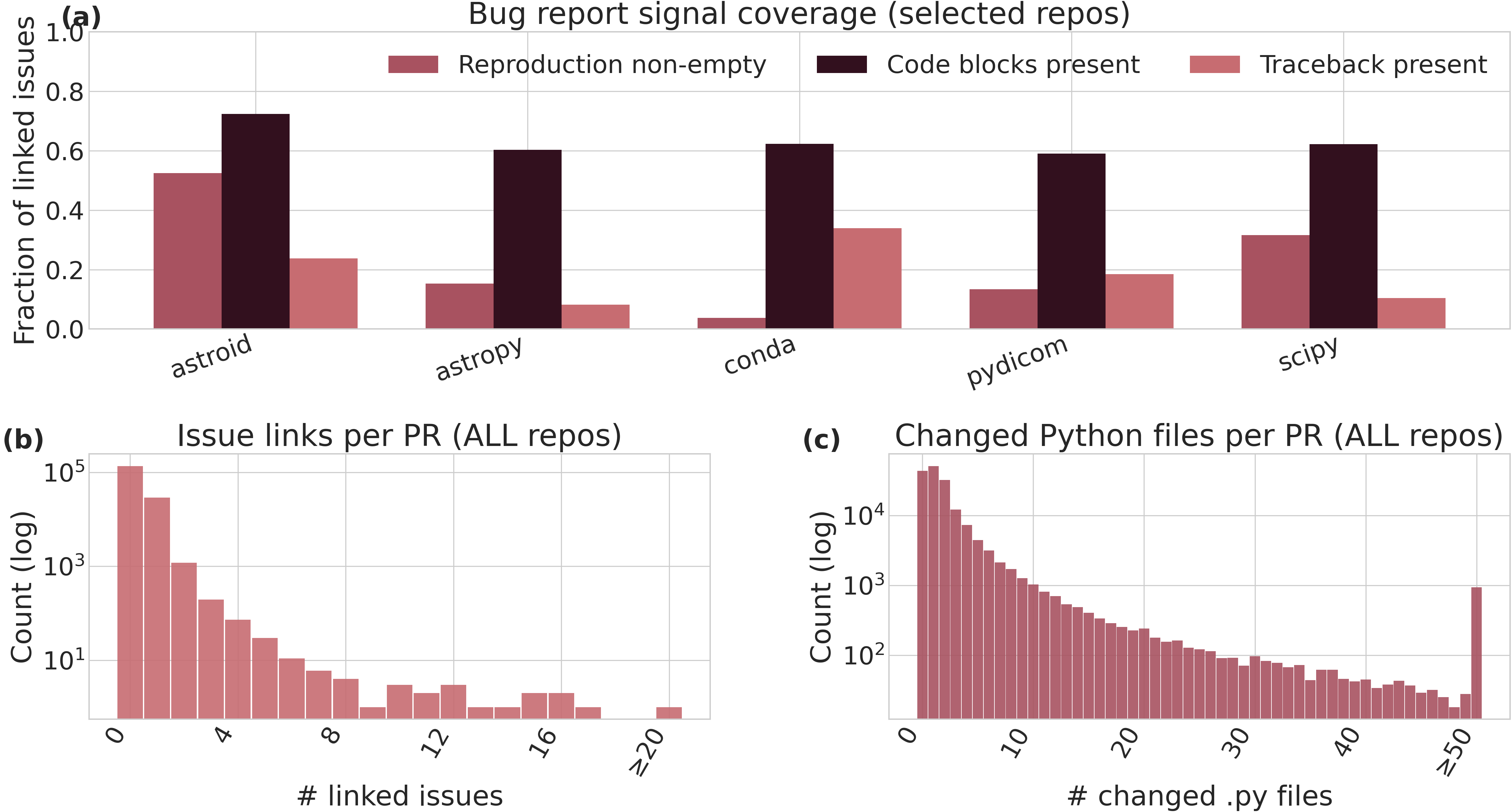}
  \caption{Label Collection statistics. (a) Bug-report signal coverage in linked issues. (b) Number of linked issues per PR. (c) Number of changed \texttt{.py} files per PR. Even under leakage-safe inputs, a non-trivial fraction of linked issues contain strong localization cues. Meanwhile, both the number of issue links per PR and the number of changed Python files per PR are sharply skewed with long tails, precisely motivating repository-level structural reasoning.}
  \Description{Label Statistics.}
  \label{fig:label_stats_main}
\end{figure}

\paragraph{Task Statistics}
Figure~\ref{fig:label_stats_main} summarizes three key dataset characteristics.
Panel~(a) reports how often linked issues contain common debugging signals (reproduction steps, code blocks, tracebacks) for a representative set of repositories. 
Panels~(b--c) report global distributions aggregated across \emph{all} repositories with collected PR data: the number of linked issues per PR and the number of changed Python files per PR (log-scaled y-axis with a tail bin).

\subsection{Graph Feature Construction}\label{sec:feature}

We construct two types of features for repository graphs: node text embeddings and query–node similarity scores. These provide strong, lightweight, and ready-to-use features for downstream graph learning models.

Specifically, we embed both node texts and a small set of rewritten issue queries using the same 4096-dimensional encoder. We then compute query–node similarities via inner product and treat these scores as an auxiliary, retrieval-oriented feature channel. This signal helps guide the model toward relevant code regions even before message passing begins.

\subsubsection{Node Text Embedding}\label{sec:node-embed}

For each node $v$ , we serialize its textual content into a single string and augment it with its file path as supplementary context. This yields a uniform textual representation across heterogeneous node types (e.g., files, classes, functions). We encode this string into a dense embedding  $\mathbf{h}_v \in \mathbb{R}^{4096}$  using the Qwen3-Embedding-8B model~\citep{zhang2025qwen3}. To improve efficiency at scale, we deploy the embedding model via the vLLM serving framework~\citep{kwon2023vllm}.

\subsubsection{Query–Node Similarity}\label{sec:sim-feature}

For each issue, we generate a small set of rewritten queries (default:  $m = 5$) and embed them using the same encoder, yielding query embeddings  $\mathbf{h}_{q_1}, \dots, \mathbf{h}_{q_m} \in \mathbb{R}^{4096}$. The similarity between a node $v$  and a query  $q$  is computed as the inner product:
\begin{equation}
\mathrm{sim}(v, q) = \mathbf{h}_v^\top \mathbf{h}_q.
\end{equation}
This similarity score serves as a low-dimensional, transferable feature that provides an initial retrieval signal across repositories---prior to any graph propagation. Empirical validation of its effectiveness is provided in Appendix~\ref{app:featval}.

\subsubsection{Anchor Nodes}

Full repository graphs at a given snapshot can be too large to fit in GPU memory. Moreover, they often contain many nodes that are irrelevant to the current issue; including all of them during GNN inference can dilute the signal and reduce the hit rate among top-ranked candidates. To address this, we restrict the input graph to the $k$-hop subgraph centered on a small set of anchor nodes.

We use two complementary strategies to select anchor nodes:

\begin{itemize}[itemsep=2pt,topsep=-2pt,parsep=0pt,leftmargin=10pt]
    \item \textbf{Semantic-based anchor nodes.}  
    Inspired by Code Graph Models (CGM)~\citep{tao2025code}, we use an LLM-based Rewriter that, given an issue report, generates up to five search-style queries along with structured code entities and keywords. These signals are used to retrieve candidate anchor nodes via lexical and semantic matching, providing high-quality entry points for downstream retrieval and graph reasoning. Full details of the method are in Appendix~\ref{app:method-details}; the exact prompt templates and output schemas for the Rewriter are in Appendix~\ref{app:prompts} (see Figure~\ref{fig:extractor-inferer-prompts}); and representative Rewriter outputs are shown in Appendix~\ref{app:rewrite-examples} (Figure~\ref{fig:rewrite-examples}).

    \item \textbf{Temporal anchor nodes.}  
    Intuitively, recently modified modules are more likely to be edited again, and code entities co-edited within a short time window often appear together in future fixes. These patterns are captured in \textbf{commit history} rather than in the static code snapshot. We train a temporal retriever on the training set using only past commit history (i.e., no future leakage) to estimate a prior probability over nodes. At inference time, this retriever selects additional anchor candidates from the test repositories. We experimented with several representative temporal GNNs for this purpose; implementation details are in Appendix~\ref{app:Tanchor}.
\end{itemize}

\begin{table*}[t]
    \caption{Comparison of LLM and GNN methods on the GREPO dataset. The highest value in each column is highlighted in \best{orange}. "Avg. Rank" denotes the average ranking of the method across all metrics based on average performance.}
    \label{tab:main_table}
    \centering
    \begin{tabular}{lc cllllllllll}
    \toprule
    \multirow{2}{*}{Method} & \multirow{2}{*}{\shortstack{Avg.\\Rank}} & \multirow{2}{*}{Metric} & \multirow{2}{*}{Average} & \multicolumn{9}{c}{Datasets} \\
    \cmidrule(lr){5-13}
    & & & & astropy & dvc & ipython & pylint & scipy & sphinx & streamlink & xarray & geopandas \\ 
    \midrule
    
    \multirow{4}{*}{CF-RAG} & \multirow{4}{*}{9.75} 
      & Hit@1  & 0.70  & 0.88 & 0.16 & 1.70 & 0.03 & 0.83 & 0.21 & 0.54 & 0.56 & 1.37 \\ 
    & & Hit@5  & 5.16  & 4.64 & 4.42 & 5.48 & 5.03 & 7.54 & 2.46 & 8.39 & 3.51 & 4.95 \\ 
    & & Hit@10 & 10.85 & 11.13 & 8.64 & 13.29 & 8.70 & 12.97 & 8.17 & 17.90 & 6.86 & 9.99 \\ 
    & & Hit@20 & 19.93 & 18.90 & 15.43 & 27.01 & 17.90 & 24.59 & 14.40 & 29.92 & 12.60 & 18.65 \\ 
    \midrule

    \multirow{4}{*}{LocAgent} & \multirow{4}{*}{9.25} 
      & Hit@1  & 5.04  & 5.09 & 0.16 & 1.02 & 3.76 & 5.23 & 0.39 & 3.42 & 10.55 & 3.85 \\ 
    & & Hit@5  & 11.30 & 8.42 & 2.73 & 8.75 & 4.08 & 14.76 & 2.36 & 8.20 & 18.39 & 16.02 \\ 
    & & Hit@10 & 12.01 & 9.15 & 3.16 & 8.84 & 4.08 & 16.88 & 2.36 & 8.57 & 18.48 & 16.44 \\ 
    & & Hit@20 & 12.23 & 9.15 & 3.16 & 8.84 & 4.08 & 17.89 & 2.36 & 8.57 & 18.48 & 16.44 \\  
    \midrule

    \multirow{4}{*}{Agentless} & \multirow{4}{*}{8.00} 
      & Hit@1  & 13.65 & 10.05 & \best{11.71} & 13.44 & 9.82 & \best{23.73} & 13.14 & 12.68 & \best{12.21} & \best{16.09} \\ 
    & & Hit@5  & 21.32 & 17.86 & 19.88 & 21.08 & 13.01 & \best{31.98} & 20.56 & 25.76 & \best{19.74} & 22.05 \\ 
    & & Hit@10 & 22.62 & 18.82 & 21.91 & 21.98 & 13.41 & \best{34.95} & 21.01 & 26.97 & 21.26 & 23.28 \\ 
    & & Hit@20 & 23.43 & 19.24 & 22.15 & 22.89 & 13.92 & \best{36.27} & 23.91 & 26.97 & 21.84 & 23.68 \\ 
    \midrule

    \multirow{4}{*}{GCN} & \multirow{4}{*}{6.25} 
      & Hit@1  & 13.74 & 12.32 & 8.91 & 22.00 & 11.92 & 13.81 & 17.00 & 19.00 & 6.77 & 11.90 \\ 
    & & Hit@5  & 30.18 & 27.27 & 18.70 & 38.23 & 28.50 & 26.63 & 32.67 & 50.53 & 17.40 & 31.65 \\ 
    & & Hit@10 & 35.52 & 30.64 & 23.81 & 43.56 & 32.54 & 30.94 & 37.18 & 62.07 & 22.64 & 36.30 \\ 
    & & Hit@20 & 39.24 & 33.37 & 25.91 & 49.25 & 34.91 & 33.78 & 40.19 & 67.50 & 26.38 & 41.88 \\ 
    \midrule

    \multirow{4}{*}{SAGE} & \multirow{4}{*}{4.75} 
      & Hit@1  & 13.68 & 12.56 & 9.25 & 18.65 & 13.71 & 12.44 & 17.84 & 18.61 & 7.40 & 12.70 \\ 
    & & Hit@5  & 31.39 & 28.06 & 20.12 & 37.82 & 31.06 & 26.82 & 35.29 & 53.30 & 18.41 & 31.64 \\ 
    & & Hit@10 & 37.04 & 32.22 & 24.91 & 45.36 & 34.43 & 32.22 & 40.42 & 62.78 & 23.83 & 37.16 \\ 
    & & Hit@20 & 40.86 & 36.31 & 28.64 & 48.70 & 36.60 & 36.42 & 42.84 & 68.69 & 28.15 & 41.39 \\ 
    \midrule

    \multirow{4}{*}{GIN} & \multirow{4}{*}{4.00} 
      & Hit@1  & 14.26 & 12.25 & 9.16 & 22.12 & 12.78 & 12.75 & 17.63 & 20.47 & 8.27 & 12.89 \\ 
    & & Hit@5  & 31.48 & 28.80 & 20.17 & 39.48 & 29.63 & 27.43 & 35.96 & 53.23 & 17.74 & 30.89 \\ 
    & & Hit@10 & 36.99 & 32.38 & 25.31 & 45.10 & 34.08 & 31.93 & 40.53 & 62.95 & 23.21 & 37.41 \\ 
    & & Hit@20 & 40.87 & 35.16 & 28.93 & 48.99 & 36.37 & 35.91 & 43.11 & 69.19 & 28.29 & 41.90 \\ 
    \midrule

    \multirow{4}{*}{GatedGCN} & \multirow{4}{*}{4.00} 
      & Hit@1  & 14.64 & 12.81 & 9.67 & \best{22.61} & 13.94 & 14.15 & 17.60 & 20.65 & 6.71 & 13.58 \\ 
    & & Hit@5  & 31.49 & 27.75 & 19.42 & \best{39.53} & 30.61 & 27.75 & 34.23 & 52.77 & 17.78 & 33.58 \\ 
    & & Hit@10 & 36.47 & 31.15 & 23.93 & \best{45.20} & 33.73 & 31.98 & 38.77 & 62.40 & 22.51 & \best{38.59} \\ 
    & & Hit@20 & 39.90 & 33.95 & 26.91 & 49.25 & 35.46 & 34.84 & 41.25 & 68.62 & 25.88 & 42.94 \\ 
    \midrule

    \multirow{4}{*}{GATv2} & \multirow{4}{*}{\best{1.00}} 
      & Hit@1  & \best{14.84} & 13.46 & 9.21 & 21.09 & 13.06 & 15.18 & 19.36 & 20.47 & 7.88 & 13.81 \\ 
    & & Hit@5  & \best{32.47} & 28.53 & \best{20.70} & 39.05 & \best{31.09} & 28.74 & \best{35.65} & \best{54.49} & 19.36 & 34.60 \\ 
    & & Hit@10 & \best{37.68} & \best{33.22} & \best{26.11} & \best{45.20} & 34.97 & 32.68 & \best{40.87} & 63.15 & \best{24.50} & 38.44 \\ 
    & & Hit@20 & \best{41.54} & \best{36.29} & \best{29.13} & 49.06 & \best{37.68} & 36.78 & 43.31 & \best{69.30} & 28.90 & 43.40 \\ 
    \midrule

    \multirow{4}{*}{GPS} & \multirow{4}{*}{6.00} 
      & Hit@1  & 14.32 & 12.80 & 9.54 & 20.36 & 14.00 & 12.84 & 18.23 & \best{21.44} & 7.12 & 12.54 \\ 
    & & Hit@5  & 30.44 & 26.38 & 19.98 & 37.43 & 30.42 & 26.32 & 34.14 & 50.34 & 17.59 & 31.32 \\ 
    & & Hit@10 & 35.30 & 30.51 & 24.01 & 42.56 & 34.01 & 30.32 & 37.55 & 60.99 & 22.38 & 35.33 \\ 
    & & Hit@20 & 38.50 & 32.86 & 27.01 & 45.19 & 36.86 & 32.83 & 40.24 & 66.61 & 25.89 & 38.97 \\ 
    \midrule

    \multirow{4}{*}{GAT} & \multirow{4}{*}{2.00} 
      & Hit@1  & 14.80 & \best{13.82} & 8.97 & 19.22 & \best{14.36} & 14.29 & \best{19.37} & 20.87 & 7.91 & 14.42 \\ 
    & & Hit@5  & 31.51 & \best{29.24} & 20.09 & 36.80 & 30.81 & 26.20 & 35.22 & 53.85 & 17.98 & 33.42 \\ 
    & & Hit@10 & 37.40 & 32.91 & 25.80 & 44.27 & \best{35.30} & 31.48 & 39.66 & \best{64.34} & 24.33 & 38.54 \\ 
    & & Hit@20 & 41.25 & 36.15 & 27.70 & \best{49.26} & 37.32 & 35.64 & \best{43.71} & 67.82 & \best{29.20} & \best{44.46} \\ 
    \bottomrule
    \end{tabular}
\end{table*}

\begin{table*}[t]
    \centering
\caption{Ablation study on the feature, edge, and anchor node provided in GREPO dataset.}\label{tab:abl}
    \begin{tabular}{cccccccccc}
    \toprule
    ~&Full&\multicolumn{3}{|c}{Feature Ablation}&\multicolumn{3}{|c}{Edge Ablation}&\multicolumn{2}{|c}{Anchor Ablation} \\
    \midrule
    ~ & GAT & woSim & woAnchor & woET & woContain & woCall & woInherit  & woSemantic & woTemporal \\ 
    \midrule
        Hit@1 & 14.80  & 13.43  & 13.61  & 13.97  & 14.12  & 14.61  & 14.07    & 14.31  & 8.09  \\ 
        Hit@5 & 31.51  & 30.16  & 30.49  & 31.21  & 30.56  & 32.01  & 30.32    & 31.47   & 19.13  \\ 
        Hit@10 & 37.40  & 36.57  & 36.15  & 36.29  & 34.87  & 37.15  & 35.94  & 36.40  & 24.18  \\ 
        Hit@20 & 41.25  & 40.67  & 40.35  & 40.04  & 37.47  & 40.80  & 40.30  & 38.85  & 28.78 \\ 
    \bottomrule  
    \end{tabular}
\end{table*}

\section{Experiments}\label{sec:experiments-corechange5}

\subsection{Experimental Settings}\label{sec:metrics-corechange5}

For each issue $q$ , let  $\mathcal{G}(q)$  denote the ground-truth set of function and class nodes that must be modified to fix the bug, and let  $\mathrm{TopK}(q)$  be the top-$K$  nodes predicted by the model. We evaluate performance using mean query recall, defined as:

\begin{equation}
\mathrm{Hit@K}(q) = \frac{|\mathcal{G}(q) \cap \mathrm{TopK}(q)|}{|\mathcal{G}(q)|}.
\end{equation}
The score reported is the average of all test issues.

We split the issues in each repository chronologically into training (80\%) and test (20\%) sets based on issue creation time. Models are trained on the \textbf{combined training sets} from all 86 repositories. However, for evaluation, we report results only on the test sets of nine representative repositories:  
\texttt{astropy}, \texttt{dvc}, \texttt{ipython}, \texttt{pylint}, \texttt{scipy}, \texttt{sphinx}, \texttt{streamlink}, \texttt{xarray}, and \texttt{geopandas}.  

To assess generalization, we also evaluate in a 0-shot setting, where these 9 evaluation repositories are entirely excluded from training.

All methods--both GNNs and baselines--use the same issue splits. Critically, $\mathrm{Hit@K}$ is computed against the full global ground-truth set  $\mathcal{G}(q)$  (as defined in Section~\ref{sec:label}). If any ground-truth node is missing from the extracted subgraph (e.g., due to anchor selection), it cannot be recovered, and the corresponding hit score is zero. This ensures that our evaluation metric is fair to LLM-based baselines. Full experimental details are in Appendix~\ref{app:exp}.

\subsection{Benchmarking GNNs on GREPO}
To comprehensively evaluate GNNs, we compare it against several representative information retrieval baselines:
\begin{enumerate}[itemsep=2pt,topsep=-2pt,parsep=0pt,leftmargin=15pt,rightmargin=5pt]
\item \textbf{LocAgent}~\citep{chen_locagent}, an agent-based method that leverages the Qwen2.5-72B-Instruct model~\citep{yang2025qwen2}.
\item \textbf{Agentless}~\citep{xia_agentless}, which uses GPT-4o~\citep{hurst2024gpt} for direct code retrieval without agent orchestration.
\item \textbf{CF-RAG}, the retrieval-augmented generation system from
CodeFuse~\citep{tao2025code}, which employs Qwen3-Embedding-8B~\citep{zhang2025qwen3} for semantic matching.
\end{enumerate}
Notably, CF-RAG also serves as the anchor node generator in our GNN pipeline. This design choice ensures a fair comparison: any performance gain from our GNN can be attributed to the graph learning component itself, rather than improvements in initial retrieval quality.
All LLM baselines are evaluated without additional training, applied directly to the test sets of each repository using the same input constraints.
For GNNs, we benchmark several established architectures: GCN~\citep{GCN}, GIN~\citep{GIN}, GraphSAGE~\citep{GraphSage}, GAT~\citep{GAT}, GATv2~\citep{GAT}, and the graph transformer GPS~\citep{GT-GPS}.
The results, shown in Table~\ref{tab:main_table}, reveal two key insights:
(1) All GNN variants consistently outperform LLM-based baselines, demonstrating the effectiveness of graph-structured reasoning for bug localization.
(2) Architectural choices matter: GNNs with attention mechanisms--such as GAT, GATv2--generally achieve higher performance than simpler models like GCN, GIN, or GraphSAGE, highlighting the importance of expressive message-passing designs for this task.

\begin{figure}[t]
\includegraphics[width=0.4\textwidth]{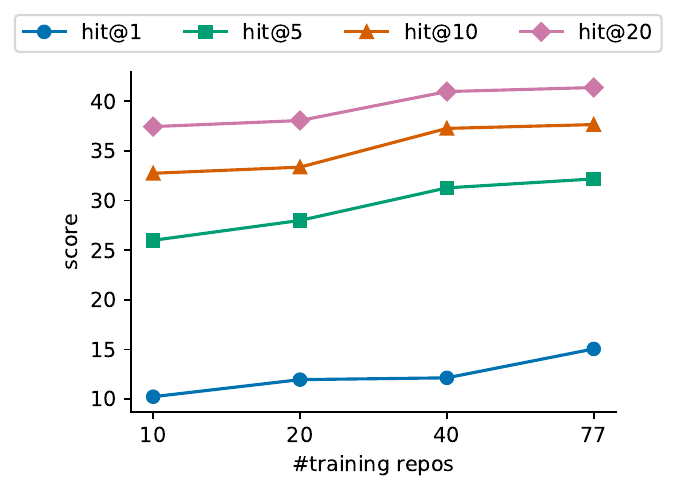}
    \caption{Scaling Law of GAT in GREPO. The training repositories for each scale are selected randomly.}\label{fig:scalinglawGAT}
    \Description{Scaling law on our GREPO dataset.}
\end{figure}
\subsection{Scaling Law}

Given the unprecedented scale of GREPO, we investigate whether it can support the training of a bug localization foundation model. To this end, we study the scaling behavior of GAT under a zero-shot setting, where the training set excludes all test repositories.

As shown in Figure~\ref{fig:scalinglawGAT}, the average Hit@K performance across the 9 held-out test repositories improves steadily as the number of training repositories increases. Notably, when trained on 77 repositories, GAT achieves performance comparable to that of fully supervised models on 86 training repositories (i.e., those trained on the same repositories as the test set). This demonstrates that GAT trained on GREPO reveals strong zero-shot generalization capability and suggests that GREPO is sufficiently large and diverse to support scalable, transferable bug localization models.

\subsection{Ablation Study}

To validate the contribution of our graph structure and node feature design, we conduct a comprehensive ablation study on GAT. In Table~\ref{tab:abl}, "Full" denotes the complete GAT model with all proposed components. We evaluate the following ablations:

\begin{itemize}[itemsep=2pt,topsep=-2pt,parsep=0pt,leftmargin=15pt,rightmargin=5pt]
    \item \textbf{Feature Ablations:} \texttt{woSim}: removes the query–node similarity feature derived from text embeddings.  \texttt{woAnchor}: disables anchor node labeling (i.e., treats all nodes in the subgraph equally).  \texttt{woET}: removes edge type embeddings, treating all edge types identically during message passing.

    \item \textbf{Edge Ablations:}  \texttt{woContain}, \texttt{woCall}, \texttt{woInherit}: remove edges of type \textsc{Contain}, \textsc{Call}, and \textsc{Inherit}, respectively, from the input graph.

    \item \textbf{Anchor Ablations:}  \texttt{woSemantic}: excludes semantic anchor nodes.  \texttt{woTemporal}: excludes temporal anchor nodes.
\end{itemize}

All ablated variants result in performance degradation, confirming that each component--features, edge types, and anchor selection strategies--contributes meaningfully to the model's effectiveness.

\section{Conclusion}
We present GREPO, the first benchmark designed to evaluate GNNs for repository-level bug localization. GREPO provides ready-to-use heterogeneous, temporally indexed repository graphs, plus a scalable pipeline for incremental construction and retrieval-guided subgraph inference. Across 86 Python repositories and 47,294 real-world problems, diverse GNNs consistently outperform strong IR/LLM baselines, and ablations confirm the importance of relation types and temporal signals. We release the benchmark and tooling to accelerate reproducible progress on structure-aware debugging and future extensions to broader languages and analyzes.

\bibliographystyle{ACM-Reference-Format}
\bibliography{sample-base}

\newpage
\appendix
\onecolumn

\section{Graph of One Commit (AST \& Jedi Details)}
\label{app:graph-one-commit}

This appendix provides implementation-level details for constructing the \emph{graph of one commit} used in our method, with special emphasis on the \textbf{exact input/output interface} of (i) Tree-sitter AST parsing and (ii) Jedi static inference. 

\subsection{Node schema and Edge schema (heterogeneous relations)}
\label{app:node-schema}

We construct a typed node set that jointly models filesystem structure and code-level definitions. Each node has:
\begin{itemize}
    \item \textbf{Node type} in \{directory, file, python\_file, class\_def, func\_def\}.
    \item \textbf{Address} by its \texttt{path} (repository-relative path) and optional \texttt{name} (qualified definition name).
    \item \textbf{Text attribute} \texttt{attr[code]} storing the raw source (full file for \texttt{python\_file}, definition text for \texttt{class\_def}/\texttt{func\_def}).
    \item \textbf{Temporal fields} \texttt{start\_commit} and \texttt{end\_commit}, plus a \texttt{previous} list for version linking (\S\ref{app:temporal}).
\end{itemize}

\paragraph{Definition join key.}
Jedi does \emph{not} know our graph node IDs; it only returns semantic targets as \texttt{(module\_path, line)} pairs (plus metadata). Conversely, Tree-sitter emits graph nodes and their start localizations, but does not robustly resolve dynamic Python name binding. The join key provides a minimal, stable interface shared by both stages. Tree-sitter and Jedi produce \emph{complementary} information: Tree-sitter reliably identifies \textbf{where} definitions/calls occur in source, while Jedi can often infer \textbf{what} symbol a given source position refers to. To connect Jedi-resolved targets back to our extracted definition nodes, we build an explicit join index (a lookup table) over definitions, using the key:
\begin{equation}
(\texttt{relpath(module\_path)}, \texttt{definition.line}) \mapsto \texttt{def\_node\_id}.
\end{equation}
This join key is the core bridge between Tree-sitter and Jedi.

\paragraph{What the join key means.}
\begin{itemize}
    \item \textbf{\texttt{module\_path}} is returned by Jedi for each inferred candidate and indicates the file where the candidate symbol is defined (an absolute path).
    \item \textbf{\texttt{relpath(module\_path)}} converts that absolute path to a repository-relative path (the same convention used when indexing Tree-sitter-extracted definitions).
    \item \textbf{\texttt{definition.line}} is the line number (1-indexed) where Jedi believes the inferred definition starts (e.g., the \texttt{class} or \texttt{def} line).
    \item \textbf{\texttt{def\_node\_id}} is the unique node identifier assigned during Tree-sitter parsing when we created the corresponding \texttt{class\_def} or \texttt{func\_def} node.
\end{itemize}

We define a two-stage process to construct the edges, and the detailed edge-construction procedure is shown in Algorithm~\ref{alg:join-driver}, where the called \textbf{ExtractQueries} is presented as Algorithm~\ref{alg:extract-queries}, and the called \textbf{InferAndJoin} is presented as Algorithm~\ref{alg:infer-join}.

\begin{algorithm}[t]
\caption{AST--Jedi Join for Call and Superclass Edge Construction (Driver)}
\label{alg:join-driver}
\begin{algorithmic}[1]
\Require Python source files $\mathcal{F}$ under repository root $\mathcal{R}$
\Ensure Definition nodes $V$ and directed edges $E_{\textsc{call}}$ and $E_{\textsc{sup}}$

\Statex \textbf{Data structures}
\State $V \gets \emptyset$
\State $E_{\textsc{call}} \gets \emptyset$, \;\; $E_{\textsc{sup}} \gets \emptyset$
\State $\texttt{line2def} \gets$ empty map \Comment{$(\textit{file},\textit{start\_line}) \mapsto$ def node}
\State $\texttt{CallSites} \gets$ empty map \Comment{def node $\mapsto$ list of $(\ell,c)$}
\State $\texttt{SuperTokens} \gets$ empty map \Comment{class node $\mapsto$ list of $(\ell,c)$}

\Statex
\State $(V,\texttt{line2def},\texttt{CallSites},\texttt{SuperTokens}) \gets \textsc{ExtractQueries}(\mathcal{F})$
\State $(E_{\textsc{call}},E_{\textsc{sup}}) \gets \textsc{InferAndJoin}(\mathcal{F},V,\texttt{line2def},\texttt{CallSites},\texttt{SuperTokens})$
\State \Return $(V, E_{\textsc{call}}, E_{\textsc{sup}})$
\end{algorithmic}
\end{algorithm}

\begin{algorithm}[t]
\caption{\textsc{ExtractQueries}: Tree-sitter Extraction of Definitions and Query Points}
\label{alg:extract-queries}
\begin{algorithmic}[1]
\Require Files $\mathcal{F}$
\Ensure $V$, $\texttt{line2def}$, $\texttt{CallSites}$, $\texttt{SuperTokens}$

\State $V \gets \emptyset$
\State $\texttt{line2def} \gets$ empty map
\State $\texttt{CallSites} \gets$ empty map
\State $\texttt{SuperTokens} \gets$ empty map

\ForAll{$f \in \mathcal{F}$}
  \State $T \gets \textsc{TreeSitterParse}(f)$

  \Comment{Register definition nodes}
  \ForAll{$d \in \textsc{DefNodes}(T)$} \Comment{\texttt{class\_definition} or \texttt{function\_definition}}
    \State $v \gets \textsc{CreateDefNode}(d)$
    \State $V \gets V \cup \{v\}$
    \State $(\ell_d,c_d) \gets \textsc{StartPoint}(d)$
    \State $\texttt{line2def}\big[(\textsc{NormPath}(f),\ell_d)\big] \gets v$
    \State $\texttt{CallSites}[v] \gets [\,]$
    \If{$d$ is a \texttt{class\_definition}}
      \State $\texttt{SuperTokens}[v] \gets [\,]$
      \ForAll{$s \in \textsc{SuperclassTokens}(d)$}
        \State $(\ell_s,c_s) \gets \textsc{StartPoint}(s)$
        \State append $(\ell_s,c_s)$ to $\texttt{SuperTokens}[v]$
      \EndFor
    \EndIf
  \EndFor

  \Comment{Collect call sites per enclosing definition}
  \ForAll{$c \in \textsc{CallNodes}(T)$} \Comment{\texttt{call} nodes}
    \State $d \gets \textsc{EnclosingDef}(c)$ \Comment{innermost enclosing class/func definition}
    \If{$d \neq \emptyset$}
      \State $(\ell_c,c_c) \gets \textsc{StartPoint}(c)$
      \State $\ell_d \gets \textsc{StartPoint}(d).\ell$
      \State $v \gets \texttt{line2def}\big[(\textsc{NormPath}(f),\ell_d)\big]$
      \State append $(\ell_c,c_c)$ to $\texttt{CallSites}[v]$
    \EndIf
  \EndFor
\EndFor

\State \Return $(V,\texttt{line2def},\texttt{CallSites},\texttt{SuperTokens})$
\end{algorithmic}
\end{algorithm}

\begin{algorithm}[t]
\caption{\textsc{InferAndJoin}: Jedi Inference, Join, and Edge Materialization}
\label{alg:infer-join}
\begin{algorithmic}[1]
\Require Files $\mathcal{F}$; nodes $V$; maps $\texttt{line2def}$, $\texttt{CallSites}$, $\texttt{SuperTokens}$
\Ensure Edges $E_{\textsc{call}}$, $E_{\textsc{sup}}$

\State $E_{\textsc{call}} \gets \emptyset$, \;\; $E_{\textsc{sup}} \gets \emptyset$

\ForAll{$f \in \mathcal{F}$}
  \State $\texttt{script} \gets \textsc{JediScript}(\text{path}=f)$
  \ForAll{$v \in V$ such that $\textsc{FileOf}(v)=f$}

    \Comment{(a) Call edges}
    \ForAll{$(\ell,c) \in \texttt{CallSites}[v]$}
      \State $\mathcal{P} \gets \textsc{Infer}(\texttt{script},\ell,c)$
      \ForAll{$p \in \mathcal{P}$}
        \If{$\textsc{ModulePath}(p)\neq\emptyset$ \textbf{and} $\textsc{DefLine}(p)\neq\emptyset$}
          \State $k \gets \big(\textsc{NormPath}(\textsc{ModulePath}(p)),\textsc{DefLine}(p)\big)$
          \If{$k \in \texttt{line2def}$}
            \State $u \gets \texttt{line2def}[k]$
            \State $E_{\textsc{call}} \gets E_{\textsc{call}} \cup \{(v,u)\}$
          \EndIf
        \EndIf
      \EndFor
    \EndFor

    \Comment{(b) Superclass edges (subclass $\rightarrow$ superclass)}
    \If{$v$ is a class\_def node}
      \ForAll{$(\ell,c) \in \texttt{SuperTokens}[v]$}
        \State $\mathcal{P} \gets \textsc{Infer}(\texttt{script},\ell,c)$
        \ForAll{$p \in \mathcal{P}$}
          \If{$\textsc{ModulePath}(p)\neq\emptyset$ \textbf{and} $\textsc{DefLine}(p)\neq\emptyset$}
            \State $k \gets \big(\textsc{NormPath}(\textsc{ModulePath}(p)),\textsc{DefLine}(p)\big)$
            \If{$k \in \texttt{line2def}$}
              \State $u \gets \texttt{line2def}[k]$
              \State $E_{\textsc{sup}} \gets E_{\textsc{sup}} \cup \{(v,u)\}$
            \EndIf
          \EndIf
        \EndFor
      \EndFor
    \EndIf

  \EndFor
\EndFor

\State \Return $(E_{\textsc{call}},E_{\textsc{sup}})$
\end{algorithmic}
\end{algorithm}

\paragraph{Important corner cases (observable in our real dumps).}
\begin{itemize}
    \item \textbf{Builtins / external libraries.} Jedi may infer builtins (e.g., \texttt{super}, \texttt{Exception}) whose \texttt{module\_path} points to a typeshed file, not the analyzed repository. These candidates cannot be joined to our repository definition nodes, so \texttt{joined\_targets} is empty and no graph edge is added.
    \item \textbf{Ambiguity.} If Jedi returns multiple joinable candidates for a query point, we conservatively add edges to all joinable targets.
\end{itemize}

For detailed information about the glossary of output fields, see \cref{tab:output_fields}.

\begin{table*}[t]
\caption{Output Fields of the AST Processing Pipeline}
\label{tab:output_fields}

\rowcolors{2}{gray!6}{white}  
\begin{tabularx}{\textwidth}{p{2.5cm} p{3.5cm} p{3cm} X}
\toprule

\rowcolor{gray!12}
\textbf{Component} & \textbf{Field Name} & \textbf{Type/Format} & \textbf{Description} \\
\midrule

\multirow{7}{*}{\textbf{Tree-sitter Output}} 
& \texttt{id} & Integer & Unique node identifier assigned incrementally during AST traversal. \\
& \texttt{type} & Enum  & Node type in the abstract syntax tree: \texttt{\{directory, python\_file, class\_def, func\_def\}} \\
& \texttt{path} & String (absolute path) & Absolute filesystem path; later converted to repository-relative path in the dataset. \\
& \texttt{qualname} & String (dotted notation) or \texttt{null} & Qualified name constructed during traversal (e.g., \texttt{.IncorrectEnvError.\_\_init\_\_}) for definition nodes; \texttt{null} otherwise. \\
& \texttt{start} & $[line, column]$  & Start coordinates of the definition (\texttt{class} or \texttt{def} keyword). \\
& \texttt{code} & String & Raw source code: full file for \texttt{python\_file}; definition text for \texttt{class\_def}/\texttt{func\_def}. \\
& \texttt{superclasses} & List of objects & For \texttt{class\_def} nodes: base-class expressions with \texttt{text}, \texttt{start}, and \texttt{end} fields. \\
& \texttt{calls} & List of objects & For definition nodes: call expressions found within the definition. Each entry includes: \\
& & & \hspace{0.5cm}-- \texttt{text}: exact source substring of the call. \\
& & & \hspace{0.5cm}-- \texttt{start}/\texttt{end}: span coordinates; \texttt{start} used as Jedi query point. \\
\midrule

\multirow{8}{*}{\textbf{Jedi Output}}
& \texttt{repo\_root} & String & Repository root path for computing relative paths. \\
& \texttt{files[*].file} & String & File being analyzed by Jedi (repository-relative). \\
& \texttt{files[*].defs[*].def} & String & Enclosing definition (caller/subclass) for which relations are inferred. \\
& \texttt{def\_start} & $[line, column]$ & Start coordinates of the enclosing definition. \\
& \texttt{calls[*]} & List of objects & Results for each Tree-sitter call-site query: \\
& & & \hspace{0.5cm}-- \texttt{call\_text}, \texttt{call\_start}: call expression and query coordinates. \\
& & & \hspace{0.5cm}-- \texttt{candidates}: raw Jedi candidates (includes builtins), each with: \\
& & & \hspace{1cm}• \texttt{name}, \texttt{type}, \texttt{full\_name}, \texttt{module\_path}, \texttt{line}. \\
& & & \hspace{0.5cm}-- \texttt{joined\_targets}: subset of candidates matching \texttt{line2defdict}. Each includes: \\
& & & \hspace{1cm}• \texttt{join\_key} = (\texttt{relpath(module\_path)}, \texttt{line}) \\
& & & \hspace{1cm}• \texttt{target\_node\_id}, \texttt{target\_qualname} \\
& \texttt{superclasses[*]} & List of objects & Similar structure to \texttt{calls[*]}, for superclass token queries in class definitions. \\
\bottomrule
\end{tabularx}
\end{table*}

For a single commit snapshot, we extract (directed) edges of four semantic families; each family is stored with both forward and reverse directions:
\begin{itemize}
    \item \textbf{Contain / ContainedIn} (\texttt{edge\_attr} 0 / 1): hierarchical containment from directory$\to$file$\to$class$\to$function.
    \item \textbf{Call / CalledBy} (\texttt{edge\_attr} 2 / 3): inter-procedural call dependencies between definition nodes.
    \item \textbf{Superclass / Subclass} (\texttt{edge\_attr} 4 / 5): inheritance dependencies where a subclass node points to its superclass node.
    \item \textbf{Previous / Next} (\texttt{edge\_attr} 6 / 7): temporal linkage between successive versions of the \emph{same} entity across commits.
\end{itemize}

\subsection{Tree-sitter stage}
\label{app:treesitter-io}

To illustrate the above process more intuitively, we take a real code repository, \texttt{poetry}\footnote{\url{https://github.com/python-poetry/poetry}}, as an example. We analyze two compact files to demonstrate both (i) joinable inheritance and (ii) a joinable call.

\begin{lstlisting}[style=repo]
poetry/src/poetry/utils/env/exceptions.py
  class EnvError(Exception): ...
  class IncorrectEnvError(EnvError): ...
  class EnvCommandError(EnvError): ...

poetry/src/poetry/utils/threading.py
  class AtomicCachedProperty(functools.cached_property[T]): ...

  def atomic_cached_property(...):
      return AtomicCachedProperty(func)
\end{lstlisting}

\paragraph{Input.}
Tree-sitter operates on each \texttt{.py} file (full source text) at a given commit. It produces an AST from which we identify:
\begin{itemize}
    \item \textbf{class\_definition} nodes $\Rightarrow$ create \texttt{class\_def} graph nodes;
    \item \textbf{function\_definition} nodes $\Rightarrow$ create \texttt{func\_def} graph nodes;
    \item \textbf{call} nodes $\Rightarrow$ collect call-site spans under the \emph{enclosing} definition node;
    \item \textbf{superclasses field} in a class definition $\Rightarrow$ collect the token spans of each base class expression.
\end{itemize}

\paragraph{Output contract for Jedi.}
Crucially, the Tree-sitter stage does \emph{not} resolve targets. Instead, it emits \textbf{source coordinates} for later inference:
\begin{itemize}
    \item \textbf{Call sites}: a list of $(\ell, c)$ spans (\texttt{start\_point}, \texttt{end\_point}) for each \texttt{call} node.
    \item \textbf{Superclass tokens}: a list of $(\ell, c)$ spans for each base class token in \texttt{class Model(Base): ...}.
\end{itemize}
All coordinates are 1-indexed, matching the expected API of \texttt{jedi.Script.infer(line, column)}.

The following excerpt shows the actual Tree-sitter extraction for Poetry and how (i) the superclass token span is recorded for local inheritance and (ii) the enclosing function contains the call-site span:

\begin{lstlisting}[style=repo]
{
  "type": "class_def",
  "path": ".../poetry/src/poetry/utils/env/exceptions.py",
  "qualname": ".IncorrectEnvError",
  "start": [16, 1],
  "superclasses": [
    { "text": "EnvError", "start": [16, 25], "end": [16, 33] }
  ]
},
{
  "id": 2,
  "type": "class_def",
  "path": ".../poetry/src/poetry/utils/threading.py",
  "qualname": ".AtomicCachedProperty",
  "start": [21, 1],
  "superclasses": [
    { "text": "functools.cached_property[T]", "start": [21, 28], "end": [21, 56] }
  ],
  "calls": []
},
{
  "id": 7,
  "type": "func_def",
  "path": ".../poetry/src/poetry/utils/threading.py",
  "qualname": ".atomic_cached_property",
  "start": [52, 1],
  "calls": [
    { "start": [69, 12], "end": [69, 38], "text": "AtomicCachedProperty(func)" }
  ]
}
\end{lstlisting}

\subsection{Jedi stage}
\label{app:jedi-io}

\paragraph{Input.}
For each Python file, we instantiate a Jedi script object bound to that file: $\texttt{script} \leftarrow \texttt{jedi.Script(path=python\_file\_path)}.
$
For each \textbf{call-site start coordinate} $(\ell, c)$ collected by Tree-sitter, we query:$
\texttt{definitions} \leftarrow \texttt{script.infer}(\ell, c).
$ Similarly, for each \textbf{superclass token} start coordinate, we query \texttt{infer} to resolve the base class definition.

\paragraph{Output.}
Jedi returns a list of candidate symbolic targets. We keep candidates only if module\_path is not None (filters out builtins / unknown) and the candidate can be joined back to a known extracted definition via the join key $(\texttt{relpath(module\_path)}, \texttt{line})$. When multiple candidates match, we conservatively add edges to \emph{all} joinable targets.

The following excerpt shows the actual Jedi outputs and join-back targets in Poetry. Importantly, edges are created only from \texttt{joined\_targets}:

\begin{lstlisting}[style=repo]
{
  "token_text": "EnvError",
  "token_start": [16, 25],
  "candidates": [
    {
      "name": "EnvError",
      "type": "class",
      "module_path": ".../poetry/src/poetry/utils/env/exceptions.py",
      "line": 12
    }
  ],
  "joined_targets": [
    {
      "join_key": ["src/poetry/utils/env/exceptions.py", 12],
      "target_qualname": ".EnvError"
    }
  ]
}

{
  "call_text": "AtomicCachedProperty(func)",
  "call_start": [69, 12],
  "candidates": [
    {
      "name": "AtomicCachedProperty",
      "type": "class",
      "module_path": ".../poetry/src/poetry/utils/threading.py",
      "line": 21
    }
  ],
  "joined_targets": [
    {
      "join_key": ["src/poetry/utils/threading.py", 21],
      "target_qualname": ".AtomicCachedProperty"
    }
  ]
}
\end{lstlisting}

\subsection{Temporal metadata and ``graph of one commit''}
\label{app:temporal}

The repository evolves across commits, and the graph must support selecting a \emph{commit-local} view. Each node is therefore associated with a lifespan interval:
\begin{equation}
[\texttt{start\_commit}, \texttt{end\_commit})
\end{equation}
where \texttt{end\_commit} is \texttt{none} if the node remains alive at the end of the analyzed history. When a path or definition is re-introduced (same \texttt{path} and \texttt{name}), we link the new node to its prior version via a directed \texttt{previous} edge (and a reverse \texttt{next} edge).

\paragraph{Commit-local subgraph.}
For any target commit with topological timestamp $t$, the \emph{graph of one commit} is obtained by selecting nodes with:
\begin{equation}
\texttt{starttimestamp} \le t < \texttt{endtimestamp},
\end{equation}
and retaining edges whose endpoints are both alive at $t$. This yields a heterogeneous snapshot graph grounded in a single commit.

\paragraph{Timestamp propagation and call-closure.}
Two additional post-processing steps improve temporal consistency:
\begin{itemize}
    \item \textbf{Containment-consistent lifespans}: end timestamps are propagated from containers to contained nodes so that a child cannot outlive its parent in the containment hierarchy.
    \item \textbf{Call-closure across versions}: if a caller calls an older version of a callee, and that callee has a \texttt{next} version, we add a call edge to the newer version when the lifespans overlap. This reduces false ``stale-target'' calls when code is updated across commits.
\end{itemize}

\section{Additional Branchy Commit-DAG Examples}
\label{app:branchy-dag-examples}

To further illustrate that our repositories exhibit genuine branching and merging (beyond a single chain), we provide additional branchy DAG visualizations extracted from real repositories in our dataset.
In each panel, we use the same color convention as Figure~\ref{fig:temporal-longest-path-b}: \textcolor{blue}{blue} for the global longest path, \textcolor{orange}{orange} for the test-time prefix ending at a bug-associated commit (when available), \textcolor{red}{red} for the bug-associated commit, and \textcolor{gray}{gray} for off-longest-path commits that realize branches and merges.

\begin{figure}[p]
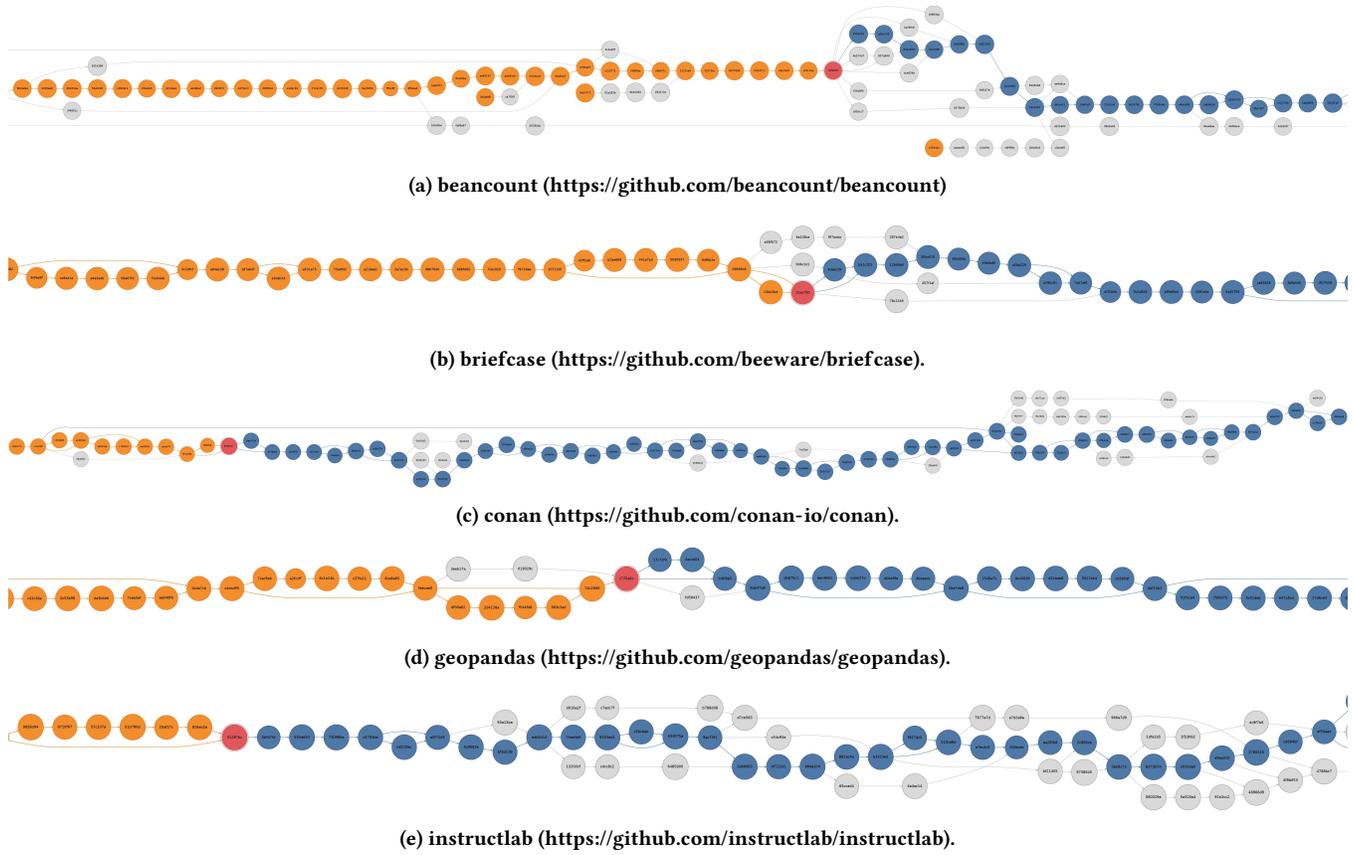

\centering
\begin{subfigure}[t]{\linewidth}
\centering
\includegraphics[width=\textwidth]{beancount_dag_branchy.pdf}
\caption{\textbf{beancount} (\url{https://github.com/beancount/beancount)}}
\label{fig:branchy-examples-a}
\end{subfigure}

\vspace{0.6em}

\begin{subfigure}[t]{\linewidth}
\centering
\includegraphics[width=\textwidth]{briefcase_dag_branchy.pdf}
\caption{\textbf{briefcase} (\url{https://github.com/beeware/briefcase}).}
\label{fig:branchy-examples-b}
\end{subfigure}

\vspace{0.6em}

\begin{subfigure}[t]{\linewidth}
\centering
\includegraphics[width=\linewidth]{conan_dag_branchy.pdf}
\caption{\textbf{conan} (\url{https://github.com/conan-io/conan}).}
\label{fig:branchy-examples-c}
\end{subfigure}

\vspace{0.6em}

\begin{subfigure}[t]{\linewidth}
\centering
\includegraphics[width=\linewidth]{geopandas_dag_branchy.pdf}
\caption{\textbf{geopandas} (\url{https://github.com/geopandas/geopandas}).}
\label{fig:branchy-examples-d}
\end{subfigure}

\vspace{0.6em}

\begin{subfigure}[t]{\linewidth}
\centering
\includegraphics[width=\linewidth]{instructlab_dag_branchy.pdf}
\caption{\textbf{instructlab} (\url{https://github.com/instructlab/instructlab}).}
\label{fig:branchy-examples-e}
\end{subfigure}

\caption{Five additional branchy commit-DAG visualizations from real repositories in our dataset (top to bottom: (a)--(e)).}
\label{fig:branchy-examples}
\end{figure}

\section{Incremental Building Details}
\label{app:incremental-building}

\subsection{Definitions and measurement protocol}
\label{app:incremental-defs}

\paragraph{Commit path.}
For temporal indexing we use a single, consistent commit sequence per repository (the same path used during graph construction), i.e., the selected ``longest path'' extracted from the repository commit DAG.
All per-commit statistics below are computed along this path.

\paragraph{$\text{Changed}/\text{Total}$ ratio.}
For each commit on the path, we compute:
(i) \textbf{Total}: the number of Python files in the repository at that commit (tree traversal);
(ii) \textbf{Changed}: the number of \emph{distinct} Python files that appear in the patch between the previous commit and the current commit (diff over \texttt{a\_path}/\texttt{b\_path}).
The ratio is $\text{Changed}/\text{Total}$.
This ratio directly proxies the fraction of the codebase that would require reparsing in an incremental pipeline, versus a full rebuild that reparses all files.

\paragraph{Node update magnitude.}
Using the temporal node semantics (\texttt{start\_commit}, \texttt{end\_commit}), we compute per-commit counts of:
(i) nodes \textbf{added} at a commit (nodes whose \texttt{start\_commit} equals that commit);
(ii) nodes \textbf{removed} at a commit (nodes whose \texttt{end\_commit} equals that commit).
These counts are shown as smoothed time series for readability, while preserving overall trends.

\paragraph{Interpretation.}
If most commits have small $\text{Changed}/\text{Total}$, then an incremental builder avoids repeated full-graph reconstruction.
If node update magnitudes remain bounded, then the temporal graph can be maintained with localized edits, which also benefits downstream pipelines (e.g., embeddings/feature updates) that can reuse untouched parts.

\subsection{Additional repositories}
\label{app:incremental-more-repos}

We provide additional per-repository diagnostic figures using the same plotting routine, to demonstrate that the observed incremental behavior is not specific to a single project.
Figures are generated from real repositories and their corresponding graph construction outputs.

\begin{figure*}[t]
  \centering
  \begin{subfigure}{0.48\textwidth}
    \centering
    \includegraphics[width=\linewidth]{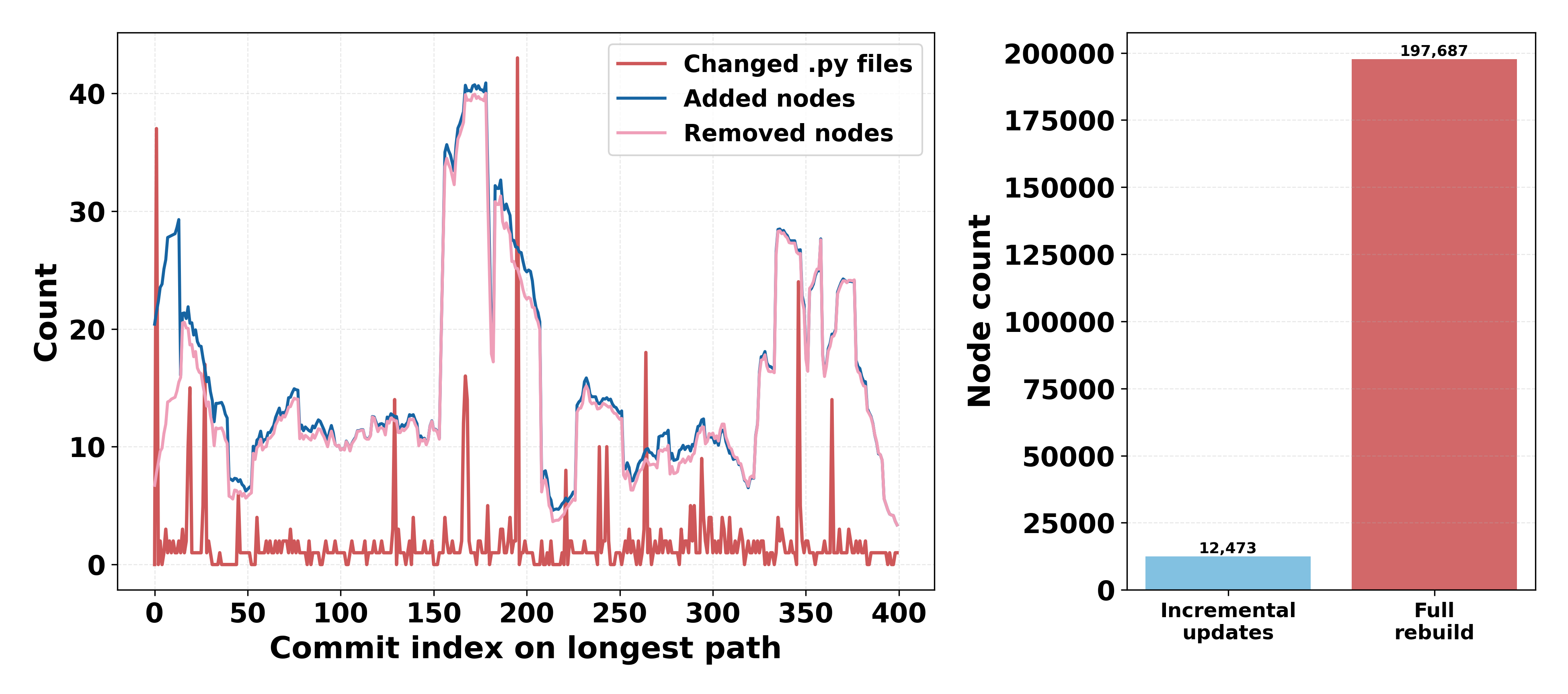}
  \end{subfigure}\hfill
  \begin{subfigure}{0.48\textwidth}
    \centering
    \includegraphics[width=\linewidth]{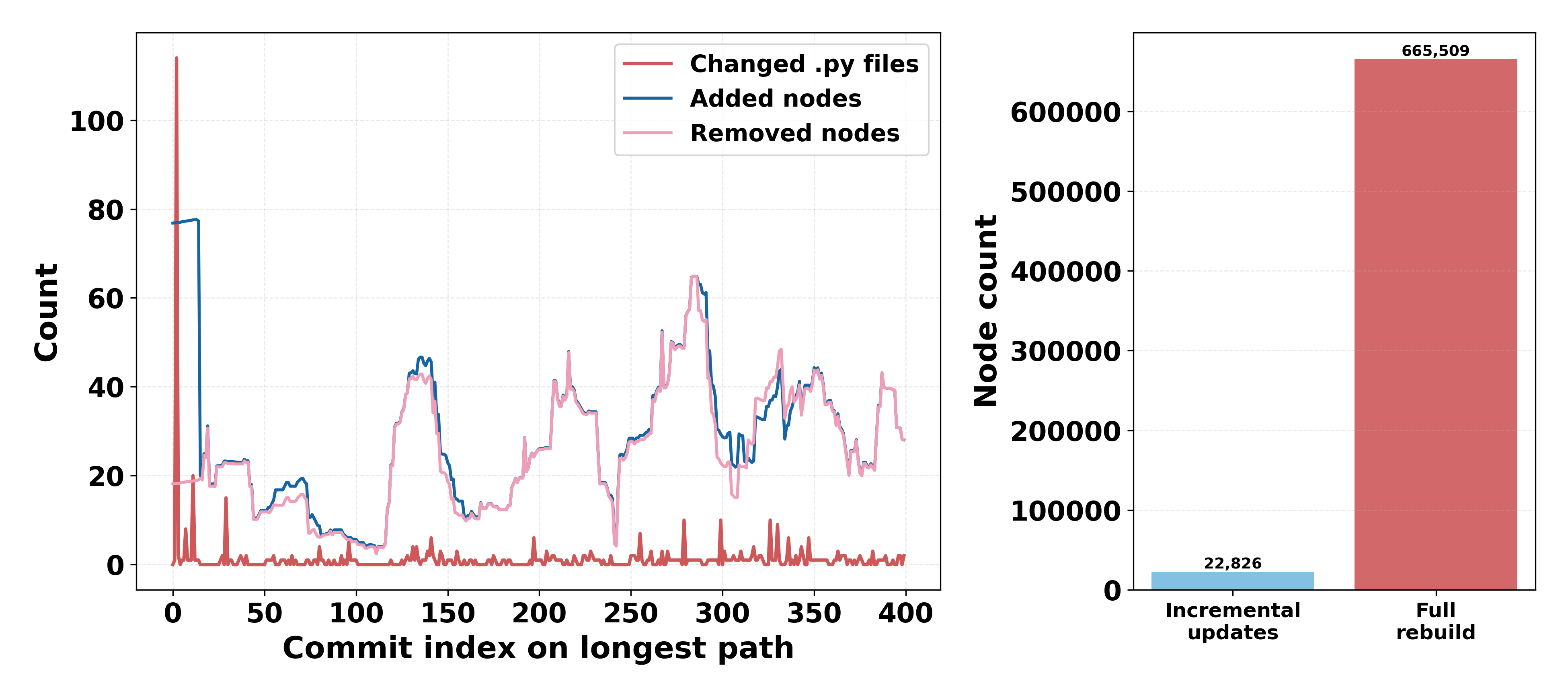}
  \end{subfigure}
  
  \vspace{0.6em}
  
  \begin{subfigure}{0.48\textwidth}
    \centering
    \includegraphics[width=\linewidth]{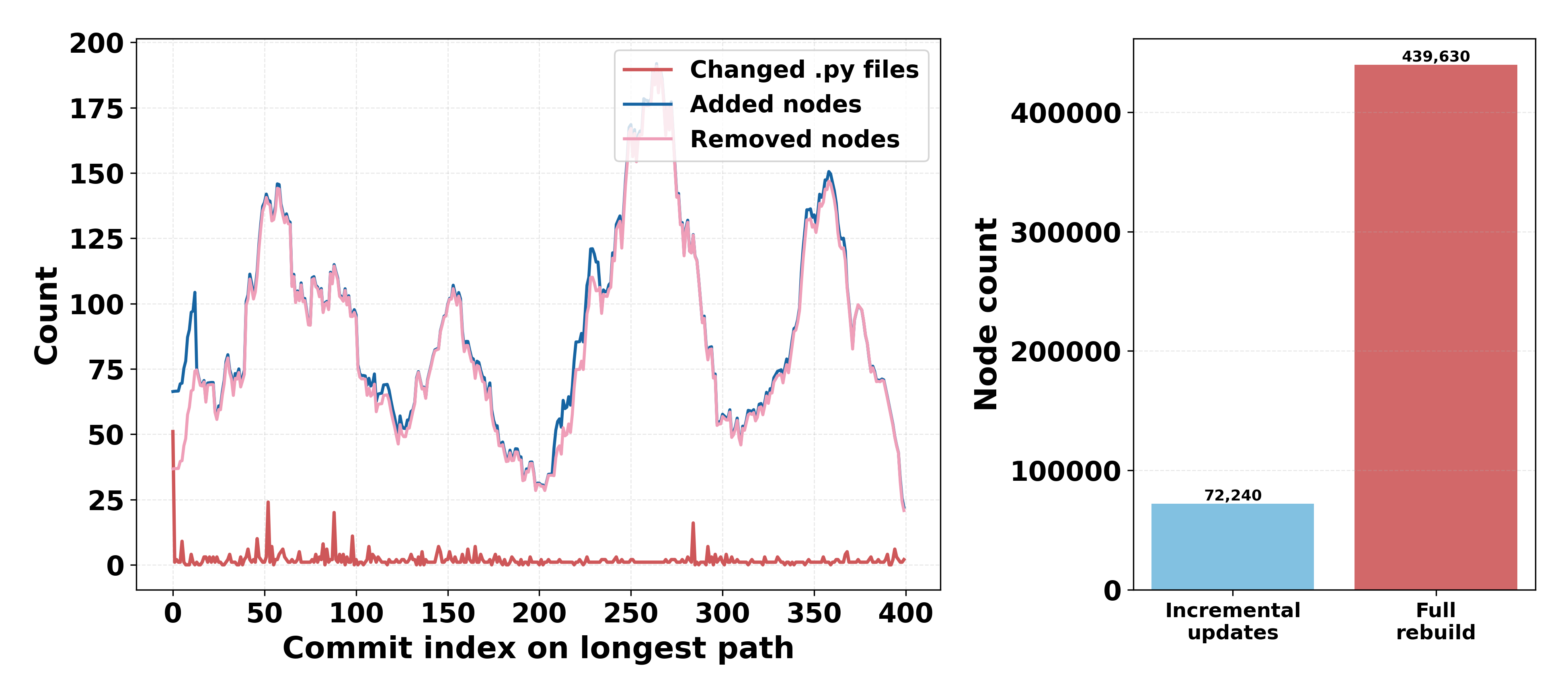}
  \end{subfigure}\hfill
  \begin{subfigure}{0.48\textwidth}
    \centering
    \includegraphics[width=\linewidth]{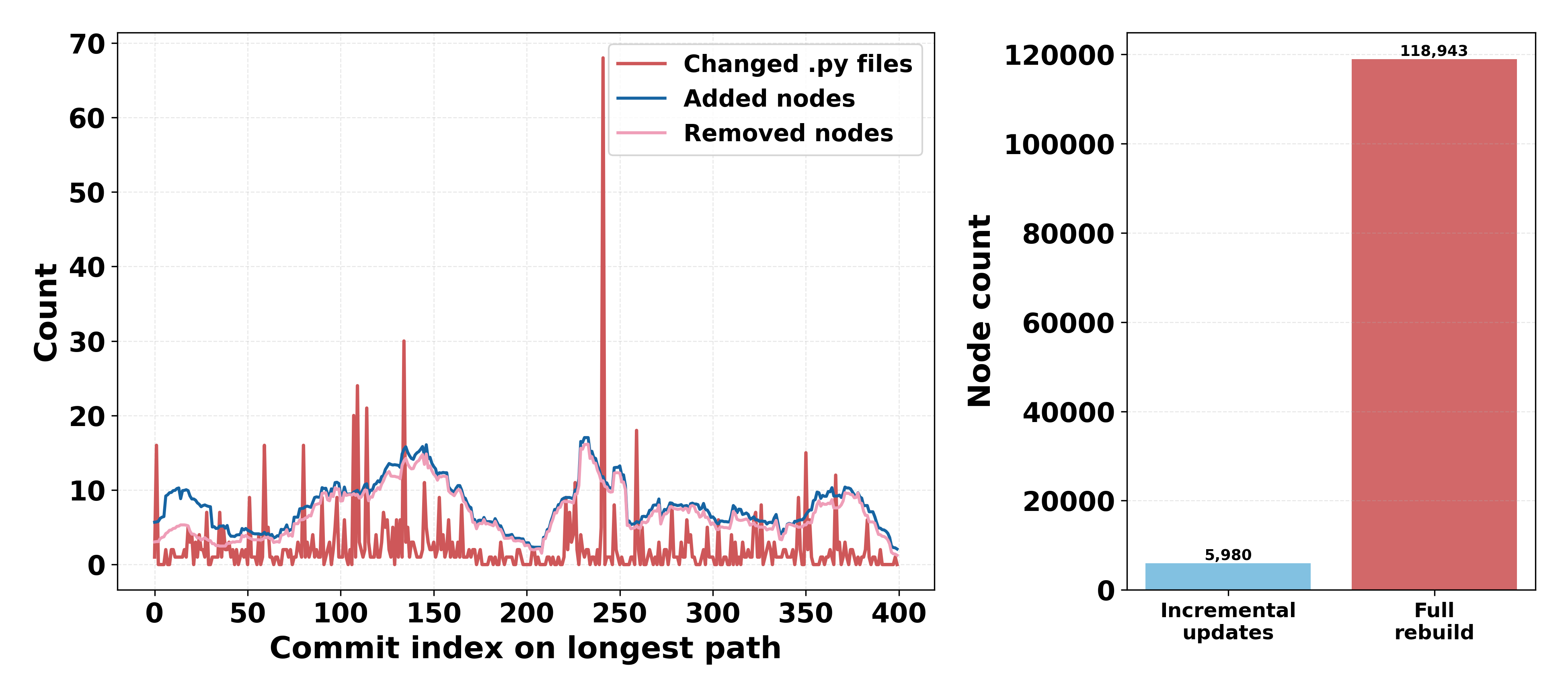}
  \end{subfigure}

  \caption{\textbf{Incremental construction diagnostics on four real repositories.}
  Each panel contains two subplots: (left) per-commit changed \texttt{.py} files with incremental node updates (added/removed), and (right) a direct cost comparison between incremental updates and full rebuilds (sum of per-commit alive nodes).
  From left to right, top to bottom: \textsc{Conda}, \textsc{Django}, \textsc{IPython}, and \textsc{Matplotlib}.}
  \label{fig:incremental-appendix-4repos}
\end{figure*}

\subsection{Implementation notes}
\label{app:incremental-impl}

\paragraph{Patch to changed-file set.}
For each adjacent commit pair on the selected path, we collect the set of files affected by the diff, filtering to \texttt{.py}.
This file set gates reparsing and local edge updates.

\paragraph{Temporal node semantics.}
Nodes are versioned by identity keys (path/name for definitions and paths), enabling reuse across commits and explicit closure when removed.
When an entity reappears, a new node is created and linked to its prior version via a version edge (e.g., \textsc{Previous}/\textsc{Next}).
This produces a compact temporal trace without duplicating the entire repository state at each commit.

\paragraph{Reverse edges.}
Edges that are straightforward inversions of forward relations (e.g., \textsc{CalledBy} from \textsc{Call}) are constructed in a post-processing pass.
This keeps incremental updates focused on forward extraction while still providing a fully usable bidirectional graph for downstream consumption.

\paragraph{Limitations of the plotted proxy.}
$\text{Changed}/\text{Total}$ is a conservative file-level proxy: a changed file may only touch a small region, and some edits may not change the extracted entities.
Nevertheless, it directly captures the dominant cost driver for AST-based extraction (parsing), and is therefore a faithful indicator of incremental savings at scale.

\section{Temporal Relation Among Commits}\label{app:commit_temp_rel}

We collect each repository's full development history (commits, pull requests, and issues) and model the commit history as a directed acyclic graph (DAG), where edges encode parent-to-child commit ancestry induced by branching and merging.
While the DAG represents the true history, it only defines a \emph{partial order}: commits on different branches are often incomparable without additional assumptions, which makes it difficult to assign a single consistent temporal index required by most sequence-based temporal models and time-conditioned graph learning objectives.
To obtain a \emph{total order} that is both well-defined and reproducible, we linearize the commit DAG by extracting a longest path and using it as the canonical timeline for the repository.
Concretely, we compute the longest path using dynamic programming on the commit DAG and use its topological order as the repository's temporal axis.

\textbf{Training vs.\ testing.}
For training, we use the global longest path of the repository history to define a consistent time index shared across all training samples.
For testing on a bug report, we restrict the timeline to the prefix of the longest path that terminates at the bug-associated commit (obtained via the linked PR/issue metadata), thereby preventing information leakage from future commits while preserving a total order within the accessible history.

\begin{figure}[t]
\centering
\begin{subfigure}[t]{\linewidth}
\centering
\includegraphics[width=\linewidth]{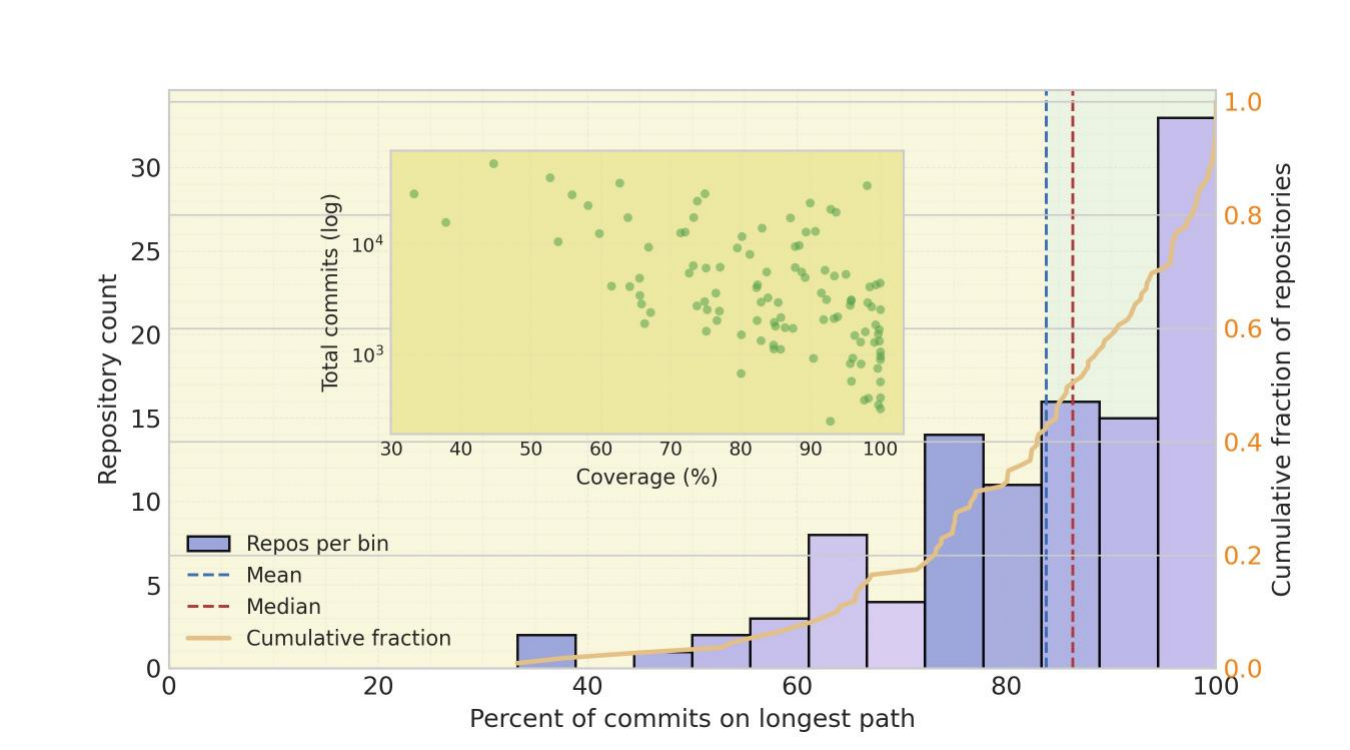}
\caption{\textbf{Longest-path coverage across repositories.}
Histogram of the fraction of commits that lie on the selected longest path, with an overlaid cumulative fraction curve. The inset shows repository size (total commits, log scale) versus longest-path coverage, illustrating how branch/merge intensity varies with repository scale.}
\label{fig:temporal-longest-path-a}
\end{subfigure}

\begin{subfigure}[t]{\linewidth}
\centering
\includegraphics[width=\linewidth]{astropy_dag_branchy.pdf}
\caption{\textbf{Example branchy commit DAG with highlighted linearization.}
Each node is a commit and each directed edge indicates parent$\rightarrow$child ancestry.
\textcolor{blue}{Blue} nodes/edges denote the selected global longest path (the canonical timeline).
\textcolor{orange}{Orange} nodes/edges denote the \emph{test path}, i.e., the longest-path prefix that ends at the bug-associated commit.
The bug-associated commit is emphasized in \textcolor{red}{red}.
\textcolor{gray}{Gray} nodes are real off-longest-path commits included to visualize branching/merging around the selected region.}
\label{fig:temporal-longest-path-b}
\end{subfigure}

\caption{Temporal linearization of commit histories via longest-path extraction.}
\label{fig:temporal-longest-path}
\end{figure}

\textbf{Empirical characteristics.}
Figure~\ref{fig:temporal-longest-path} summarizes the coverage distribution of longest-path linearization across our repositories and illustrates a concrete example\footnote{This schematic shows a local portion of the real commit DAG from the \texttt{astropy} repository: \url{https://github.com/astropy/astropy}.}  how the bug commit and the selected paths are positioned within a branchy DAG.
The distribution indicates that many repositories exhibit strong mainline dominance (high longest-path coverage), while some repositories are substantially more branch/merge heavy.
Importantly, regardless of the exact coverage, longest-path linearization provides a principled and deterministic way to impose an ordering when cross-branch temporal comparison is otherwise ill-defined.

\section{Task/Label Collection Details}\label{app:label_details}

\subsection{Issue--PR linking regex and keyword set.}
We follow GitHub's documented closing keywords and common variants. Let $\mathcal{K}$ be the keyword set:
\begin{tcolorbox}[colback=black!3,colframe=black!60,boxrule=0.6pt,arc=2pt,left=6pt,right=6pt,top=6pt,bottom=6pt]
\begin{quote}\small
\texttt{\{close, closes, closed, fix, fixes, fixed, resolve, resolves, resolved,\newline
close issue, closes issue, closed issue, fix issue, fixes issue, fixed issue,\newline
resolve issue, resolves issue, resolved issue, close the issue, closes the issue,\newline
closed the issue, fix the issue, fixes the issue, fixed the issue, resolve the issue,\newline
resolves the issue, resolved the issue, solve, solves, solved, solve issue, solves issue,\newline
solved issue, solve the issue, solves the issue, solved the issue\}}.
\end{quote}
\end{tcolorbox}
We match issue numbers with:
\begin{tcolorbox}[colback=black!3,colframe=black!60,boxrule=0.6pt,arc=2pt,left=6pt,right=6pt,top=6pt,bottom=6pt]
\begin{quote}\small
\texttt{(?i)(?:\textbackslash b(?:KEYWORDS)\textbackslash b\textbackslash s*\#(\textbackslash d+))}
\end{quote}
\end{tcolorbox}
where \texttt{KEYWORDS} is the \texttt{|}-joined set above. Before matching, we remove HTML comments with:
\begin{tcolorbox}[colback=black!3,colframe=black!60,boxrule=0.6pt,arc=2pt,left=6pt,right=6pt,top=6pt,bottom=6pt]
\begin{quote}\small
\texttt{(?s)<!--.*?-->}.
\end{quote}
\end{tcolorbox}
We search over the concatenation of PR title, PR body, and all commit messages in the PR.
We de-duplicate issue IDs and drop self-references where the extracted issue number equals the PR number.

\begin{tcolorbox}[colback=black!3,colframe=black!60,boxrule=0.6pt,arc=2pt,left=6pt,right=6pt,top=6pt,bottom=6pt]
\textbf{Implementation note (dataset schema).}
In our crawler output, the \texttt{issues} field is a list of integers whose first element is a dummy sentinel \texttt{-1}; the remaining elements are linked issue numbers.
For storage efficiency, the \texttt{issues\_info} field is a single string formed by concatenating JSON-serialized issue dictionaries with the separator \texttt{\#@!@\#} .
\end{tcolorbox}

\subsection{Exact regexes and text normalization.}
We explicitly mirror GitHub's closing-keyword convention.\footnote{\url{https://docs.github.com/en/issues/tracking-your-work-with-issues/linking-a-pull-request-to-an-issue}}
Our extraction proceeds in three steps:
\textbf{(i)} build a search string by concatenating the PR title, PR body, and all commit messages in the PR;
\textbf{(ii)} remove HTML comments; and
\textbf{(iii)} match keyword--issue-number patterns case-insensitively.

\begin{tcolorbox}[colback=black!3,colframe=black!60,boxrule=0.6pt,arc=2pt,left=6pt,right=6pt,top=6pt,bottom=6pt]
\ttfamily\small
\textbf{(HTML comment stripping)}\\
(?s)<!--.*?-->\\begin{equation}6pt]
\textbf{(Issue reference extraction)}\\
(?i)(?:\textbackslash b(?:KEYWORDS)\textbackslash b\textbackslash s*\#(\textbackslash d+))
\end{tcolorbox}

\noindent
\textbf{Post-processing rules.}
We deduplicate extracted issue numbers, and remove self-references where an extracted number equals the PR number.
If a PR references multiple issues (common in maintenance PRs), we keep \emph{all} matched issue IDs; the downstream task can either treat them as separate examples or keep the unioned issue text depending on the experiment design.
Our default dataset construction keeps all linked issues but enforces leakage-safe text usage (next subsection).

\subsection{Leakage control and input text construction.}
The benchmark task is \emph{localization from a bug report}. To preserve realism and avoid trivial shortcuts, we strictly control what textual fields are used as model input:
\begin{itemize}
  \item \textbf{Used as input:} issue title and the issue's initial description body (the first message written by the reporter).
  \item \textbf{Excluded from input:} issue comments (discussion thread), PR discussion/reviews, PR body (fix explanation), and code diffs/patches.
\end{itemize}
\noindent
Rationale: issue comments and PR text frequently include the solution, a patch, or explicit file paths, which would leak labels and substantially inflate localization performance.
We still retain these fields in the raw crawl output for auditing and analysis, but they are not fed to models during training/evaluation.

\subsection{Rule-based issue body parsing (template headings).}
Many mature repositories use issue templates (e.g., GitHub forms or Markdown templates) that organize reports into heading-delimited sections.
We exploit this structure to obtain cleaner, comparable fields across repos.

\paragraph{Heading segmentation.}
We split the issue body into sections using Markdown headings (two or more \texttt{\#}) and capture each heading title with its following content:
\begin{tcolorbox}[colback=black!3,colframe=black!60,boxrule=0.6pt,arc=2pt,left=6pt,right=6pt,top=6pt,bottom=6pt]
\ttfamily\small
\^ \#{2,}\textbackslash s*(.+?)\textbackslash s*\textbackslash n(.*?)(?=\^\#{2,}\textbackslash s|\textbackslash Z)
\end{tcolorbox}
\noindent
We run this with \texttt{re.MULTILINE | re.DOTALL}.
Any residual text not under a recognized heading is appended into an \texttt{others} field.

\paragraph{Slot mapping heuristics.}
We normalize headings to lowercase and map common headings to canonical slots:
\texttt{bug\_desc}, \texttt{reproduce}, \texttt{expected\_behavior}, \texttt{actual\_behavior}, \texttt{version}, \texttt{require}, \texttt{solution}, and \texttt{others}.
We include both exact template matches (repo-specific phrasing) and robust fuzzy matching rules (e.g., headings containing ``expected'', ``actual'', ``reproduce'', ``version'').
This design makes the parser resilient to minor template variations and partial/informal reports.

\paragraph{Additional signal extraction (code and tracebacks).}
Beyond headings, we explicitly extract three signals that are highly predictive for localization:
\begin{itemize}
  \item \textbf{Code blocks} (verbatim content inside triple backticks), often containing minimal reproductions.
  \item \textbf{Tracebacks} (if present), which frequently reveal the failing call chain.
  \item \textbf{Traceback frames} (file paths, line numbers, function names), enabling fine-grained error localization analysis.
\end{itemize}
\begin{tcolorbox}[colback=black!3,colframe=black!60,boxrule=0.6pt,arc=2pt,left=6pt,right=6pt,top=6pt,bottom=6pt]
\ttfamily\small
\textbf{(Code blocks)}\\
```(.+?)```\\begin{equation}4pt]
\textbf{(Traceback body)}\\
Traceback \textbackslash (most recent call last\textbackslash ):(.+?)(?=\^```|\textbackslash Z)\\begin{equation}4pt]
\textbf{(Traceback frames)}\\
File "(.+?)", line (\textbackslash d+), in (.+?)(?:\textbackslash s*\textbackslash n|\$)
\end{tcolorbox}
\noindent
We store the full traceback text, the final non-empty error line (\texttt{error\_statement}), and the last referenced file/function (when available).

\subsection{LLM segmentation prompt.}
When LLM segmentation outputs are available, we use them as an optional, higher-recall alternative to rule-based parsing.
Critically, the prompt instructs the model to \textbf{preserve original wording} and perform \textbf{non-overlapping assignment} (each span goes to exactly one slot), minimizing paraphrase-induced distribution shift.

\textbf{Bug-report segmentation (verbatim prompt excerpt):}

\begin{tcolorbox}[colback=black!3,colframe=black!60,boxrule=0.6pt,arc=2pt,left=6pt,right=6pt,top=6pt,bottom=6pt]
\ttfamily\small
Your task is to categorize the content of an bug report issue into the following sections. Use the exact original text from the issue---do not modify or paraphrase anything. Assign each part to exactly one category. If a category has no relevant content, leave it empty. Return a json format dictionary with the specified keys:
\\
1. Bug Description \\
2. Reproduction \\
3. Expected Behavior \\
4. Actual Behavior \\
5. Environment \\
6. Other \\
\\
Rules: preserve wording; do not omit/merge; empty slot $\rightarrow$ "".
\\
<issues> \{issue\_str\} </issues>
\end{tcolorbox}

\textbf{Feature-request segmentation (verbatim prompt excerpt):}

\begin{tcolorbox}[colback=black!3,colframe=black!60,boxrule=0.6pt,arc=2pt,left=6pt,right=6pt,top=6pt,bottom=6pt]
\ttfamily\small
Your task is to categorize the content of an feature request issue into the following sections. Use the exact original text---do not modify or paraphrase.
\\
1. Feature Description \\
2. Proposed Solution \\
3. Other
\end{tcolorbox}

\noindent
\textbf{Post-processing.}
We discard any intermediate ``thoughts'' field (if present) and keep only the segmented text spans for downstream use.
This makes the segmentation output auditable and prevents hidden rationales from entering the dataset.

\subsection{Label extraction and graph mapping details.}
We define ground truth using \emph{what the fixing PR actually changed}:
\begin{itemize}
  \item \textbf{Source of truth:} PR-level file list and patches from the GitHub API.
  \item \textbf{Python-only filter:} keep only files ending with \texttt{.py}.
  \item \textbf{Snapshot alignment:} associate the PR with its \texttt{base\_commit\_sha}; map file paths to nodes in the repository graph snapshot valid at that commit.
  \item \textbf{Empty-label handling:} if no changed Python file maps to a snapshot file node, the example is discarded.
\end{itemize}
\noindent
This procedure yields objective, reproducible labels and ensures the supervision aligns with the exact node universe visible to the model at that time. At the same time, the detailed information about Quality Controls and edge cases is as follows:

\begin{itemize}
  \item \textbf{Patch availability:} we skip PRs with missing file patches (rare API artifact) to avoid silently corrupting labels.
  \item \textbf{Duplicate issue references:} repeated issue numbers are removed to prevent overweighting a single issue.
  \item \textbf{Multiple issues per PR:} we retain all linked issues; downstream experiments can select the primary issue, concatenate multiple issue texts, or sample one issue per PR for evaluation stability.
  \item \textbf{Non-templated issues:} if no headings are present, the parser falls back to storing remaining text in \texttt{others} and still extracts code/traceback signals when possible.
\end{itemize}

\section{Similarity Feature}
\label{app:sim-val}

\begin{figure}[t]
  \centering
  \includegraphics[width=0.5\linewidth]{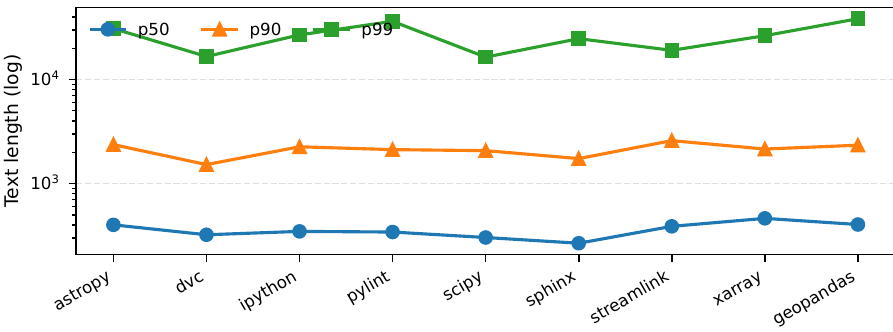}
  \caption{\textbf{Node text-length quantiles across nine repositories (log-scale).}
  We report p50/p90/p99 of canonicalized node text length (characters). All repositories exhibit a heavy-tailed distribution: most nodes are short, while a small fraction are extremely long, motivating truncation and batched embedding computation for stable and efficient feature construction.}
  \label{fig:node-text-length-9repos}
\end{figure}

Figure~\ref{fig:node-text-length-9repos} summarizes the distribution of node text lengths across our nine repositories using the 50th/90th/99th percentiles (computed over sampled nodes per repository; lengths are measured in characters on the canonicalized node text).
A consistent \emph{heavy-tailed} pattern emerges: the median node text is relatively short, while the upper tail (p99) is orders of magnitude larger, reflecting that most nodes correspond to small code units (e.g., short functions or identifiers), whereas a small fraction corresponds to very large entities (e.g., large files or long definitions).
This long-tail structure is important for feature construction because embedding computation must be robust to extreme-length inputs: without truncation and batching, a few very long nodes would dominate memory/latency and introduce instability.
Accordingly, our implementation uses batched encoding with a fixed maximum token length, ensuring predictable computational cost while still capturing semantics for the majority of nodes; the remaining long-text nodes are represented by truncated embeddings, which our downstream results show are sufficient for stable similarity-based retrieval and learning.
Finally, the fact that the quantile curves are broadly consistent across repositories suggests that the text/feature pipeline is not overfit to a single codebase and can be applied uniformly in a multi-repository setting.

Similarity is computed by inner product between node embeddings and query embeddings, yielding a per-node similarity vector (one dimension per rewritten query). We validate similarity as a feature via a \textbf{positives vs.\ negatives} probe:
\begin{itemize}
  \item \textbf{Positives}: nodes edited by the patch.
  \item \textbf{Hard negatives}: class/function nodes in the same modified files but not edited (obtained from file node \verb|contain| lists, excluding positives).
  \item \textbf{Random negatives}: randomly sampled class/function nodes outside the positive/hard-negative sets.
\end{itemize}
For each node, we score similarity as $\max_{q}\mathrm{sim}(v,q)$ (maximum over rewritten queries).
We report ROC-AUC using a tie-aware rank statistic (Mann--Whitney formulation), which directly measures the probability that a randomly drawn positive has a higher similarity score than a randomly drawn negative.

\subsection{Linear Probes on Frozen Embeddings}
\label{app:probe}
To further verify that embeddings preserve code-related semantics, we conduct linear-probe experiments: if a simple linear classifier can accurately predict labels from frozen embeddings, then the embeddings must encode linearly accessible semantic/structural information \cite{alain2016understanding}.
We evaluate 3 tasks:
\begin{enumerate}
  \item \textbf{node\_type\_5way}: directory / file / Python file / class def / func def.
  \item \textbf{is\_test\_binary}: test-like paths vs non-test (path contains \texttt{test/tests/pytest} or suffix \texttt{\_test.py}).
  \item \textbf{topdir\_multi}: top-level directory classification (top-$K$ most frequent, others $\rightarrow$ \texttt{other}).
\end{enumerate}

We train a single linear layer with cross-entropy (AdamW) on an 80/20 train/test split of sampled nodes (sample\_size=20,000 per repository; 10 epochs). Table~\ref{tab:linear-probe-acc-9repos} reports the final test accuracy.

\begin{table}[t]
\centering
\small
\setlength{\tabcolsep}{5pt}
\rowcolors{2}{blue!6}{white}
\begin{tabular}{lccc}
\toprule
\rowcolor{blue!12}
Repository &
Type-5 &
Test-Bin &
Topdir \\
\midrule
\texttt{astropy}    & 0.85475  & 0.97975 & 0.98125 \\
\texttt{dvc}        & 0.81475  & 0.98925 & 0.97150 \\
\texttt{ipython}    & 0.82450  & 0.93950 & 0.94400 \\
\texttt{pylint}     & 0.83250  & 0.87475 & 0.51375 \\
\texttt{scipy}      & 0.87500 & 0.96675 & 0.90475 \\
\texttt{sphinx}     & 0.83525  & 0.95300 & 0.93325 \\
\texttt{streamlink} & 0.75450  & 0.97700 & 0.94075 \\
\texttt{xarray}     & 0.92950  & 0.91225 & 0.77400 \\
\texttt{geopandas}  & 0.92350  & 0.98775 & 0.94750 \\
\bottomrule
\end{tabular}
\caption{\textbf{Linear probe test accuracy on frozen node embeddings across nine repositories.}
 (each repo: sample\_size=20{,}000; 80/20 train/test split; 10 epochs).
Tasks: Type-5 = directory/file/python-file/class/function; Test-Bin = test-related path vs non-test; Topdir = top-level directory multi-class (top-K frequent + \texttt{other}).}
\label{tab:linear-probe-acc-9repos}
\end{table}

\subsection{Similarity Case Studies}
\label{app:cases}

To complement our quantitative analyses, we provide qualitative similarity case studies.
For each repository, we show (i) a real issue report and (ii) several top-ranked code snippets retrieved purely by our similarity feature.
These examples serve two purposes: they make the retrieval signal interpretable to readers, and they verify that high similarity corresponds to semantically relevant code regions rather than accidental lexical overlap.

\begingroup
\graphicspath{{issue_snippet}}
\setlength{\parindent}{0pt}
\setlength{\parskip}{0pt}

\newcommand{\caseimg}[1]{%

\begin{center}

\includegraphics[width=0.8\linewidth]{#1}

\end{center}

\vspace{-0.35em}%

}
\newcommand{\repocap}[1]{%
  \vspace{0.15em}\par{\small\textit{#1}}\par\vspace{0.8em}%
}

\caseimg{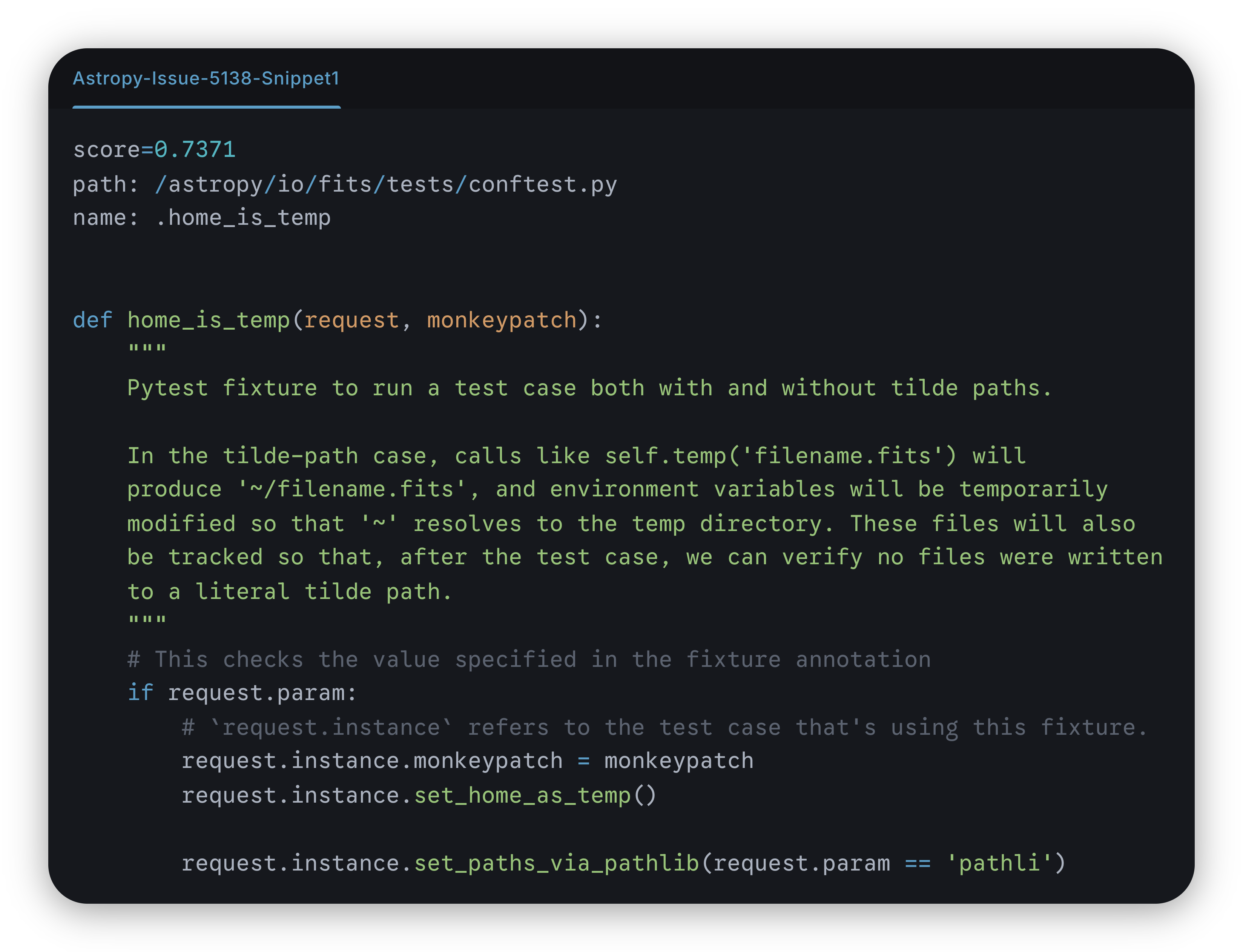}
\caseimg{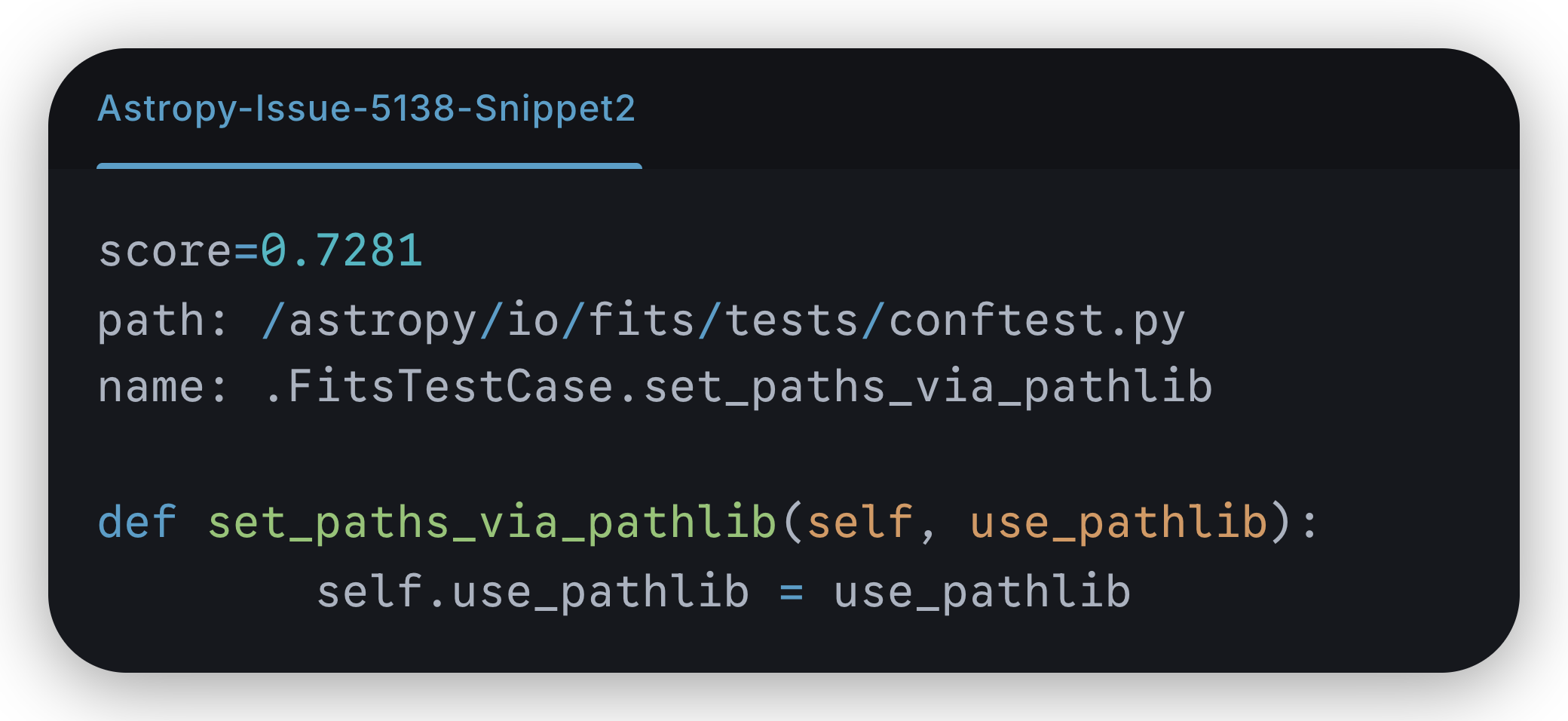}
\caseimg{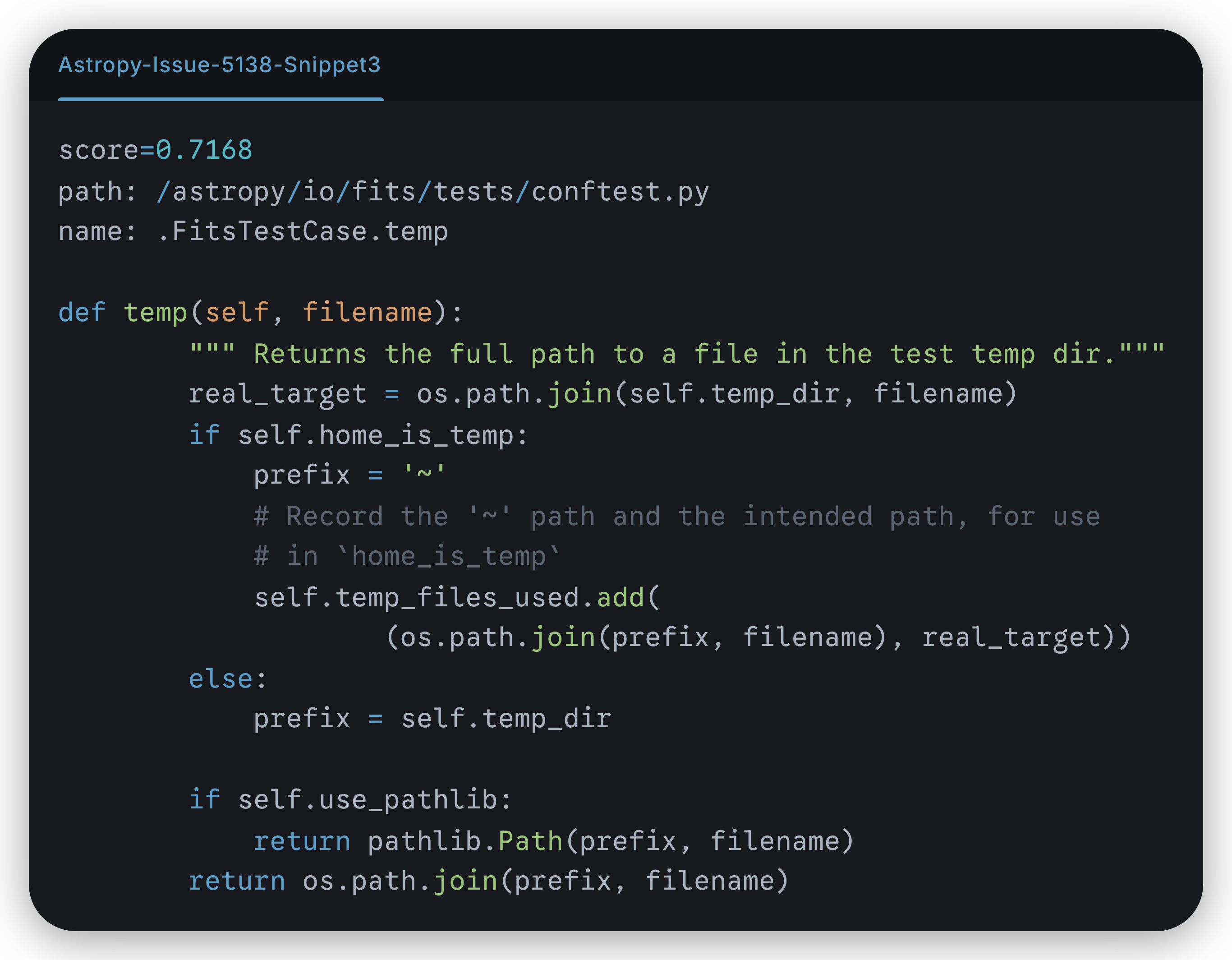}
\repocap{\textbf{Astropy.} Issue \#5138 and its top-ranked retrieved snippets (sorted by similarity).}

\caseimg{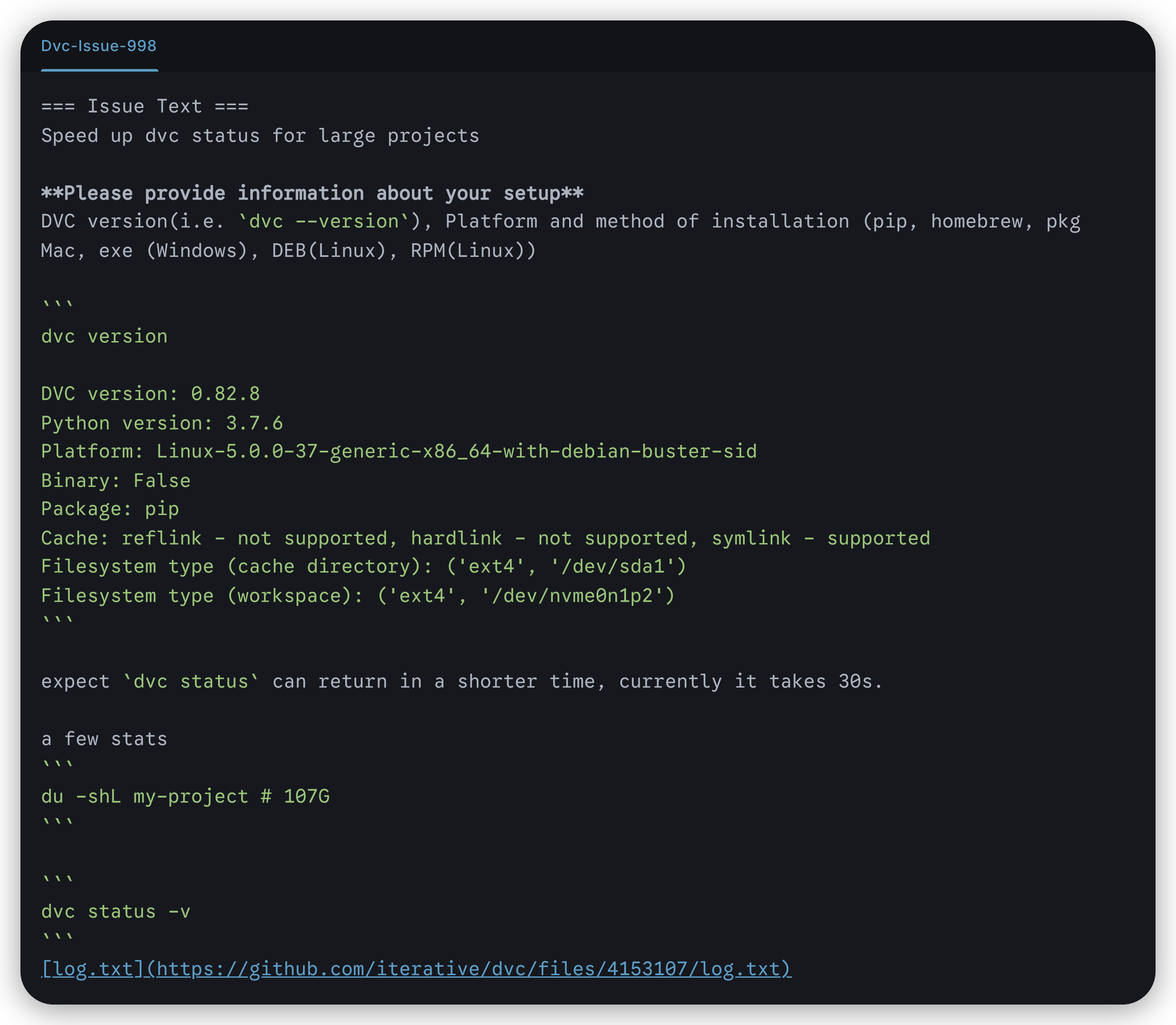}
\caseimg{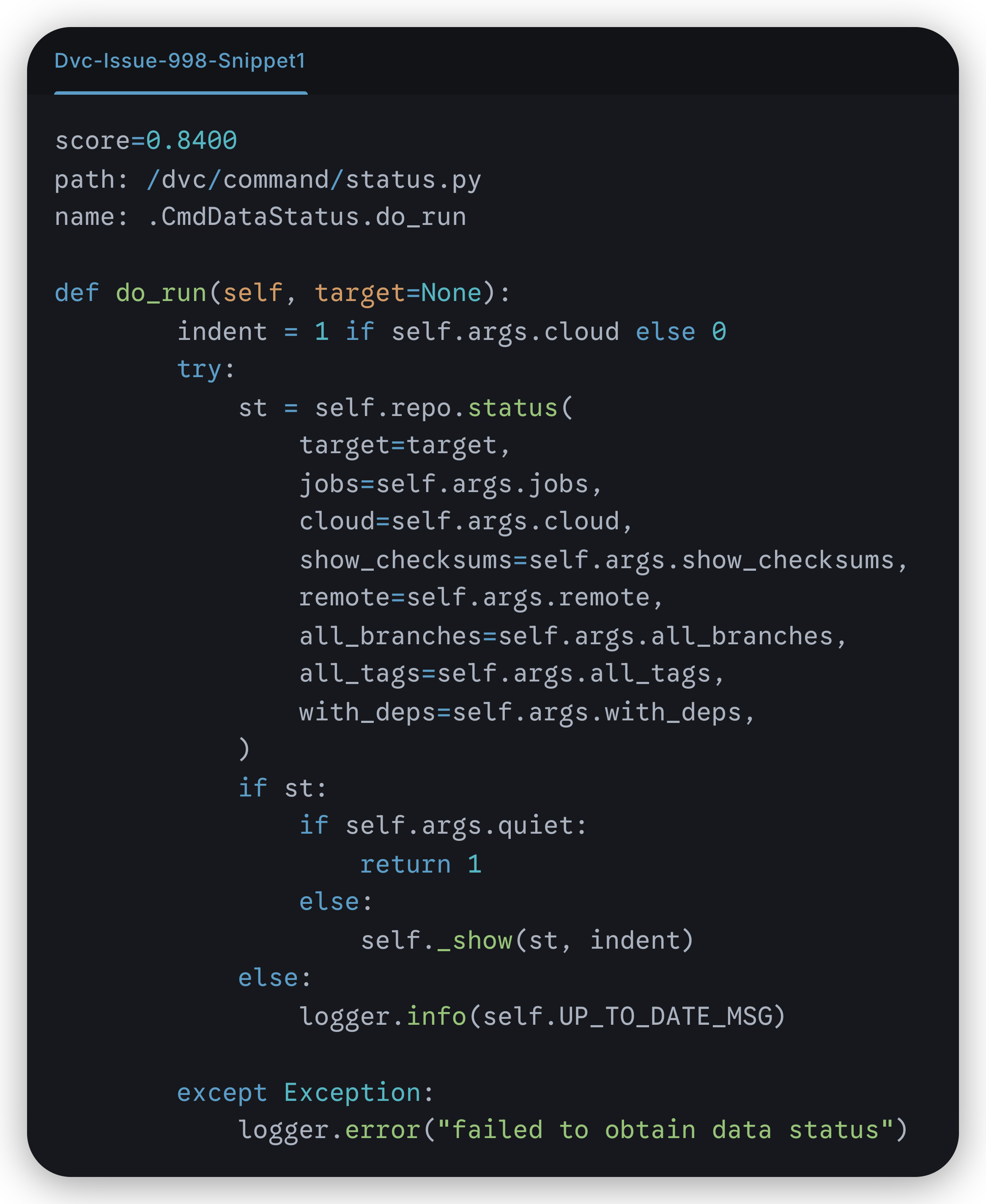}
\caseimg{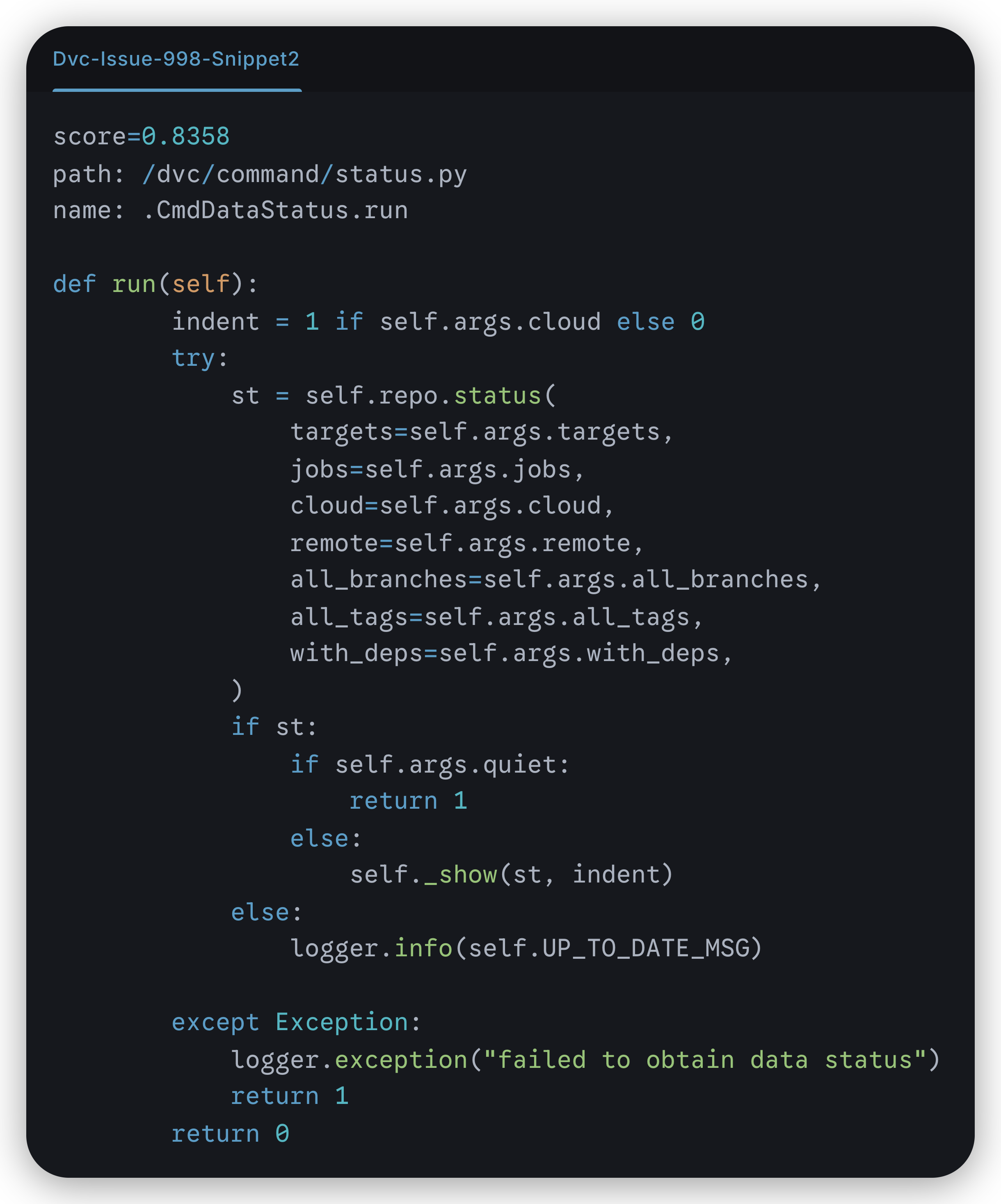}
\caseimg{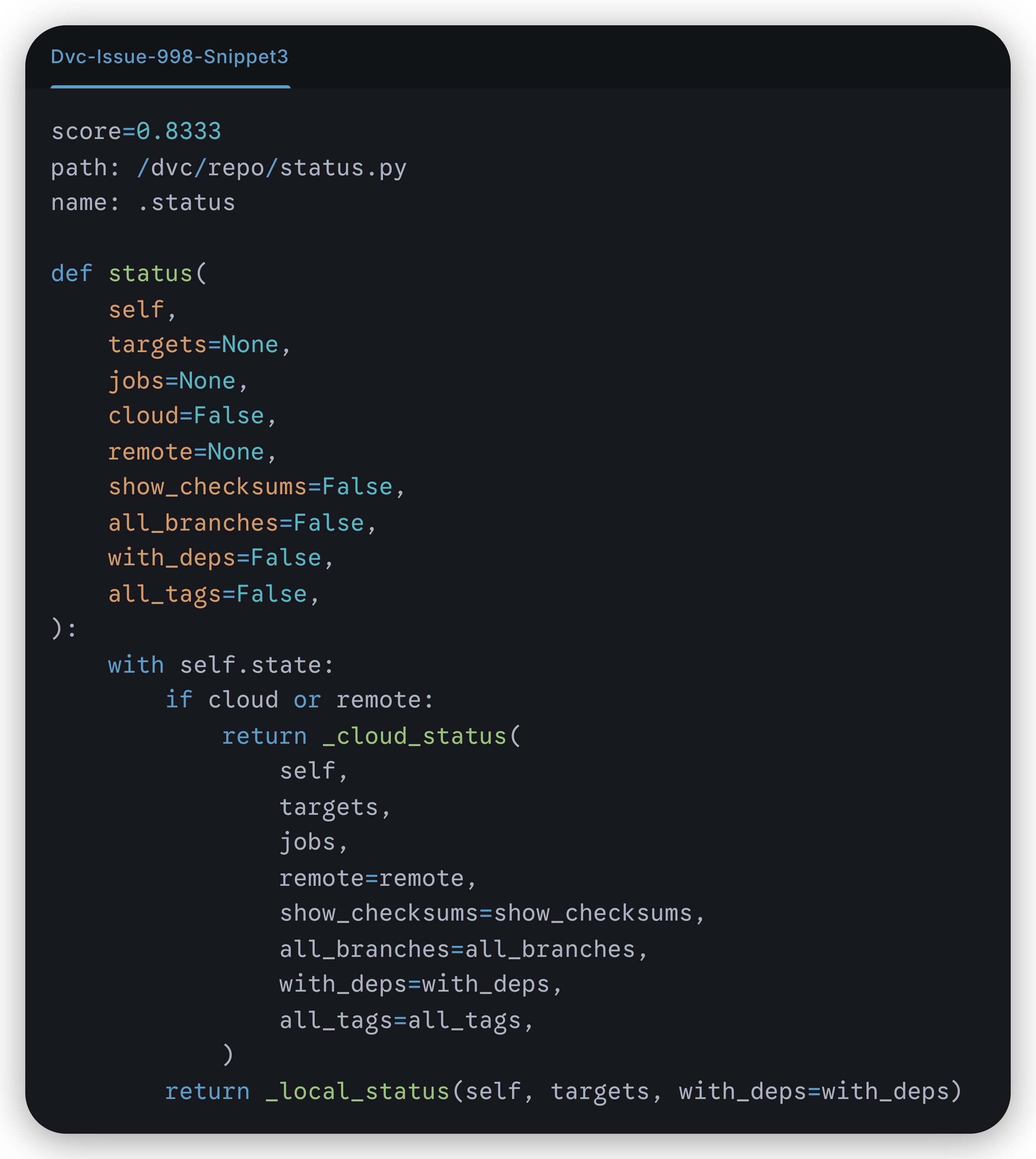}
\repocap{\textbf{DVC.} Issue \#998 and its top-ranked retrieved snippets (sorted by similarity).}

\caseimg{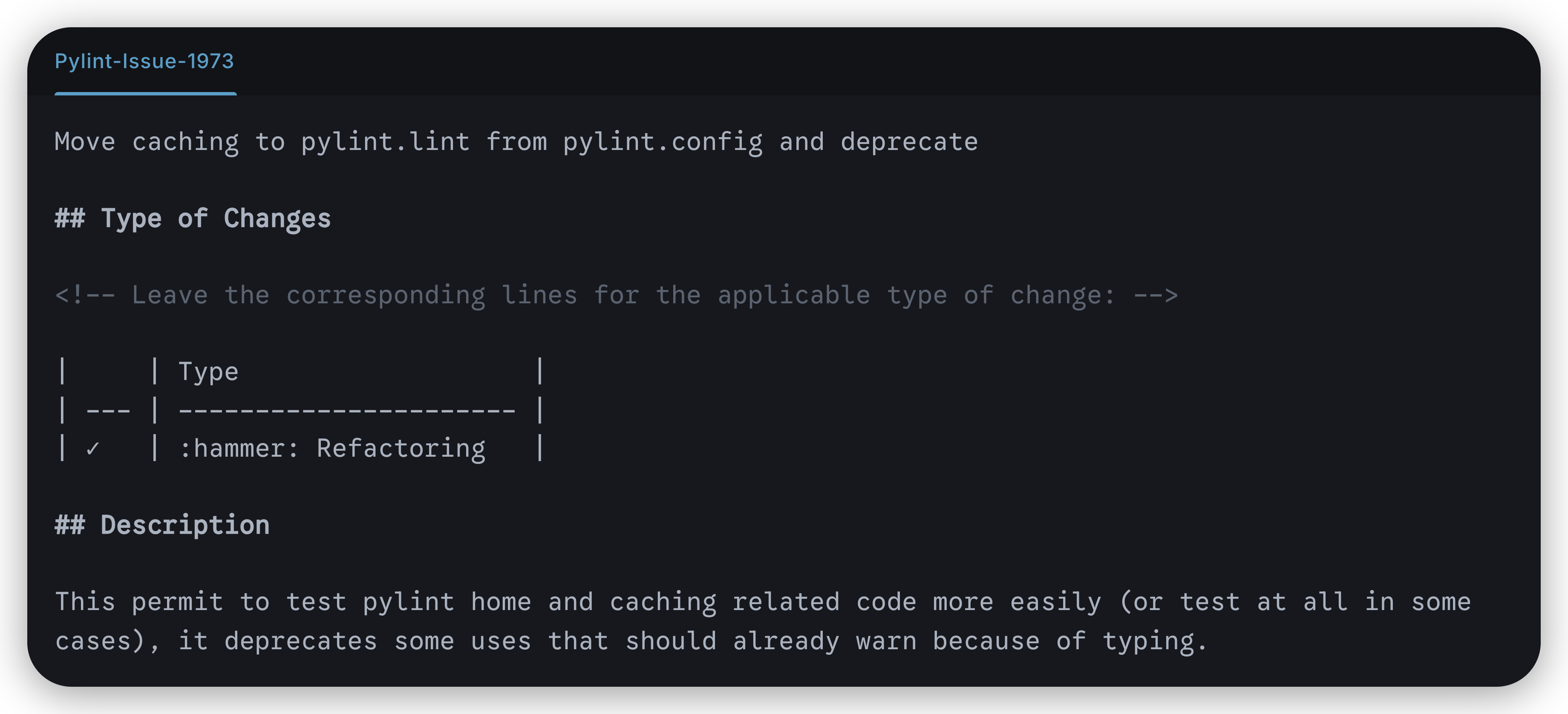}
\caseimg{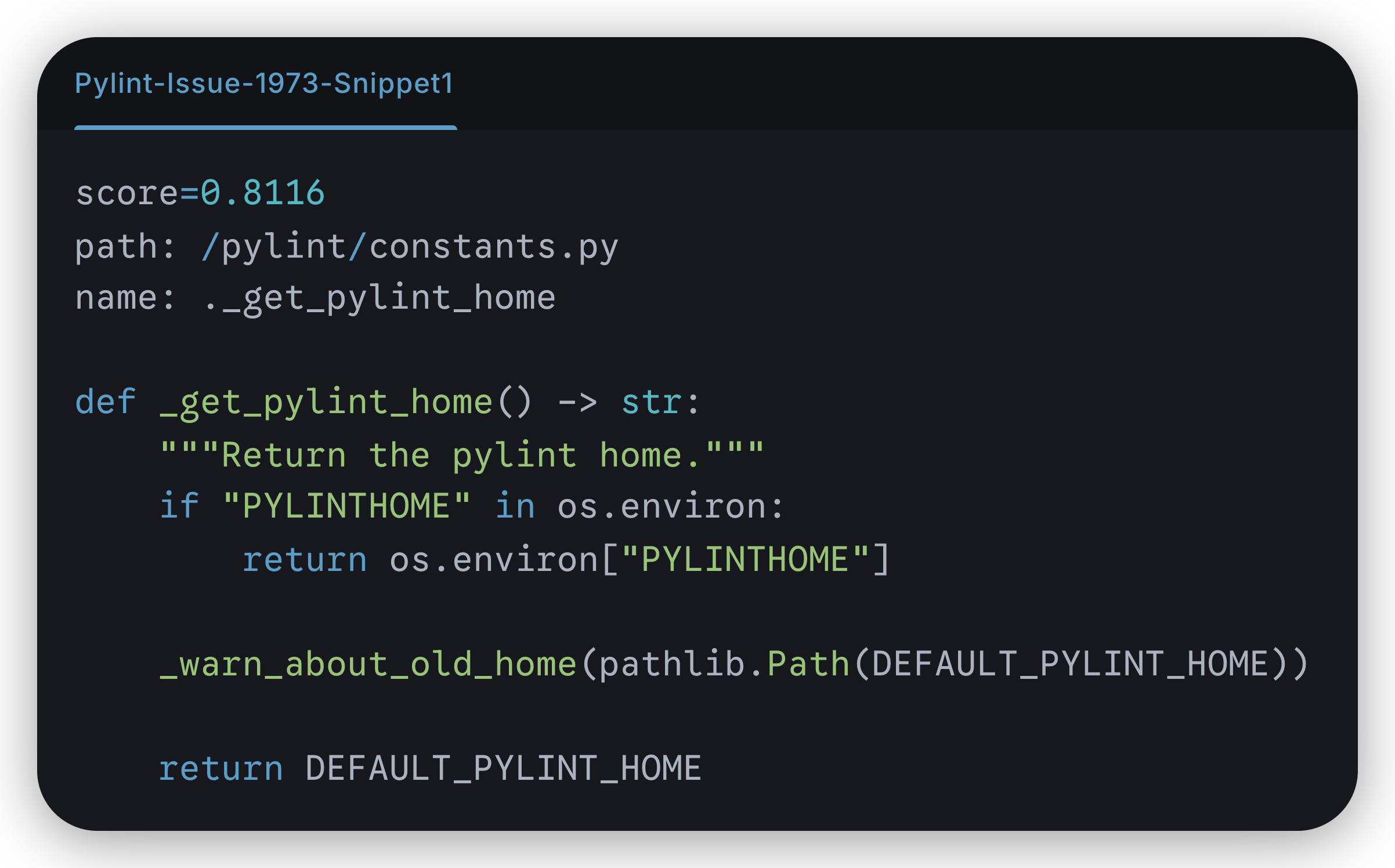}
\caseimg{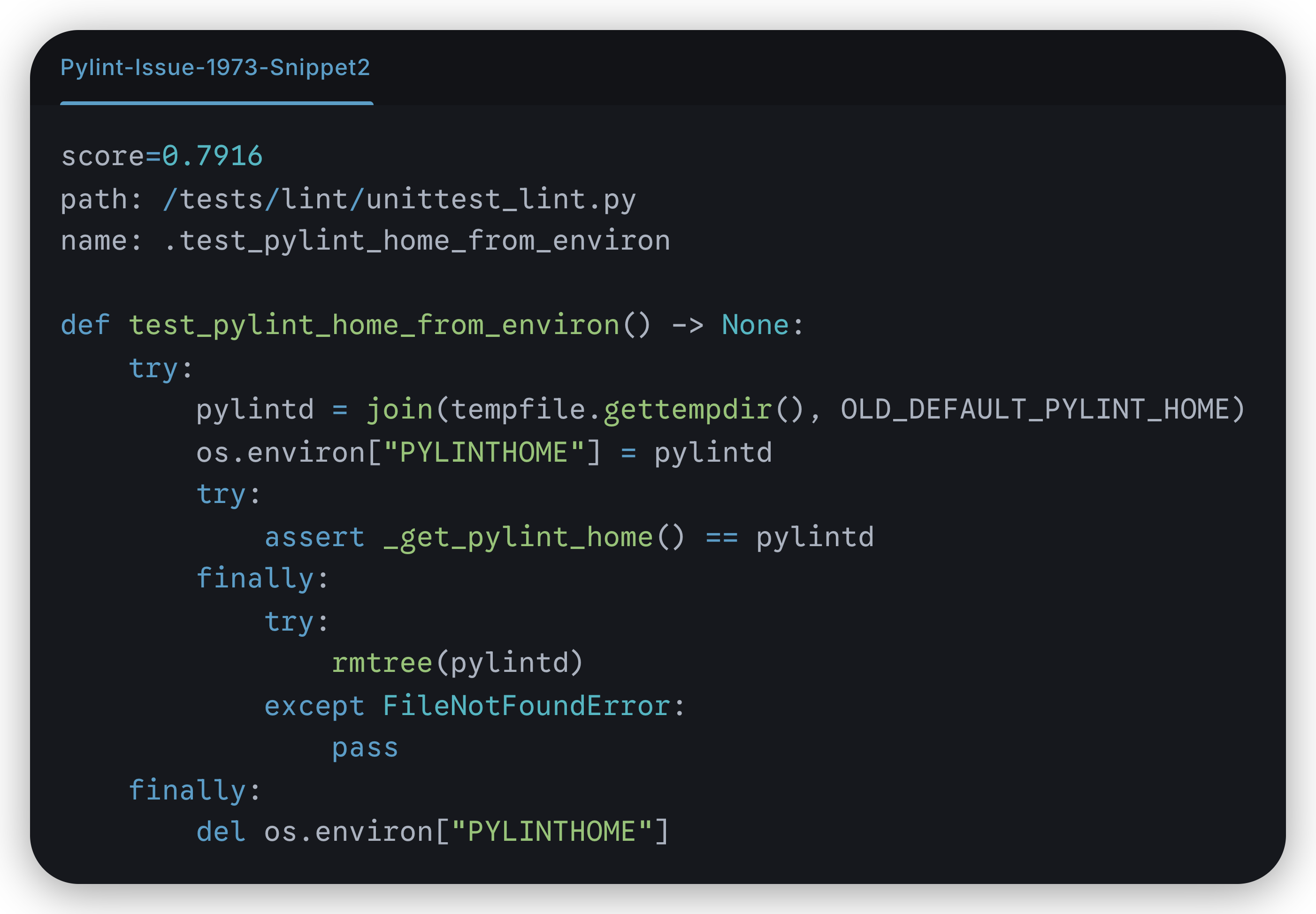}
\repocap{\textbf{Pylint.} Issue \#1973 and its top-ranked retrieved snippets (sorted by similarity).}

\captionsetup{type=figure}
\captionof{figure}{\textbf{Qualitative similarity case studies across repositories.}
For each repository, we present a real issue report followed by several code snippets with high similarity to the issue-derived queries.
This qualitative evidence complements the quantitative AUC results by illustrating what the similarity feature actually retrieves: top-ranked snippets are typically topically and semantically aligned with the bug report, providing an interpretable retrieval-oriented prior for downstream graph learning.}
\label{fig:similarity-case-studies}
\endgroup

\section{Feature Validation}\label{app:featval}
\begin{figure}[t]
  \centering
  \includegraphics[width=\linewidth]{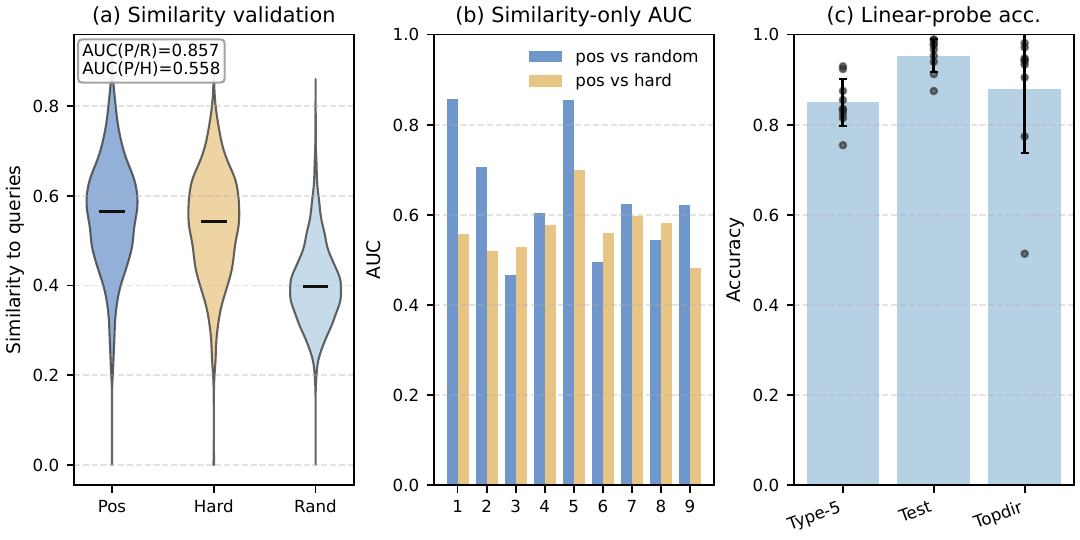}
  \caption{\textbf{Feature validation for similarity and embedding semantics.}
  (a) \textbf{Per-issue example (astropy):} similarity distributions for positive (edited) nodes, hard negatives (same-file but unedited), and random negatives, with AUC inset.
  (b) 
  \textbf{Aggregate similarity-only ranking (9 repos):} AUC for separating positives from random negatives vs.\ hard negatives using similarity scores alone.
  (c) \textbf{Linear probe on frozen embeddings (9 repos):} test accuracy (mean$\pm$std over repos, dots are per-repo) on three representative semantic/structural tasks (node kind, test vs.\ non-test, and top-level module
  ).}
  \label{fig:summary9}
\end{figure}

We validate feature construction on \textbf{9 repositories}:
\texttt{astropy}, \texttt{dvc}, \texttt{ipython}, \texttt{pylint}, \texttt{scipy}, \texttt{sphinx}, \texttt{streamlink}, \texttt{xarray}, and \texttt{geopandas}.
Figure~\ref{fig:summary9} reports our main \emph{feature-validation} results (aggregate across repositories, plus one representative per-issue example). We keep the detailed protocols and additional qualitative evidence in Appendix~\ref{app:sim-val}--\ref{app:cases}.

\paragraph{What this validates.}
\textbf{Similarity as a feature (Fig.~\ref{fig:summary9}a--b).}
We score each node by $s(v)=\max_i \mathbf{h}_v^\top\mathbf{h}_{q_i}$ and evaluate whether similarity alone can rank \emph{patched} (edited) code nodes above negatives.
We quantify this with ROC-AUC on two pairings: positives vs.\ random negatives and positives vs.\ hard negatives, where hard negatives are unedited class/function nodes in the same modified file(s). As expected, similarity strongly separates positives from random negatives (high AUC), while separation against hard negatives is more challenging because those nodes share file-level topical context by construction. \textbf{Embedding semantics (Fig.~\ref{fig:summary9}c).} We additionally train linear probes (single-layer classifiers) on frozen node embeddings and obtain high test accuracy on multiple structural/semantic tasks (node kind, test vs.\ non-test, and top-level module), indicating that embeddings preserve meaningful signals beyond surface lexical overlap. The encoder outputs near unit-norm vectors, which keeps similarity scores numerically stable across repositories (Appendix Fig.~\ref{fig:node-text-length-9repos}).

\section{Method details and diagnostics}
\label{app:method-details}

The overarching goal of anchor node and query augmentation is to convert an unstructured issue report into \emph{graph-addressable} signals: a compact set of candidate code nodes that can serve as a task-conditioned entry point for downstream components (retrieval, reranking, or query-aware graph learning).
The specific methodology is largely inspired by Code Graph Models (CGM)~\citep{tao2025code}.
Importantly, we treat the anchor node as a \emph{task-specific interface} rather than a permanent graph rewrite: the repository graph remains unchanged, while per-issue connectivity is stored as artifacts that are easy to cache, inspect, and reproduce.

\paragraph{Rewriter outputs as structured retrieval cues.}
Given an issue report, the Rewriter produces two complementary structured views.
The \textbf{Extractor} output lists concrete code entities (especially file paths) and a small set of meaningful keywords, encouraging precise lexical grounding when identifiers are explicitly mentioned.
The \textbf{Inferer} output generates up to five search-style queries as complete sentences, encouraging semantic generalization when reports describe behavior without naming specific symbols.
We enforce a delimiter-based output schema (Appendix~\ref{app:prompts}) so that outputs are machine-readable and robust to minor formatting drift.

\paragraph{Lexical anchors (Extractor channel).}
Entities and keywords are matched against node strings (node \texttt{name} and, for file/module nodes, node \texttt{path}) using RapidFuzz\footnote{\url{https://github.com/rapidfuzz/RapidFuzz}}.
For each extracted item we keep the top-3 matches and take the union as $P_{\text{ext}}$, stored as \texttt{extractor\_anchor\_nodes}.
This channel is typically high-precision when the report contains explicit identifiers, but it can under-cover purely behavioral descriptions; this motivates adding a semantic channel.

\paragraph{Semantic anchors (Inferer channel).}
Rewritten queries are embedded by Qwen3-Embedding-8B served through vLLM~\citep{kwon2023vllm} and cached in SafeTensors format for efficient reuse.
We perform dense similarity search using FAISS~\citep{johnson2019billion}; for each query we retrieve top-$k$ nearest nodes and store them as \texttt{inferer\_anchor\_nodes} (a list of $m$ top-$k$ lists).
In our stored artifacts, inferer anchors are recorded as local indices within the time-sliced node list; for analysis, these indices are mapped back to global node ids so that set operations against ground-truth node ids are well-defined.

\paragraph{Time-consistent retrieval.}
Repository graphs are temporal: nodes may appear and disappear across commits. For each issue we compute an \texttt{issue\_time} index and restrict retrieval to nodes whose lifespan covers that time.
This time slicing enforces causality (no retrieving nodes that did not yet exist) and reduces the candidate pool, which improves both efficiency and interpretability of the retrieved anchors.

\paragraph{Unified anchor set and hit/recall metric.}
We define the final prediction set as $P=P_{\text{ext}}\cup P_{\text{inf}}$, where $P_{\text{inf}}$ denotes the semantic anchors mapped to global ids.
To quantify how well anchors cover the ground-truth region, we report per-issue hit/recall: $\mathrm{hit}=\frac{|P\cap G|}{|G|}$,
where $G$ is either (i) modified file nodes or (ii) patched class/function nodes. We average the metric over issues with non-empty $G$.

\begin{figure}[t]
  \centering
  \includegraphics[width=\linewidth]{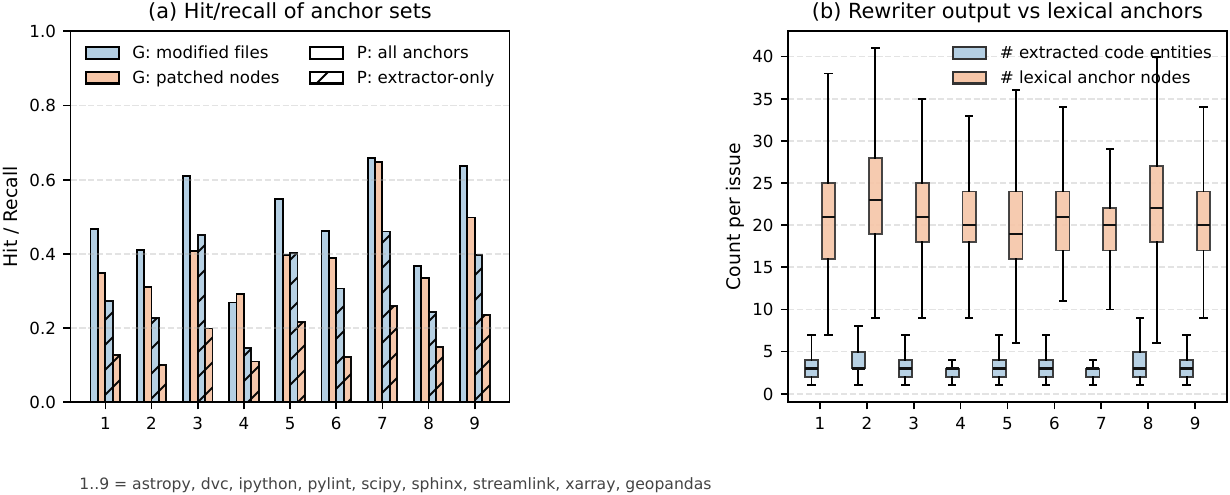}
  \caption{\textbf{Anchor-set hit/recall and output sizes across nine repositories.}
  \textbf{(a) Hit/recall of anchor sets.} For each issue we compute $\mathrm{hit}=|P\cap G|/|G|$ and report the mean over issues with non-empty $G$.
  Solid bars use the full anchor set $P=P_{\text{ext}}\cup P_{\text{inf}}$, while hatched bars use Extractor-only anchors $P=P_{\text{ext}}$.
  Colors indicate the ground-truth set $G$: modified file nodes (blue) and patched class/function nodes (orange).
  \textbf{(b) Output sizes.} Boxplots show the distribution of extracted code entities (Rewriter/Extractor output) and the resulting number of lexical anchor nodes after fuzzy matching.
  Together, the panels quantify both effectiveness (coverage via recall) and practicality (compact interface size) of anchor node augmentation.}
  \label{fig:anchor-hit}
\end{figure}

Figure~\ref{fig:anchor-hit} provides a compact diagnostic of anchor quality.
Using \textbf{all anchors} achieves an issue-count weighted mean hit/recall of $\approx$0.467 on modified files and $\approx$0.380 on patched nodes, whereas \textbf{Extractor-only} anchors achieve $\approx$0.306 and $\approx$0.161, respectively.
Beyond the numbers, the qualitative interpretation is consistent: the Extractor channel excels when the issue names identifiers, while the Inferer channel compensates when the report is descriptive and behavioral; their union improves robustness across repositories and writing styles.

\section{Rewriter prompt templates}
\label{app:prompts}
Following the Rewriter design in CGM~\citep{tao2025code}, we use two prompts per issue: an \textbf{Extractor} prompt (entities + keywords) and an \textbf{Inferer} prompt (search queries).
The Extractor prompt is optimized for \emph{precise grounding}: it asks the model to identify concrete code entities---especially file paths---and to distill a few meaningful keywords.
The Inferer prompt is optimized for \emph{semantic coverage}: it asks the model to express the issue as up to five repository-scoped search queries written as complete sentences.
Both prompts enforce explicit delimiter blocks so that a deterministic parser can reliably recover lists; this is crucial because the Retriever consumes these outputs directly (Appendix~\ref{app:method-details}).
Figure~\ref{fig:extractor-inferer-prompts} shows the full templates used in our implementation.

\begingroup
\graphicspath{{}}
\setlength{\parindent}{0pt}
\setlength{\parskip}{0pt}

\newcommand{\caseimg}[1]{%

\begin{center}

\includegraphics[width=0.8\linewidth]{#1}

\end{center}

\vspace{-0.35em}%

}
\newcommand{\repocap}[1]{%
  \vspace{0.15em}\par{\small\textit{#1}}\par\vspace{0.8em}%
}

\caseimg{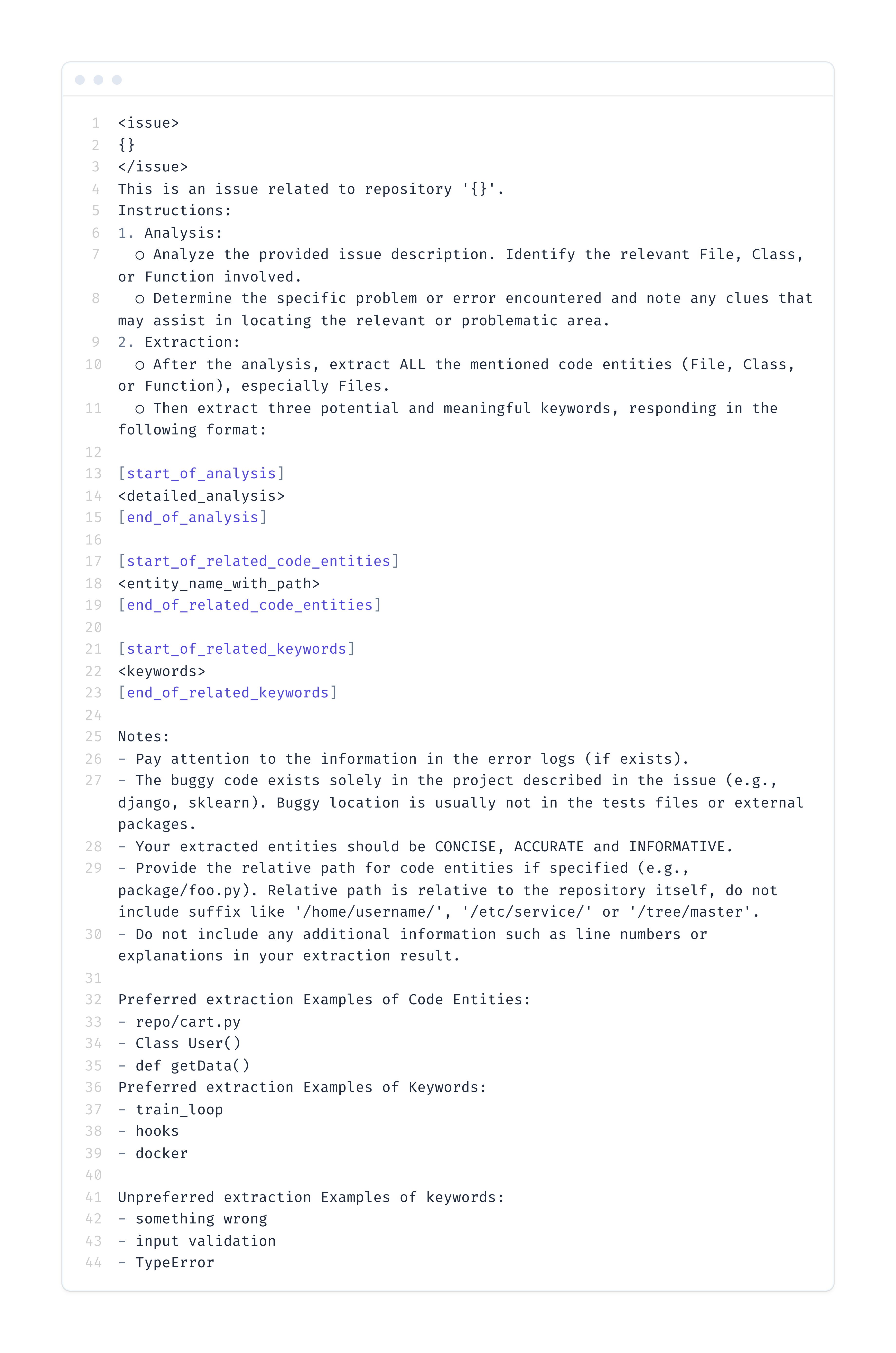}
\repocap{\textbf{Extractor} prompt (entities + keywords).}

\caseimg{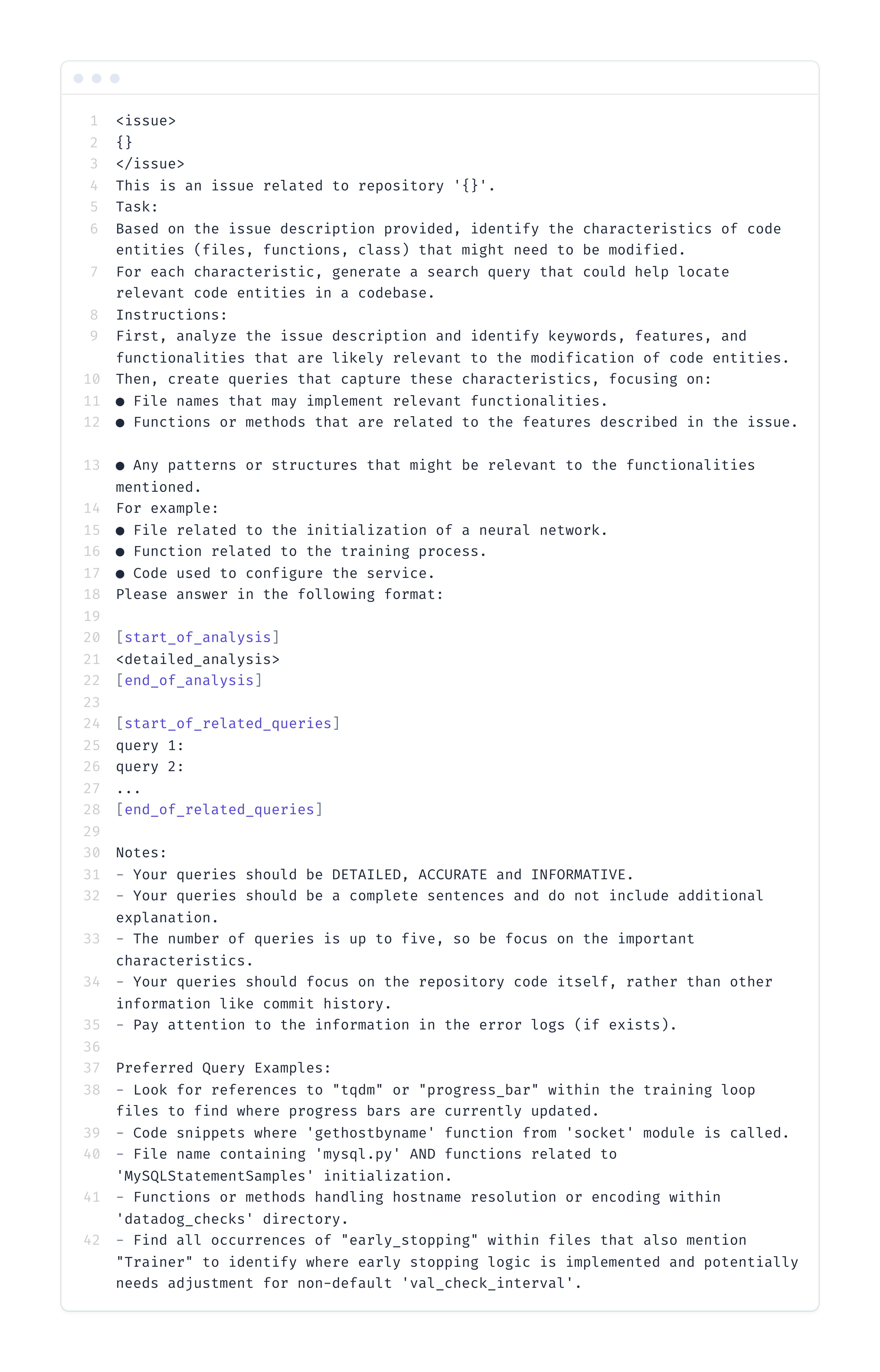}
\repocap{\textbf{Inferer} prompt (search queries)}

\captionsetup{type=figure}
\captionof{figure}{\textbf{Prompt templates used by the Rewriter.}
The \textbf{Extractor} prompt requests (i) a brief analysis and (ii) structured extraction of all mentioned code entities (especially file paths) and a small set of keywords, returned within explicit delimiter blocks.
The \textbf{Inferer} prompt requests repository-scoped search queries (up to five) phrased as complete sentences, again within delimiter blocks.
This strict output schema decouples the LLM's free-form reasoning from the machine-readable signals required by the Retriever and makes the pipeline robust and reproducible.}
\label{fig:extractor-inferer-prompts}
\endgroup

By rewriting the original issue content, we reduce noise (long discussions, environment logs) and convert free-form descriptions into retrieval-friendly signals.
To make the intermediate representation concrete, Figure~\ref{fig:rewrite-examples} shows real rewriting outputs produced by the prompts above and illustrates what is fed into the Retriever in Appendix~\ref{app:method-details}.

\section{Rewrite examples}
\label{app:rewrite-examples}

We provide three representative cases (Astropy, DVC, and Pylint) to illustrate the structure and content of the rewritten outputs.
Each example includes (i) extracted code entities/keywords that enable precise lexical grounding and (ii) search-style queries that enable semantic retrieval, bridging the gap between natural language issue reports and graph-addressable code nodes.

\begingroup
\graphicspath{{}}
\setlength{\parindent}{0pt}
\setlength{\parskip}{0pt}

\newcommand{\caseimg}[1]{%

\begin{center}

\includegraphics[width=0.8\linewidth]{#1}

\end{center}

\vspace{-0.35em}%

}
\newcommand{\repocap}[1]{%
  \vspace{0.15em}\par{\small\textit{#1}}\par\vspace{0.8em}%
}

\caseimg{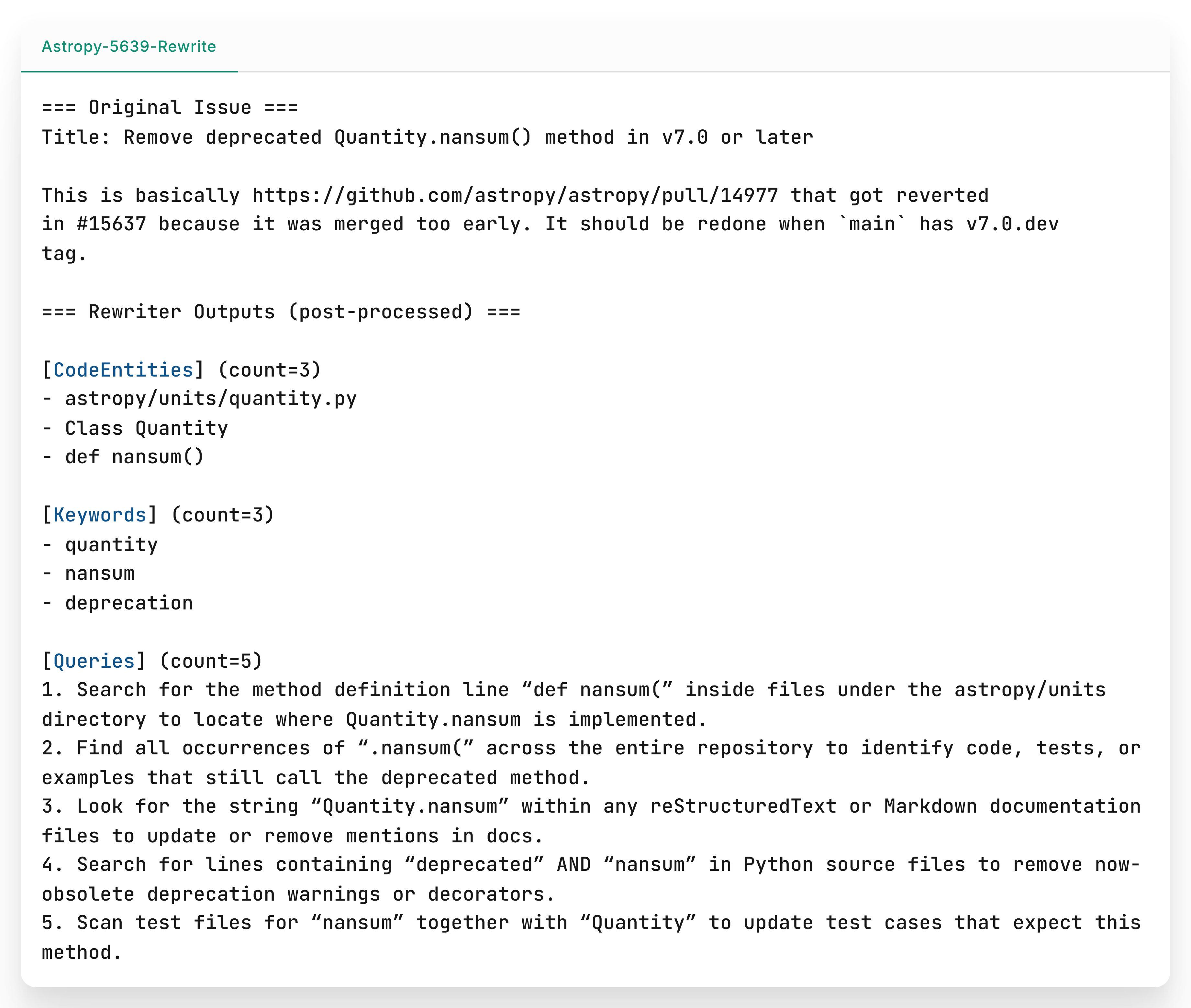}
\repocap{Astropy-5639-Rewrite}

\caseimg{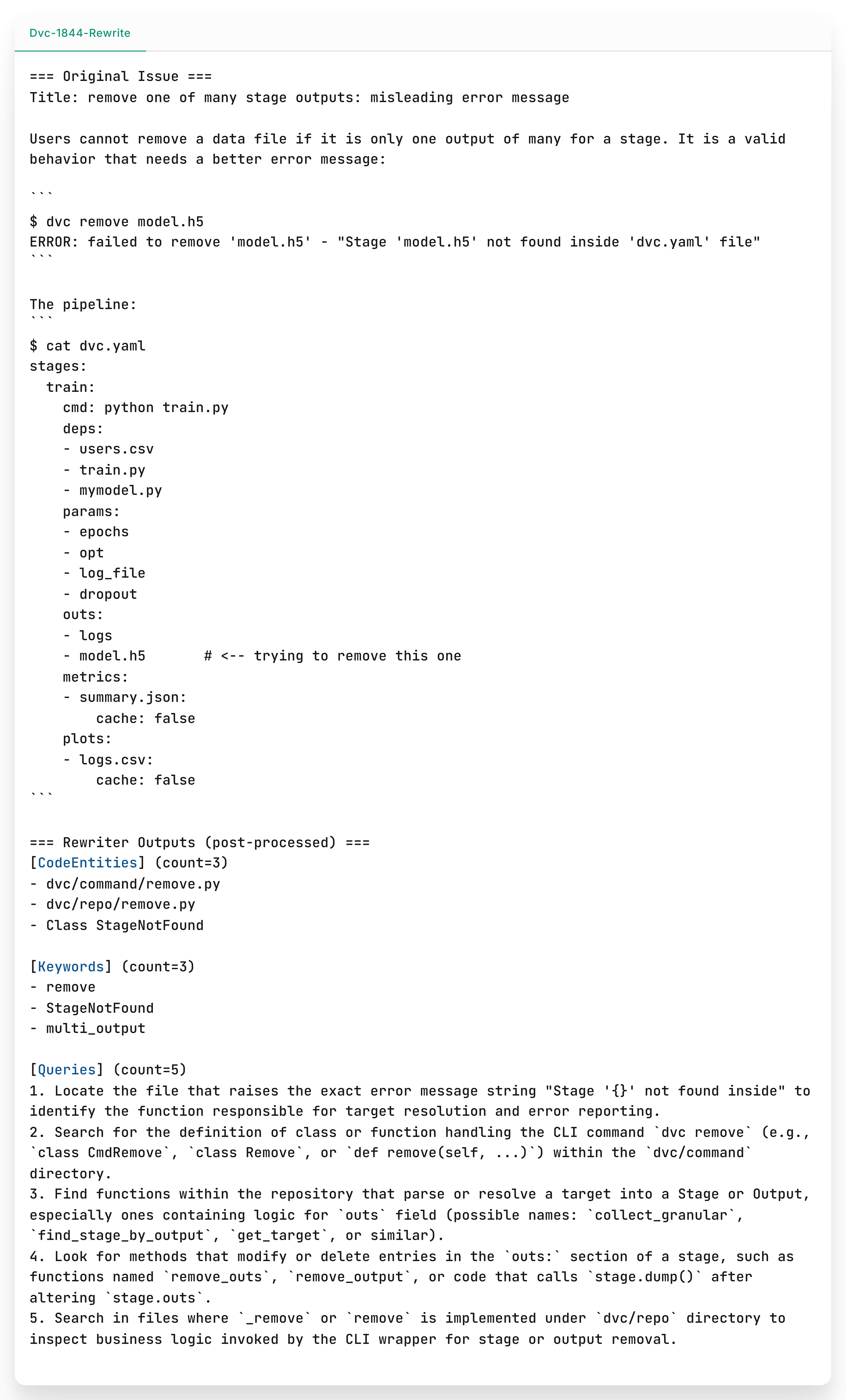}
\repocap{Dvc-1844-Rewrite}

\caseimg{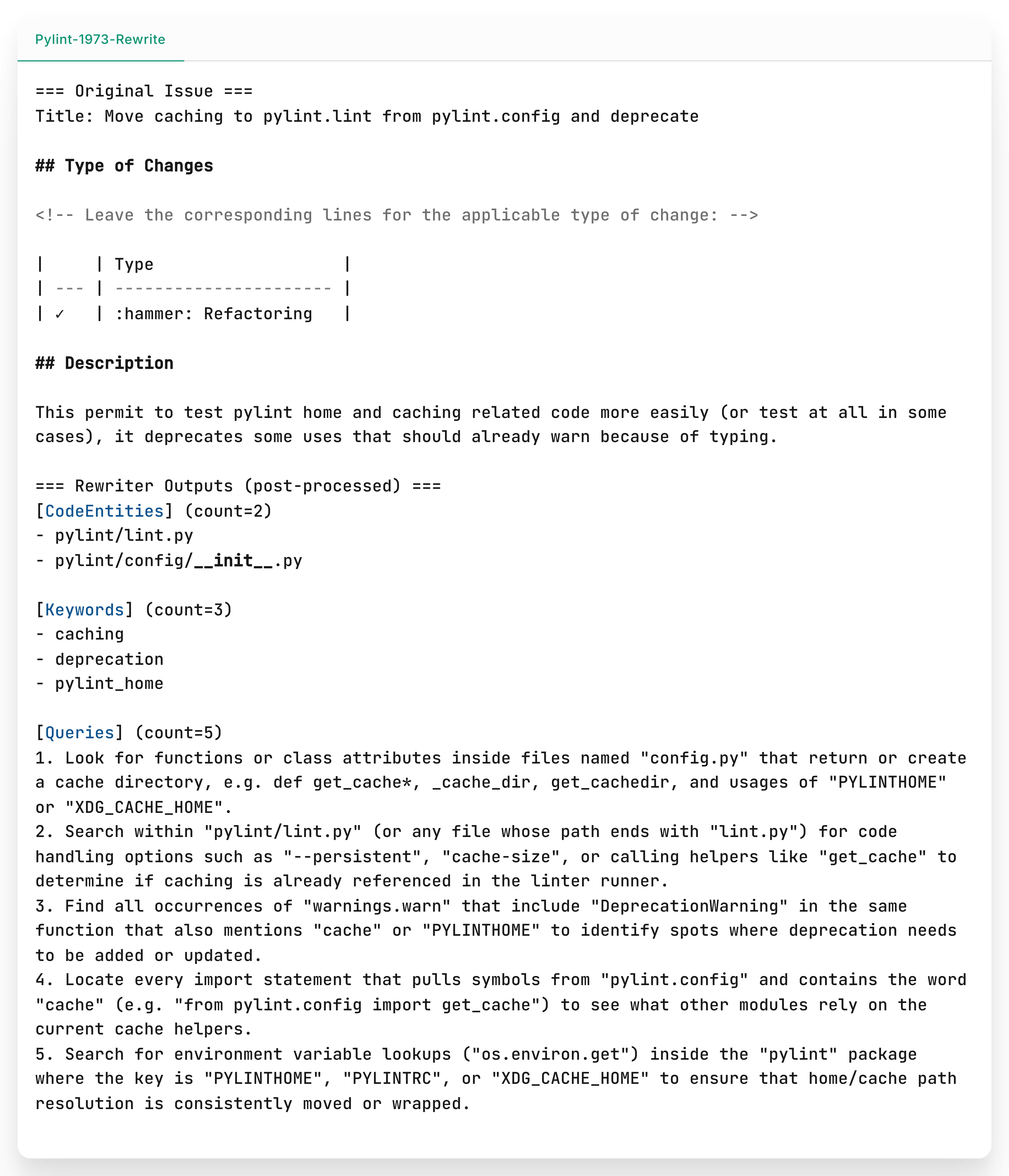}
\repocap{Pylint-1973-Rewrite}

\captionsetup{type=figure}
\captionof{figure}{\textbf{Examples of Rewriter outputs (Extractor + Inferer) across repositories.}
For each issue, the Rewriter produces a concise, structured representation: a list of code entities (e.g., file paths and referenced symbols), a small set of keywords, and up to five search-style queries phrased as complete sentences.
Entities/keywords are used for fuzzy lexical anchoring over node names and paths, while queries are embedded for semantic similarity search.
These intermediate artifacts make the subsequent retrieval step interpretable and enable reproducible, task-conditioned graph access.}
\label{fig:rewrite-examples}
\endgroup

\section{Temporal Anchors}\label{app:Tanchor}

\paragraph{Motivation.}
Repository-level bug fixing is often path-dependent: recently edited modules are more likely to be edited again, and entities co-touched in a short window frequently co-occur in subsequent fixes.
These dynamics live in the commit history rather than in the static snapshot graph.
A natural approach is to retrieve temporal candidates and inject them into the reranker subgraph, but injection is brittle under a fixed budget: larger candidate lists improve recall yet quickly introduce many irrelevant nodes and edges.
More importantly, an ``inject then rerank'' pipeline can look like a candidate-source swap.
We instead make temporal signals change the \emph{graph reasoning core}.

\paragraph{Overview.}
We first train an issue-conditioned temporal retriever to output a node prior $\pi_q(v)$ under strict no-future history (Stage~I).
At inference, $\pi_q$ guides where to expand the reranker subgraph (Stage~II) and is also converted into residual edge gates that modulate GAT message passing (Stage~IV), while keeping the evaluation protocol unchanged.

\subsection{Issue-Conditioned Temporal Prior (\textsc{GET}: Global Event Transformer Retriever)}\label{sec:getv2-corechange5}

\paragraph{Retrieval objective aligned to bug localization.}
Instead of training a temporal model for generic link prediction, we define an issue-conditioned retrieval task:
\begin{equation}
(t_{bug},\mathcal{A}(q),\mathcal{H}_{\le t_{bug}})\ \Rightarrow\ \pi_q(v)\ \ \text{for nodes } v\in V(t_{bug}),
\end{equation}
with supervision from the patched set $\mathcal{G}(q)$.
The retriever is trained to rank patch-related nodes higher.

\paragraph{No-future candidate pool.}
Given anchors $\mathcal{A}(q)$, we build a candidate pool $\mathcal{C}(q)$ by sampling historical neighbors using only interactions $\tau\le t_{bug}$.
To obtain a stable pool under truncation, we rank candidates by anchor support and recency:
\begin{equation}
\mathrm{support}(v)=\sum_{a\in\mathcal{A}(q)}\mathbb{I}[v\in \mathcal{N}_{hist}(a,t_{bug})],\quad
\end{equation}
\begin{equation}
\mathrm{recency}(v)=\max_{a\in\mathcal{A}(q)}\max\{\tau\mid (a,v,\tau)\in\mathcal{H}_{\le t_{bug}}\},
\end{equation}
and keep top candidates sorted by $(\mathrm{support},\mathrm{recency})$.

\paragraph{Temporal Transformer scoring and loss.}
For each anchor $a\in\mathcal{A}(q)$ we sample a historical neighbor sequence $\{(n_j,\tau_j)\}_{j=1}^L$ and encode each token with a trainable node embedding plus a time encoding over $\Delta t=t_{bug}-\tau_j$.
A Transformer encoder yields anchor vectors; we average them to form an issue vector $\mathbf{z}_q$ and score a candidate node $v$ by cosine similarity:
\begin{equation}
s_q(v)=\cos\bigl(\mathbf{W}_q\mathbf{z}_q,\ \mathbf{W}_v\mathbf{z}_v\bigr).
\end{equation}
We optimize a pairwise ranking loss (implemented with \texttt{softplus}):
\begin{equation}
\mathcal{L}_{retr}(q)=\frac{1}{|\mathcal{P}(q)|\,|\mathcal{N}(q)|}\sum_{g\in\mathcal{P}(q)}\sum_{n\in\mathcal{N}(q)}
\mathrm{softplus}\bigl(s_q(n)-s_q(g)+m\bigr),
\end{equation}
where $\mathcal{P}(q)=\mathcal{G}(q)\cap \mathcal{C}(q)$ are positives found in the pool and $\mathcal{N}(q)$ are sampled negatives.
At inference, the retriever outputs a sparse prior $\pi_q(v)$ for each issue.

\subsection{Prior-Guided Routing for Subgraph Construction}\label{sec:routing-corechange5}

\paragraph{Routing under a fixed budget.}
Let $V_q^{base}$ be the standard GREPO anchor subgraph.
We use $\pi_q$ to decide where to spend additional subgraph budget rather than unioning a long candidate list.
Concretely, we take the top-$R$ prior nodes as seeds,
\begin{equation}
S(q)=\mathrm{TopR}_{v\in V(t_{bug})}\ \pi_q(v),
\end{equation}
expand their neighborhoods on the snapshot graph with hop $H$ and cap $B_{exp}$,
\begin{equation}
V^{exp}_q=\mathrm{Extract}(G(t_{bug}),S(q);\;H,\;B_{exp}),
\end{equation}
and form the reranker node set
\begin{equation}
V_q=(V^{base}_q \cup V^{exp}_q)\cap V(t_{bug}).
\end{equation}
In our strongest setting, we keep \textsc{no-inject}: the prior affects the reranker through routing/expansion and edge gating, without unioning the full Top-$N$ candidate list.

\subsection{Query-Aware GAT Reranker}\label{sec:reranker-corechange5}

\paragraph{Node features.}
The reranker operates on $G_q=(V_q,E_q)$ and predicts a relevance score $r_q(v)$ for each $v\in V_q$.
We concatenate query--node similarity, an anchor indicator, and the temporal prior score when available:
\begin{equation}
\mathbf{h}_v^{(0)}=\bigl[\ \mathrm{Sim}(v,q)\ ;\ a_v\ ;\ \sigma(\pi_q(v))\ \bigr].
\end{equation}
Optionally, we soften the anchor indicator with the prior (\texttt{score\_into\_anchor}): $a_v\leftarrow \max(a_v,\sigma(\pi_q(v)))$.

\paragraph{Training objective.}
We train with the same ranking objective as GREPO: patched nodes in $\mathcal{G}(q)$ are positives and other nodes in $V_q$ are negatives.
The reranker is trained on 86 repositories and evaluated on 9 held-out repositories under the same filtered issue splits.

\subsection{Residual Edge Gating for Structure-Level Fusion}\label{sec:edge-gate-corechange5}

\paragraph{From node priors to edge gates.}
We convert the node prior into per-edge gates that modulate message passing.
For an edge $e=(u,v)$ we aggregate endpoint priors by
\begin{equation}
\eta_q(e)=\max\{\sigma(\pi_q(u)),\sigma(\pi_q(v))\},
\end{equation}
and define a residual gate (``no signal'' stays neutral):
\begin{equation}
g_q(e)=1+\alpha\cdot\Bigl(\mathrm{sigmoid}\bigl(\gamma(\eta_q(e)-b)\bigr)-\tfrac{1}{2}\Bigr)\cdot 2,
\end{equation}
where $\alpha$ controls the strength, $\gamma$ controls the slope, and $b$ is a bias.
To prevent global damping when the temporal prior has no overlap with the extracted subgraph, we enforce: if both endpoints have zero prior, then $g_q(e)=1$; if all nodes in $V_q$ have zero prior, then $g_q(e)=1$ for all edges.

\paragraph{Gated message passing.}
We pass $g_q(e)$ as an edge weight (or attention bias) to the query-aware GAT so that messages on edges with higher temporal support are amplified.
This changes the reranker's message passing without altering the evaluation protocol.

\subsection{Optional: Query-Conditioned Virtual Edges}\label{sec:virtual-edges-corechange5}

\paragraph{Compact connectivity augmentation.}
As an alternative to importing full neighborhoods, we can attach a small number of high-prior nodes as isolated candidates and connect them to anchors using a new edge type $r_{virt}$:
\begin{equation}
E_q^{virt}=\{(a,c,r_{virt}) \mid a\in\tilde{\mathcal{A}}(q),\ c\in \tilde{\mathcal{C}}(q)\}\cup\text{reverse},
\end{equation}
where $\tilde{\mathcal{C}}(q)$ are the top-$K$ candidates by $\pi_q$ (capped) and $\tilde{\mathcal{A}}(q)$ are selected anchors.
Virtual edges provide a low-noise ablation that still makes the reasoning graph issue-conditioned.

\begin{algorithm}[t]
\caption{Core-changed inference for bug localization.}\label{alg:corechange5-infer}
\small
\begin{algorithmic}[1]
\Require Issue $q$, bug time $t_{bug}$, snapshot graph $G(t_{bug})$, anchors $\mathcal{A}(q)$, dumped temporal prior $\pi_q$.
\Require Budgets: base extraction $(k,B)$, expansion $(H,B_{exp})$, seed size $R$.
\State $V^{base}_q \gets \mathrm{Extract}(G(t_{bug}),\mathcal{A}(q);k,B)$
\State $S(q)\gets \mathrm{TopR}\ \pi_q(v)$
\State $V^{exp}_q \gets \mathrm{Extract}(G(t_{bug}),S(q);H,B_{exp})$
\State $V_q \gets (V^{base}_q \cup V^{exp}_q)\cap V(t_{bug})$
\State Build node features $\mathbf{h}^{(0)}_v=[\mathrm{Sim}(v,q);a_v;\sigma(\pi_q(v))]$ for $v\in V_q$
\State Build residual edge gates $g_q(e)$ from endpoint priors (neutral when no temporal signal)
\State Run query-aware GAT on $G_q=(V_q,E_q)$ with edge weights/bias $g_q(e)$
\State \Return $\mathrm{TopK}(q)$ by reranker scores $r_q(v)$
\end{algorithmic}
\end{algorithm}

\begin{table}[t]
\caption{Key hyperparameters for our best core-changed run (function-level).}\label{tab:corechange5-hparams}
\setlength{\tabcolsep}{3pt}
\small
\begin{tabular}{ll}
\toprule
Component & Setting \\
\midrule
Reranker backbone & GAT (4 layers, 4 heads, hidden dim 128) \\
Optimization & epochs=10, lr=$10^{-4}$, weight decay=0 \\
Subgraph extraction & $k{=}1$ hop around anchors, max $B{=}80$k nodes \\
Temporal prior input & dumped GETv2 candidates, add\_topn=2000 (no-inject) \\
Routing / expansion & expand\_hops=1, expand\_topk=150, expand\_max\_size=80k \\
Residual edge gate & style=residual\_sigmoid, $\alpha{=}1.0$, $\gamma{=}2.0$, $b{=}0.0$, mode=max \\
\bottomrule
\end{tabular}
\end{table}

\section{Experimental Details}\label{app:exp}

All experiments are conducted on a Linux server equipped with one NVIDIA RTX 4090 GPU. We use PyTorch and PyTorch Geometric to implement the graph neural networks (GNNs). The large language model (LLM) baselines are implemented using their official codebases. We employ the AdamW optimizer with the following hyperparameters: batch size = 1, learning rate = 1e-4, weight decay = 0, number of training epochs = 10, and subgraph hop  = 1.

\section{The Repositories in GREPO }\label{app:reponame}
The full repository names of the GREPO benchmark are shown in Table~\ref{tab:grepo_benchmark}.

{
\footnotesize  
\setlength{\tabcolsep}{4pt} 

\begin{longtable}{p{0.22\textwidth} p{0.43\textwidth} p{0.28\textwidth}}
\caption{The full repository names of the GREPO benchmark.} \label{tab:grepo_benchmark} \\

\toprule
\textbf{Repository} & \textbf{Description} & \textbf{URL} \\
\midrule
\endfirsthead

\multicolumn{3}{c}{{\bfseries \tablename\ \thetable{} -- continued from previous page}} \\
\toprule
\textbf{Repository} & \textbf{Description} & \textbf{URL} \\
\midrule
\endhead

\midrule
\multicolumn{3}{r}{{Continued on next page}} \\
\bottomrule
\endfoot

\bottomrule
\endlastfoot

ntc-templates & Multi-vendor network parsing templates & \url{https://github.com/networktocode/ntc-templates} \\
wemake-python-styleguide & The strictest and most opinionated python linter & \url{https://github.com/wemake-services/wemake-python-styleguide} \\
cryptography & Cryptographic recipes and primitives for Python & \url{https://github.com/pyca/cryptography} \\
sphinx & Python documentation generator & \url{https://github.com/sphinx-doc/sphinx} \\
xarray & N-D labeled arrays and datasets & \url{https://github.com/pydata/xarray} \\
ipython & Interactive computing in Python & \url{https://github.com/ipython/ipython} \\
jupyter-ai & Generative AI extension for JupyterLab & \url{https://github.com/jupyterlab/jupyter-ai} \\
keras & Deep learning for humans & \url{https://github.com/keras-team/keras} \\
llama-stack & Composable building blocks for Llama models & \url{https://github.com/meta-llama/llama-stack} \\
pylint & Static code analysis for Python & \url{https://github.com/pylint-dev/pylint} \\
transformers & State-of-the-art Machine Learning for Pytorch/TF/JAX & \url{https://github.com/huggingface/transformers} \\
django & High-level Python Web framework & \url{https://github.com/django/django} \\
matplotlib & Plotting with Python & \url{https://github.com/matplotlib/matplotlib} \\
checkov & Infrastructure as Code (IaC) security scanner & \url{https://github.com/bridgecrewio/checkov} \\
tox & Command line driven CI frontend and test runner & \url{https://github.com/tox-dev/tox} \\
mypy & Optional static typing for Python & \url{https://github.com/python/mypy} \\
transitions & A lightweight, object-oriented state machine & \url{https://github.com/pytransitions/transitions} \\
yt-dlp & A command-line program to download videos & \url{https://github.com/yt-dlp/yt-dlp} \\
mesa & Agent-based modeling framework & \url{https://github.com/projectmesa/mesa} \\
conan & The open-source C/C++ package manager & \url{https://github.com/conan-io/conan} \\
twine & Utility for publishing packages on PyPI & \url{https://github.com/pypa/twine} \\
urllib3 & HTTP library with thread-safe connection pooling & \url{https://github.com/urllib3/urllib3} \\
falcon & The no-nonsense web API framework & \url{https://github.com/falconry/falcon} \\
feature\_engine & Feature engineering package with sklearn-like APIs & \url{https://github.com/feature-engine/feature_engine} \\
filesystem\_spec & A specification for pythonic file-systems & \url{https://github.com/fsspec/filesystem_spec} \\
Flexget & The multi-purpose automation tool for content & \url{https://github.com/Flexget/Flexget} \\
geopandas & Python tools for geographic data & \url{https://github.com/geopandas/geopandas} \\
haystack & Orchestration framework for LLM applications & \url{https://github.com/deepset-ai/haystack} \\
instructlab & Taxonomy-driven model alignment and tuning & \url{https://github.com/instructlab/instructlab} \\
jax & Autograd and XLA for high-performance machine learning & \url{https://github.com/google/jax} \\
kedro & A framework for creating reproducible data pipelines & \url{https://github.com/kedro-org/kedro} \\
litellm & Call all LLM APIs using the OpenAI format & \url{https://github.com/BerriAI/litellm} \\
marshmallow & Simplified object serialization & \url{https://github.com/marshmallow-code/marshmallow} \\
conda & OS-agnostic package and environment manager & \url{https://github.com/conda/conda} \\
llama\_deploy & Deployment tool for LlamaIndex agentic workflows & \url{https://github.com/run-llama/llama_deploy} \\
networkx & Network analysis in Python & \url{https://github.com/networkx/networkx} \\
aider & AI pair programming in your terminal & \url{https://github.com/Aider-AI/aider} \\
aiogram & Asynchronous framework for Telegram Bot API & \url{https://github.com/aiogram/aiogram} \\
ansible-lint & Linter for Ansible playbooks and roles & \url{https://github.com/ansible/ansible-lint} \\
arviz & Exploratory analysis of Bayesian models & \url{https://github.com/arviz-devs/arviz} \\
astroid & A common base representation of python source code & \url{https://github.com/pylint-dev/astroid} \\
astropy & Community Python library for Astronomy & \url{https://github.com/astropy/astropy} \\
attrs & Python classes without boilerplate & \url{https://github.com/python-attrs/attrs} \\
babel & Internationalization utilities & \url{https://github.com/python-babel/babel} \\
beancount & Double-entry bookkeeping computer language & \url{https://github.com/beancount/beancount} \\
beets & Music library management system & \url{https://github.com/beetbox/beets} \\
briefcase & Tools to package Python code as an app & \url{https://github.com/beeware/briefcase} \\
cfn-lint & CloudFormation Linter & \url{https://github.com/aws-cloudformation/cfn-lint} \\
Cirq & Library for creating and running quantum circuits & \url{https://github.com/quantumlib/Cirq} \\
crawlee-python & Reliable web scraping and browser automation & \url{https://github.com/apify/crawlee-python} \\
csvkit & A suite of utilities for converting/working with CSV & \url{https://github.com/wireservice/csvkit} \\
datasets & Access to audio, computer vision, and NLP datasets & \url{https://github.com/huggingface/datasets} \\
dspy & Framework for programming with language models & \url{https://github.com/stanfordnlp/dspy} \\
dvc & Data Version Control for ML projects & \url{https://github.com/iterative/dvc} \\
dynaconf & Configuration management for Python & \url{https://github.com/dynaconf/dynaconf} \\
faststream & Framework for asynchronous services (Kafka/RabbitMQ) & \url{https://github.com/airtai/faststream} \\
flask & A lightweight WSGI web application framework & \url{https://github.com/pallets/flask} \\
fonttools & Library for manipulating fonts & \url{https://github.com/fonttools/fonttools} \\
icloud\_photos\_downloader & Command-line tool to download iCloud Photos & \url{https://github.com/icloud-photos-downloader/icloud_photos_downloader} \\
openai-agents-python & Lightweight framework for multi-agent workflows & \url{https://github.com/openai/openai-agents-python} \\
patroni & Template for PostgreSQL High Availability & \url{https://github.com/patroni/patroni} \\
pipenv & Python Development Workflow for Humans & \url{https://github.com/pypa/pipenv} \\
poetry & Python dependency management and packaging & \url{https://github.com/python-poetry/poetry} \\
privacyidea & Open Source Two Factor Authentication & \url{https://github.com/privacyidea/privacyidea} \\
pvlib-python & Photovoltaic energy modeling & \url{https://github.com/pvlib/pvlib-python} \\
PyBaMM & Python Battery Mathematical Modelling & \url{https://github.com/pybamm-team/PyBaMM} \\
pydicom & Read, modify and write DICOM files & \url{https://github.com/pydicom/pydicom} \\
pyomo & Python Optimization Modeling Objects & \url{https://github.com/Pyomo/pyomo} \\
PyPSA & Python for Power System Analysis & \url{https://github.com/PyPSA/PyPSA} \\
python-control & Systems analysis and design & \url{https://github.com/python-control/python-control} \\
python & The Python programming language (CPython) & \url{https://github.com/python/cpython} \\
python-telegram-bot & Wrapper for the Telegram Bot API & \url{https://github.com/python-telegram-bot/python-telegram-bot} \\
pyvista & 3D plotting and mesh analysis & \url{https://github.com/pyvista/pyvista} \\
qtile & A full-featured, hackable tiling window manager & \url{https://github.com/qtile/qtile} \\
Radicale & A simple CalDAV (calendar) and CardDAV (contact) server & \url{https://github.com/Kozea/Radicale} \\
scipy & Fundamental algorithms for scientific computing & \url{https://github.com/scipy/scipy} \\
scrapy-splash & Scrapy+Splash for JavaScript integration & \url{https://github.com/scrapy-plugins/scrapy-splash} \\
segmentation\_models.pytorch & Semantic segmentation models with pre-trained backbones & \url{https://github.com/qubvel/segmentation_models.pytorch} \\
shapely & Manipulation and analysis of geometric objects & \url{https://github.com/shapely/shapely} \\
smolagents & Minimalist library for agents that think in code & \url{https://github.com/huggingface/smolagents} \\
Solaar & Linux device manager for Logitech devices & \url{https://github.com/pwr-Solaar/Solaar} \\
sqlfluff & A SQL linter and auto-formatter & \url{https://github.com/sqlfluff/sqlfluff} \\
streamlink & CLI utility to pipe streams to video players & \url{https://github.com/streamlink/streamlink} \\
tablib & Format-agnostic tabular dataset library & \url{https://github.com/jazzband/tablib} \\
torchtune & PyTorch native library for LLM fine-tuning & \url{https://github.com/pytorch/torchtune} \\
WeasyPrint & Converts HTML/CSS documents to PDF & \url{https://github.com/Kozea/WeasyPrint} \\

\end{longtable}
}

\section{Temporal Candidate Overlap Analysis}
\label{app:temporal-candidate-overlap}

\providecommand{\code}[1]{\texttt{\detokenize{#1}}}

\paragraph{Why do we look at overlap?}
Temporal candidates are used as an \emph{issue-time-conditioned prior} for bug localization, not as the final prediction. A basic sanity check is whether candidate sets are \emph{temporally smooth}: issues that are close in time should share more candidates than issues that are far apart or randomly paired.
Importantly, for bug localization in code repositories, \textbf{ground-truth patch nodes (GT) are often sparse and diverse}, so \textbf{GT overlap between nearby issues can be low}. Low GT overlap is therefore not necessarily a failure mode; it primarily indicates that consecutive changes may touch different concrete functions/files even within the same development window.

\subsection{Overlap Beyond Adjacent Issues (Larger Window)}
\label{app:temporal-candidate-overlap-lag}

We extend the overlap analysis from strictly adjacent issue pairs to a \emph{time-lag} setting. For each repository, we sort issues by \code{ts\_query} and pair issue $i$ with issue $i+\ell$ (lag $\ell$ in the sorted order). For each issue, we take Top-$K$ candidates (here $K{=}200$), deduplicate them into a set $C$, and compute:
\begin{equation}
\textsc{NZ}(\ell) = \Pr\left(|C_i \cap C_{i+\ell}| > 0\right), \quad
\textsc{Jacc}(\ell) = \mathbb{E}\left[\frac{|C_i \cap C_{i+\ell}|}{|C_i \cup C_{i+\ell}|}\right].
\end{equation}
For reference, we compute the same statistics for GT sets ($G_i$ from \code{patch\_related\_node\_ids}). All numbers are \textbf{MacroAvg over the 9 eval repositories} (each repo has equal weight).

\begin{table}[t]
\centering
\caption{Overlap vs. lag $\ell$ (Top-$K{=}200$). \textsc{NZ} is the percentage of non-empty intersections; \textsc{Jacc} is the mean Jaccard similarity. ``GT'' uses \code{patch\_related\_node\_ids}.}
\label{tab:temporal-candidate-overlap-lag}
\resizebox{\columnwidth}{!}{
\begin{tabular}{@{}c|cc|cc|cc|cc@{}}
\toprule
\multirow{2}{*}{Lag $\ell$} &
\multicolumn{2}{c|}{CRAFT (co-change)} &
\multicolumn{2}{c|}{DyGFormer (co-change)} &
\multicolumn{2}{c|}{GETv2 (issue-conditioned)} &
\multicolumn{2}{c}{GT (patch nodes)} \\
 & \textsc{NZ} & \textsc{Jacc} & \textsc{NZ} & \textsc{Jacc} & \textsc{NZ} & \textsc{Jacc} & \textsc{NZ} & \textsc{Jacc} \\
\midrule
1  & 35.9\% & 0.118 & 35.9\% & 0.118 & 38.7\% & 0.102 & 5.5\% & 0.013 \\
2  & 31.5\% & 0.090 & 31.5\% & 0.090 & 37.3\% & 0.088 & 4.2\% & 0.009 \\
5  & 25.7\% & 0.061 & 25.7\% & 0.061 & 34.0\% & 0.066 & 2.8\% & 0.005 \\
10 & 20.3\% & 0.041 & 20.3\% & 0.041 & 30.7\% & 0.049 & 1.6\% & 0.003 \\
\midrule
Random & 3.1\% & 0.006 & 2.9\% & 0.005 & 5.7\% & 0.005 & -- & -- \\
\bottomrule
\end{tabular}
}
\end{table}

\noindent\textbf{Interpretation.}
(1) Candidate overlap decreases smoothly as $\ell$ increases, while remaining substantially above random baselines, supporting the existence of a temporal locality signal.
(2) GT overlap is much lower and drops quickly with $\ell$, which is expected in bug localization: consecutive PRs can be temporally close but touch different patch nodes.

\subsection{Is Candidate Overlap Simply Anchor/GT Overlap? (Control Statistics)}
\label{app:temporal-candidate-overlap-control}

Candidate sets are conditioned on anchors, so a natural concern is that the overlap signal may be trivially explained by \emph{anchor overlap} or \emph{GT overlap}. To control for this, we compute overlap for \textbf{(i) candidates}, \textbf{(ii) the anchors used by the dump}, and \textbf{(iii) GT patch nodes}, all under the same adjacent-pair protocol. In addition, we report a small but diagnostic case rate:
\begin{equation}
\textsc{Case\%} = \Pr\big(|G_A \cap G_B|>0 \ \wedge\ |C_A \cap C_B|>0 \ \wedge\ |A_A \cap A_B|=0\big),
\end{equation}
where $A$ denotes the anchors used for candidate generation. A non-zero \textsc{Case\%} means that candidate overlap cannot be fully attributed to identical (or overlapping) anchor inputs.

\begin{table}[t]
\centering
\caption{Control statistics for adjacent issue pairs (Top-$K{=}200$). \textsc{NZ} is the non-empty intersection rate; \textsc{Jacc} is mean/median Jaccard. ``GT'' uses \code{patch\_related\_node\_ids}. Numbers are MacroAvg over the 9 eval repositories.}
\label{tab:temporal-candidate-overlap-control}
\resizebox{\columnwidth}{!}{
\begin{tabular}{@{}l|cc|cc|cc|c@{}}
\toprule
\multirow{2}{*}{Source} &
\multicolumn{2}{c|}{Candidates} &
\multicolumn{2}{c|}{Anchors used} &
\multicolumn{2}{c|}{GT (patch nodes)} &
\multirow{2}{*}{\textsc{Case}\%} \\
 & \textsc{NZ} & \textsc{Jacc} & \textsc{NZ} & \textsc{Jacc} & \textsc{NZ} & \textsc{Jacc} & \\
\midrule
CRAFT (FULL86) & 35.9\% & 0.118/0.006 & 28.2\% & 0.040/0.000 & 5.5\% & 0.013/0.000 & 0.93\% \\
DyGFormer (FULL86) & 35.9\% & 0.118/0.006 & 28.2\% & 0.040/0.000 & 5.5\% & 0.013/0.000 & 0.93\% \\
GETv2 (FULL86) & 38.7\% & 0.102/0.019 & 26.2\% & 0.037/0.000 & 5.3\% & 0.013/0.000 & 1.27\% \\
\bottomrule
\end{tabular}
}
\end{table}

\noindent In short, GT overlap is low, anchor overlap is moderate, while candidate overlap is substantially higher. The non-zero \textsc{Case\%} indicates that temporal candidate smoothness is not purely an artifact of overlapping anchors.

\subsection{Concrete Examples}
\label{app:temporal-candidate-overlap-examples}

\paragraph{Example 1: Co-change candidates (CRAFT), disjoint anchors but shared GT is retrieved.}

\textbf{Repo: dvc.} We consider two temporally adjacent issues with $\Delta t \approx 3$ (in \code{ts\_query} units).
The \textbf{anchor sets used by the candidate dump are disjoint} (anchor overlap $=0$), while the candidate sets still have a non-trivial overlap ($|C_A \cap C_B|{=}23$, Jaccard $=0.264$).
Crucially, the two issues have a \textbf{large GT intersection} ($|G_A \cap G_B|{=}12$), and \textbf{all 12 shared GT nodes appear in both issues' Top-$K$ candidate lists} (with $K{=}200$).

\begin{itemize}
    \item Issue A: \code{issue\_id=569} (PR \#1661), \code{``remote local: add dir state update after processing the files''}.
    Key files: \code{dvc/remote/local.py}, \code{tests/test\_add.py}. Key diff context includes \code{def \_save\_dir(...)}.
    \item Issue B: \code{issue\_id=570} (PR \#1662), \code{``stage: check if local path contains symlink ...''}.
    Key files: \code{dvc/stage.py}, \code{dvc/utils/fs.py}, \code{tests/test\_add.py}. Key diff contexts include \code{def \_stage\_fname(...)} and \code{def get\_mtime\_and\_size(...)}.
\end{itemize}

\noindent\textbf{Shared GT nodes and their ranks in the candidate list.}
The shared GT nodes $G_A \cap G_B$ and their ranks in each issue's candidate list are shown below. Unlike the previous example, these shared GT nodes are retrieved in \emph{both} issues:
\begin{center}
\begin{tabular}{@{}rrr@{}}
\toprule
Shared GT node (\code{orig\_node\_id}) & Rank in Issue A & Rank in Issue B \\
\midrule
47451 & 13 & 8 \\
47453 & 19 & 10 \\
47455 & 24 & 12 \\
47457 & 14 & 19 \\
47459 & 5 & 14 \\
47461 & 3 & 13 \\
47463 & 4 & 15 \\
47465 & 21 & 27 \\
47467 & 17 & 23 \\
47469 & 2 & 1 \\
47471 & 8 & 4 \\
47473 & 7 & 3 \\
\bottomrule
\end{tabular}
\end{center}

\noindent\textbf{Shared top candidates and human-readable evidence.}
Table~\ref{tab:overlap-dvc-evidence} lists several shared top candidates (by minimum rank across the two issues), together with an evidence PR where the node appears in GT and the corresponding patch context.
\begin{table}[h]
\centering
\caption{Example 1 (dvc): shared top candidates (CRAFT) with evidence. The evidence PR is obtained by back-looking up the candidate node in \code{patch\_related\_node\_ids} and then reading the corresponding PR patch.}
\label{tab:overlap-dvc-evidence}
\resizebox{\columnwidth}{!}{
\begin{tabular}{@{}rrrrlll@{}}
\toprule
\code{orig\_node\_id} & Rank A & Rank B & Evidence Issue & PR \# & Key file(s) & Patch context (subset) \\
\midrule
50171 & 0 & 0 & 567 & 1647 & \code{dvc/remote/local.py} & \code{def changed\_cache(self, md5):} \\
47469 & 2 & 1 & 555 & 1583 & \code{dvc/project.py} & \code{def add(self, fname, recursive=False):} \\
50178 & 6 & 2 & 567 & 1647 & \code{dvc/state.py} & \code{def changed\_cache(self, md5):} \\
47461 & 3 & 13 & 555 & 1583 & \code{tests/test\_add.py} & \code{def add(self, fname, recursive=False):} \\
\bottomrule
\end{tabular}
}
\end{table}

\noindent This illustrates that co-change temporal candidates can capture a stable ``active area'' over time in DVC (local remote/state caching and project-level \code{add}), even when issue anchors are not identical. In this pair, the temporal continuity is also reflected by the large shared GT set and its strong coverage in both candidate lists.

\paragraph{Example 2: Co-change candidates (DyGFormer), disjoint anchors but shared GT is retrieved.}

\textbf{Repo: xarray.} We consider two temporally nearby issues with $\Delta t \approx 26$ (in \code{ts\_query} units; lag$=10$ in the time-sorted sequence).
The anchors used for candidate generation are \textbf{disjoint} (anchor overlap $=0$), while the candidate sets still have a strong overlap ($|C_A \cap C_B|{=}32$, Jaccard $=0.552$).
The two issues also have a relatively large shared patch set ($|G_A \cap G_B|{=}37$), among which \textbf{18 shared GT nodes are retrieved in both issues' Top-$K$ candidate lists} ($K{=}200$).

\begin{itemize}
    \item Issue A: \code{issue\_id=3256} (PR \#8780), \code{introduce .vindex property for Explicitly Indexed Arrays}.
    Key files include \code{xarray/core/indexing.py} and \code{xarray/core/variable.py}, with contexts such as \code{def transpose(self, order):} and \code{def \_oindex\_get(self, key):}.
    \item Issue B: \code{issue\_id=3281} (PR \#8857), \code{increase typing annotations coverage in xarray/core/indexing.py}.
    Key files include \code{xarray/core/indexing.py}, \code{xarray/namedarray/core.py}, and \code{xarray/tests/test\_indexing.py}, with contexts such as \code{def map\_index\_queries(...):} and \code{class ExplicitIndexer:}.
\end{itemize}

\noindent\textbf{Shared GT nodes and ranks.}
The shared GT nodes that are retrieved in both issues (under DyGFormer candidates) are listed below, together with their ranks in each candidate list (0-based):
\begin{center}
\begin{tabular}{@{}rrr@{}}
\toprule
Shared GT node (\code{orig\_node\_id}) & Rank in Issue A & Rank in Issue B \\
\midrule
796935 & 4 & 3 \\
796941 & 7 & 7 \\
796946 & 9 & 9 \\
796956 & 13 & 12 \\
796964 & 15 & 15 \\
796871 & 19 & 19 \\
796875 & 20 & 22 \\
796876 & 21 & 23 \\
796877 & 22 & 24 \\
796882 & 24 & 26 \\
796883 & 25 & 27 \\
796884 & 26 & 28 \\
796892 & 29 & 48 \\
796901 & 31 & 34 \\
796902 & 32 & 35 \\
796908 & 35 & 37 \\
796910 & 36 & 39 \\
796911 & 37 & 40 \\
\bottomrule
\end{tabular}
\end{center}

\noindent This pair is a concrete example where candidate overlap and shared GT hits persist even when the anchors used for candidate generation are disjoint, supporting the interpretation that temporal candidates encode a stable ``active region'' prior beyond trivial anchor overlap.

\paragraph{Example 3: Issue-conditioned candidates (GETv2), high overlap and perfect GT coverage.}
\textbf{Repo: astropy.} We consider two temporally adjacent issues with $\Delta t \approx 7$.
GETv2 produces candidate sets with strong overlap ($|C_A \cap C_B|{=}75$, Jaccard $=0.424$), and \textbf{both issues' GT nodes are fully covered by Top-$K$ candidates} (GT coverage $=1.0$ for both issues under $K{=}200$).

\begin{itemize}
    \item Issue A: \code{issue\_id=4147} (PR \#10814), \code{``Simplify prepare\_earth\_position\_vel ...''}. Key file: \code{astropy/coordinates/builtin\_frames/utils.py}.
    \item Issue B: \code{issue\_id=4169} (PR \#10881), \code{``fix division by zero warnings for values near sun''}. Key file: \code{astropy/coordinates/builtin\_frames/utils.py}, with contexts including \code{def aticq(...)} and \code{def atciqz(...)}.
\end{itemize}

\noindent\textbf{GT nodes and their ranks.}
\begin{center}
\begin{tabular}{@{}lrr@{}}
\toprule
Issue & GT node(s) (\code{orig\_node\_id}) & Rank(s) in Top-$K$ candidates \\
\midrule
\code{issue\_id=4147} & 910572 & 1 \\
\code{issue\_id=4169} & 910571, 910570 & 3, 26 \\
\bottomrule
\end{tabular}
\end{center}

\noindent\textbf{Shared candidates with evidence (subset).}
Table~\ref{tab:overlap-astropy-evidence} shows a subset of shared candidates together with representative evidence PRs and contexts from the coordinates stack.
\begin{table}[h]
\centering
\caption{Example 2 (astropy): shared candidates (GETv2) with evidence.}
\label{tab:overlap-astropy-evidence}
\resizebox{\columnwidth}{!}{
\begin{tabular}{@{}rrrrlll@{}}
\toprule
\code{orig\_node\_id} & Rank A & Rank B & Evidence Issue & PR \# & Key file(s) & Patch context (subset) \\
\midrule
910572 & 1 & 1 & 4147 & 10814 & \code{astropy/coordinates/builtin\_frames/utils.py} & \code{prepare\_earth\_position\_vel}; \code{epv00} \\
923708 & 2 & 6 & 4004 & 10475 & \code{astropy/coordinates/attributes.py} & \code{def transform\_to(...)}; \code{def gcrs\_to\_gcrs(...)} \\
910615 & 53 & 2 & 4003 & 10474 & \code{astropy/coordinates/...} & \code{coordinates} (remote\_data cleanup; related tests) \\
\bottomrule
\end{tabular}
}
\end{table}

\noindent Overall, GETv2 retrieves a coherent neighborhood of temporally related code entities in the coordinates subsystem, which provides a useful prior for downstream bug localization.

To isolate the effect of our temporal-candidates module and GNN reranking, we also include a broad set of \emph{single-model} temporal GNN baselines that operate directly on the commit co-change interaction stream.
These baselines do not inject temporal candidates nor use our reranker, and thus serve as a reference point for the temporal backbone capacity under the same eval9 protocol.

\begin{table*}[t]
\centering
\caption{Dynamic temporal GNN baselines without temporal-candidates injection (eval9). We evaluate each model as a single temporal graph model on the commit co-change interaction graph. Hit@K is mean over issues of $|\mathrm{GT}\cap \mathrm{TopK}|/|\mathrm{GT}|$ (empty GT $\rightarrow$ 0). CandCov.\ is the fraction of issues whose ground truth intersects the candidate list.}
\label{tab:appendix-tgnn-baselines}
\begin{tabular}{llrrrrr}
\toprule
Model & Setting & Hit@1 & Hit@5 & Hit@10 & Hit@20 & CandCov. \\
\midrule
DyGFormer & 1-hop & 5.69\% & 21.65\% & 27.99\% & 31.06\% & 62.8\% \\
GraphMixer & 1-hop & 6.61\% & 25.32\% & 32.05\% & 34.35\% & 62.8\% \\
TGAT & 1-hop & 6.73\% & 25.39\% & 32.13\% & 34.49\% & 62.8\% \\
TGN & 1-hop & 5.97\% & 22.59\% & 28.85\% & 31.92\% & 62.8\% \\
DyRep & 1-hop & 5.78\% & 22.01\% & 28.46\% & 31.62\% & 62.8\% \\
CAWN & 1-hop & 5.72\% & 21.74\% & 27.94\% & 31.06\% & 62.8\% \\
TCL & 1-hop & 6.67\% & 25.10\% & 31.79\% & 34.18\% & 62.8\% \\
\bottomrule
\end{tabular}
\end{table*}

\end{document}